\documentclass{article}

\PassOptionsToPackage{numbers, compress}{natbib}

\usepackage[final]{neurips_data_2023}

\usepackage[utf8]{inputenc} 
\usepackage[T1]{fontenc}    
\usepackage{hyperref}       
\usepackage{url}            
\usepackage{booktabs}       
\usepackage{amsfonts}       
\usepackage{nicefrac}       
\usepackage{microtype}      
\usepackage{xcolor}         

\usepackage{xspace}

\usepackage{hyperref}

\usepackage{soul}
\usepackage{graphicx}
\graphicspath{{./figures/}}
\usepackage{caption}
\usepackage{subcaption}
\usepackage{multirow}
\usepackage{diagbox}
\usepackage{cancel}
\usepackage{float}
\usepackage{threeparttable}
\usepackage{amssymb}
\usepackage{bbding}
\usepackage{pifont}
\usepackage{enumitem}
\usepackage{amsmath}
\usepackage{xcolor}
\usepackage{color}
\usepackage{colortbl}
\usepackage{amsmath}
\usepackage{multicol}

\setenumerate[1]{itemsep=0pt,partopsep=0pt,parsep=\parskip,topsep=0pt,leftmargin=10pt}
\setitemize[1]{itemsep=0pt,partopsep=0pt,parsep=\parskip,topsep=0pt,leftmargin=10pt}

\definecolor{light-gray}{gray}{0.95}

\definecolor{ao}{rgb}{0.0, 0.0, 1.0}
\definecolor{darkcyan}{rgb}{0.0, 0.55, 0.55}
\definecolor{blue}{rgb}{0.0, 0.0, 1.0}
\definecolor{dartmouthgreen}{rgb}{0.05, 0.5, 0.06}
\definecolor{cyan(process)}{rgb}{0.0, 0.72, 0.92}
\definecolor{babypink}{rgb}{0.96, 0.76, 0.76}
\definecolor{cornflowerblue}{rgb}{0.39, 0.58, 0.93}

\newcommand{\fv}[1]{\textcolor{red}{\textbf{#1}}}
\newcommand{\sv}[1]{\textcolor{cornflowerblue}{\underline{#1}}}

\newcommand{\apc}[1]{\textcolor{purple}{#1}}

\definecolor{ao(english)}{rgb}{0.0, 0.5, 0.0}
\newcommand{\cmark}{\color{ao(english)}\Checkmark}

\newcommand{\sys}{\textsc{BenchTemp}\xspace}

\title{\sys: A General Benchmark for Evaluating Temporal Graph Neural Networks}

\author{%
	Qiang Huang$^{1}$ ~ Jiawei Jiang$^{1}$ ~ Susie Xi Rao$^{2}$ ~ Ce Zhang$^{2}$ ~ Zhichao Han$^{3}$ ~ \textbf{Zitao Zhang}$^{3}$  \\
	\textbf{Xin Wang}$^{1}$ ~ \textbf{Yongjun He}$^{2}$~ \textbf{Quanqing Xu}$^{4}$ ~ \textbf{Yang Zhao}$^{3}$ ~ \textbf{Chuang Hu}$^{1}$ ~ \textbf{Shuo Shang}$^{5}$  ~ \textbf{Bo Du}$^{1}$ \\
	$^{1}$School of Computer Science, Wuhan University
	\quad $^{2}$ETH Z\"{u}rich \quad $^{3}$eBay \\
	$^{4}$OceanBase, Ant Group \quad $^{5}$University of Electronic Science and Technology of China \\
	$^{1}$\texttt{jonnyhuanghnu@gmail.com} \quad $^{1}$\texttt{\{jiawei.jiang,tbbdfl,handc,dubo\}@whu.edu.cn} \\
	$^{2}$\texttt{\{raox,ce.zhang,yongjun.he\}@inf.ethz.ch} \quad $^{3}$\texttt{\{zhihan,zitzhang, yzhao5\}@ebay.com} \\
	$^{4}$\texttt{xuquanqing.xqq@oceanbase.com} \quad
	$^{5}$\texttt{jedi.shang@gmail.com} \\
}

\begin{document}
	
	\maketitle
	
	\begin{abstract}
		To handle graphs in which features or connectivities are evolving over time,
		a series of temporal graph neural networks (TGNNs) have been proposed.
		Despite the success of these TGNNs,
		the previous TGNN evaluations reveal several limitations regarding four critical issues:
		1) inconsistent datasets, 2) inconsistent evaluation pipelines, 3) lacking workload diversity, and 4) lacking efficient comparison.
		Overall, there lacks an empirical study that puts TGNN models onto the same ground and compares them comprehensively.
		To this end, we propose \sys, a general benchmark for evaluating TGNN models on various workloads.
		\sys provides a set of benchmark datasets so that different TGNN models can be fairly compared.
		Further, \sys engineers a standard pipeline that unifies the TGNN evaluation.
		With \sys, we extensively compare the representative TGNN models on different tasks (e.g., link prediction and node classification) and settings (transductive and inductive),
		w.r.t.
		both effectiveness and efficiency metrics.
		We have made \sys publicly available at \url{https://github.com/qianghuangwhu/benchtemp}.
	\end{abstract}

	\section{Introduction}
	\label{sec:intro}
	
	Temporal graphs provide a natural abstraction for many real-world systems, in which the topology structure and node/edge attributes change over time.
	Often, a temporal graph can be modeled as an interaction stream, and it is critical to capture the latent evolution pattern in many domains, such as social network, transportation system, and biology.
	Recently, researchers have proposed a range of temporal graph neural network (TGNN) models, e.g.,
	JODIE~\cite{kumar2019predicting}, DyRep~\cite{trivedi2019dyrep}, TGN~\cite{rossi2020temporal}, TGAT~\cite{xu2020inductive}, CAWN~\cite{wanginductive}, NeurTW~\cite{jin2022neural}, and NAT~\cite{luo2022neighborhood}.
	As shown in Table~\ref{tab:taxonomy},
	these models introduce different techniques,
	such as memory, attention, RNN, and temporal walk, to detect temporal changes and update node representations.
	

	\textbf{Limitation of Prior Works.}
	In the presence of different paradigms of TGNNs,
	it is therefore meaningful to compare their performance in various tasks.
	Although each of the above methods has provided an evaluation study,
	none of them has covered a wide spectrum of the literature.
	A few premier works have tried to compare these models~\cite{poursafaeitowards};
	however,
	if we look at those TGNN benchmarks, we find that:
	
	\begin{quote}
		\em
		The scope of current benchmarks is \underline{limited to some specific tasks}.
		Worse, their evaluation of the existing TGNN models shows controversial and inconsistent results, mainly due to \underline{the lack of standard datasets and processing pipeline}.
	\end{quote}
	
	\vspace{0.3em}
	
	\begin{table}[!b]
		\scriptsize
		\centering
		\caption{\footnotesize Anatomy of state-of-the-art TGNN models.
			\textit{Memory}, \textit{Attention}, \textit{RNN}, and \textit{TempWalk} indicate whether the model adopts memory modules, attention mechanism, RNN, and temporal walk strategy, respectively. 
			\textit{Scalability} means whether the model is claimed to be scalable.
			\textit{LP} and \textit{NC} indicate whether the corresponding work implements link prediction and node classification tasks, respectively. 
			\textit{\# Datasets} and \textit{\# Features} represent the 
			number of evaluated datasets and the number of initial node features.
			\textit{Supervised} indicates whether the method is supervised, unsupervised, or self(semi)-supervised.
		}
		\label{tab:taxonomy}
		\setlength{\tabcolsep}{4pt}
		\begin{tabular}{l|ccccccc|cr|r}
			\toprule
			Model &\textit{Memory}&\textit{Attention}& \textit{RNN} &\textit{TempWalk} & \textit{Scalability}&\textit{LP} & \textit{NC}&\textit{\# Datasets}&\textit{\# Features}& \textit{Supervised}\\
			\midrule
			JODIE &\cmark&\cmark&\cmark&&\cmark& \cmark & \cmark&4&128&\scriptsize{self (semi)-supervised}\\
			DyRep &&\cmark&&&\cmark& \cmark & &2&172&\scriptsize{unsupervised}\\
			TGN&\cmark&\cmark&\cmark&& &\cmark&\cmark&4&172&\scriptsize{self (semi)-supervised}\\
			TGAT&&\cmark&&& &\cmark&\cmark&3&172&\scriptsize{self (semi)-supervised}\\
			NAT&\cmark&\cmark&\cmark&&\cmark&\cmark& &8&172, 32, 2&\scriptsize{self-supervised}\\
			\midrule
			CAWN &&\cmark&\cmark&\cmark&\cmark& \cmark & &6&172, 32, 2& \scriptsize{self-supervised}\\
			NeurTW&&&\cmark&\cmark &&\cmark&\cmark&6&172, 32, 4& \scriptsize{self (semi)-supervised}\\
			\bottomrule
		\end{tabular}  
		
	\end{table}

	For example, EdgeBank~\cite{poursafaeitowards} compares a set of TGNNs only on link prediction task and does not unify the data loaders (e.g., train/validation/test split and inductive masking);
	some works choose different dimensions of initial node features for the evaluated temporal graphs~\cite{wanginductive, jin2022neural, luo2022neighborhood}; some works use different seeds for edge samplers~\cite{rossi2020temporal, xu2020inductive, wanginductive}.
	Without unifying the datasets and evaluation pipeline, previous work unsurprisingly reports different relative performance for the same set of TGNNs~\cite{rossi2020temporal,xu2020inductive,wanginductive,jin2022neural, luo2022neighborhood,poursafaeitowards}.

	\textbf{Our Goals.}
	Despite the efforts of previous works,
	the landscape of TGNN has not been fully explored.
	We try to achieve the following goals, which are vital, yet unsolved, for a mature TGNN benchmark.

	\begin{itemize}
		\item{\em Benchmark datasets.}
		A vital reason for the inconsistent results in previous work is that the temporal graph models are not compared using the same datasets.
		The original graph is preprocessed in different manners, and the temporal order is handled from different perspectives.
		Without standard datasets and their preprocessing steps, it is not possible to compare different TGNN models fairly.
		
		\item{\em A unified pipeline.}
		The execution of a TGNN model naturally contains multiple phases, including but not limited to data loader, graph sampler, trainer, etc.
		Unfortunately, there is no benchmark library that establishes a general processing pipeline for temporal graphs.
		
		\item{\em Diverse workloads.}
		Temporal graphs evolve over time, and therefore there exist diverse workloads.
		For example, there are transductive and inductive tasks, and the inductive task has New-Old and New-New scenarios (see Section~\ref{sec:pipeline_lp}).
		A solid benchmark should include diverse workloads.
		
		\item{\em Efficiency comparison.}
		In addition to model quality, it is also indispensable to compare the efficiency performance of models to help us understand their performance trade-offs.
		However, this issue has not been well studied in the literature.
		
	\end{itemize}

	In this work, we conduct a comprehensive benchmark on the state-of-the-art TGNN models.
	To avoid orange-to-apple comparison, we first study how to construct standard datasets, 
	and then study how to build a framework that evaluates various TGNN models in a unified pipeline.
	The major contributions of this work are summarized below.

	\begin{enumerate}
		\item We present \sys, the first general benchmark for evaluating temporal graph neural network (TGNN) models over a wide range of tasks and settings.
		
		\item We construct a set of benchmark temporal graph datasets and
		a unified benchmark pipeline.
		In this way, we standardize the entire lifecycle of benchmarking TGNNs.
		
		\item On top of \sys, we extensively compare representative TGNN models on the benchmark datasets, regarding different tasks, settings, metrics, and efficiency.
		
		\item 
		We thoroughly discuss the empirical results and draw insights for TGNN.
		Further, we have released the \href{https://drive.google.com/drive/folders/1HKSFGEfxHDlHuQZ6nK4SLCEMFQIOtzpz?usp=sharing}{datasets}, \href{https://pypi.org/project/benchtemp/}{PyPI}, and \href{https://my-website-6gnpiaym0891702b-1257259254.tcloudbaseapp.com/}{leaderboard}.
		Please refer to our \href{https://github.com/qianghuangwhu/benchtemp}{github project} and \href{https://github.com/qianghuangwhu/benchtemp/blob/master/appendix.pdf}{appendix} for details.
		
	\end{enumerate}

	\begin{figure}[!t]
		\centering
		\includegraphics[width=0.92\textwidth]{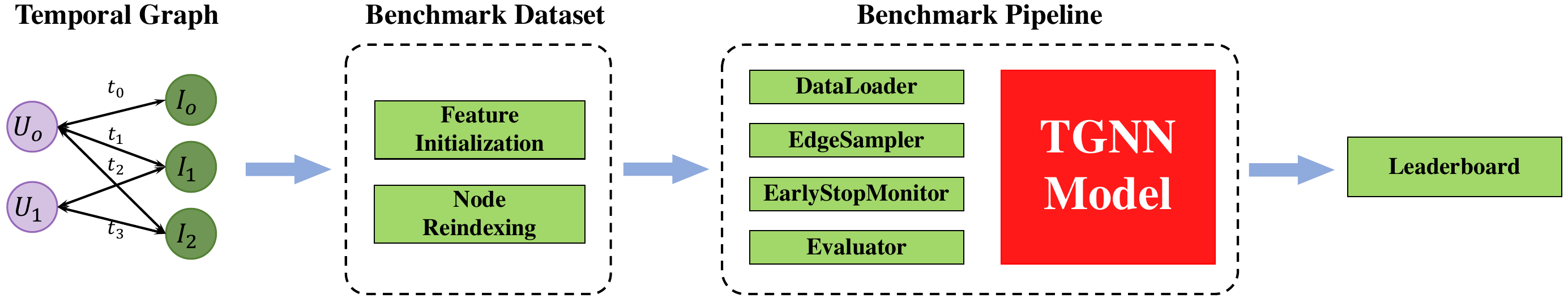}
		\caption{\footnotesize Overview of \sys.}
		\label{fig:overview}
	\end{figure}
	
	\section{Anatomy of Current TGNN Models}
	\label{sec:models}
	
	Before presenting our benchmark, we revisit the current TGNN models and investigate the following question:
	{\em How do the existing models handle temporal graphs? 
		Are there problems and inconsistencies during their processing procedures?}
	
	
	We consider seven state-of-the-art TGNN models --- JODIE, DyRep, TGN, TGAT, NeurTW, CAWN, and NAT~\cite{kumar2019predicting,trivedi2019dyrep,rossi2020temporal, xu2020inductive, wanginductive, jin2022neural, luo2022neighborhood}.
	We summarize the anatomy results of these models in Table~\ref{tab:taxonomy}. 
	JODIE~\cite{kumar2019predicting} models the dynamic evolution of user-item interactions through joint user and item embeddings, which are updated by two RNNs. 
	Furthermore, JODIE trains a temporal attention layer to predict trajectories.
	DyRep~\cite{trivedi2019dyrep} implements a temporal attention mechanism to encode the evolution of temporal structure information into interleaved and nonlinear dynamics over time.
	TGN~\cite{rossi2020temporal} establishes a memory module to memorize node representations, and uses RNN and graph operators to update the memory.
	TGAT~\cite{xu2020inductive} utilizes self-attention to aggregate temporal-topological structure, and learns the temporal evolution of node embeddings through a continuous-time encoding.
	CAWN~\cite{wanginductive} samples a set of causal anonymous walks (CAWs) by temporal random walks, retrieves motifs to represent network dynamics, and encodes and aggregates these CAWs via RNNs.
	NeurTW~\cite{jin2022neural} introduces a spatio-temporal random walk to retrieve representative motifs and irregularly-sampled temporal nodes, and then learns time-aware node representations.
	NAT~\cite{luo2022neighborhood} constructs structural features using joint neighborhoods of different nodes, and designs a data structure to support parallel access and update of node representations.

	\textbf{The Landscape of TGNNs.}
	As seen in Table~\ref{tab:taxonomy},
	the abovementioned models choose different techniques and perform different \underline{tasks} (some do not implement the node classification task~\cite{trivedi2019dyrep,wanginductive,luo2022neighborhood,poursafaeitowards}).
	Regarding the \underline{datasets},
	they choose different dimensions (e.g., 2, 4, 32, 128, and 172) for the initial node features and cause potentially inconsistent results.
	Regarding \underline{negative edge sampler}, both fixed~\cite{kumar2019predicting, trivedi2019dyrep, rossi2020temporal} and random~\cite{xu2020inductive, wanginductive, jin2022neural, luo2022neighborhood} seeds are used.
	Regarding \underline{node visibility},
	JODIE, DyRep, TGN, TGAT, and NAT choose traditional transductive and inductive settings, while
	CAWN and NeurTW further consider inductive New-New and inductive New-Old settings.
	Besides, most models only report model performance metrics and neglect efficiency metrics.

	\section{Temporal Graph Benchmark}
	
	In this section, we describe our proposed \sys, including the construction of benchmark datasets and the general benchmark pipeline, as shown in Figure~\ref{fig:overview}.
	
	
	\subsection{Benchmark Datasets}
	\label{subsec:dataset} 
	
	\textbf{Formalization.}
	A temporal graph can be represented as an ordered sequence of temporal interactions. The $r$-th interaction $I_{r}=(u_{r}, i_{r}, t_{r}, e_{r})$ happens at time $t_{r}$ between the source node $u_{r}$ and the destination node $i_{r}$ with edge feature $e_{r}$. 
	
	\begin{table}[!t]
		\tiny
		\centering
		\caption{\footnotesize Dataset statistics.}
		\label{tab:datasets}
		\begin{tabular}{llrrrrcc}
			\toprule
			\multirow{2}{*}{Dataset}  & \multirow{2}{*}{\textit{Domain}} & \multirow{2}{*}{\textit{\# Nodes}} & \multirow{2}{*}{\textit{\# Edges}} &\multicolumn{1}{c}{\textit{Avg. Degree}}& \multicolumn{1}{c}{\textit{Edge Density}} &  \multirow{2}{*}{\textit{Heterogeneous} }&\multirow{2}{*}{\textit{Homogeneous}}\\
			&&&&$\left(\frac{\# Edges}{\# Nodes}\right)$&$\left(\frac{\# Edges}{\# Users\times{\# Items}} \right)$&&\\
			\midrule
			\href{http://snap.stanford.edu/jodie/reddit.csv}{Reddit}~\cite{kumar2019predicting,url_reddit} &	Social&	10,984&	672,447&61.22 &0.06&\cmark&\\
			\href{http://snap.stanford.edu/jodie/wikipedia.csv}{Wikipedia}~\cite{kumar2019predicting}	&Social	&9,227	&157,474&17.07& 0.01	&\cmark&\\
			\href{http://snap.stanford.edu/jodie/mooc.csv}{MOOC}~\cite{kumar2019predicting, url_mooc} &	Interaction	&7,144&	411,749&57.64&0.60 &	\cmark&\\
			\href{http://snap.stanford.edu/jodie/lastfm.csv}{LastFM}~\cite{kumar2019predicting,hidasi2012fast}&	Interaction&	1,980	&1,293,103&653.08&1.32 	&\cmark&\\
			\href{https://tianchi.aliyun.com/dataset/649}{Taobao}~\cite{jin2022neural, zhu2018learning}	&E-commerce&	82,566	&77,436&0.94 &5.55	&\cmark &\\
			\href{https://www.cs.cmu.edu/~./enron/}{Enron}~\cite{url_enron,leskovec2014snap}&	Social	&184	&125,235&	680.63& 3.76	& &\cmark\\
			\href{http://realitycommons.media.mit.edu/socialevolution.html}{SocialEvo}~\cite{madan2011sensing}	&Proximity&	74	&2,099,519&	28,371.88 &405.31	& &\cmark\\
			\href{http://konect.cc/networks/opsahl-ucforum/}{UCI}~\cite{opsahl2009clustering}	&Social	&1,899&	59,835	&31.51 &0.02	& &\cmark\\
			\href{http://snap.stanford.edu/data/CollegeMsg.html}{CollegeMsg}~\cite{panzarasa2009patterns}&Social&	1,899	&59,834	&31.51& 0.02& &\cmark\\
			\href{https://github.com/shenyangHuang/LAD}{CanParl}~\cite{poursafaeitowards, huang2020laplacian}&	Politics&	734	&74,478	&101.47 &0.42	& &\cmark\\
			\href{https://springernature.figshare.com/articles/dataset/Metadata_record_for_Interaction_data_from_the_Copenhagen_Networks_Study/11283407/1}{Contact}~\cite{poursafaeitowards,sapiezynski2019interaction}	&Proximity	&692	&2,426,279		&3,506.18&5.31 & &\cmark\\
			\href{https://zenodo.org/record/3974209/#.Yf62HepKguU}{Flights}~\cite{poursafaeitowards,schafer2014bringing}&	Transport	&13,169	&1,927,145	&146.34 	&0.01& &\cmark\\
			\href{https://www.fao.org/faostat/en/#data/TM}{UNTrade}~\cite{poursafaeitowards,macdonald2015rethinking}	&Economics&	255&	507,497&1,990.18 &7.84	& &\cmark\\
			\href{https://github.com/shenyangHuang/LAD}{USLegis}~\cite{poursafaeitowards,huang2020laplacian}&	Politics&	225	&60,396	&268.43&1.19 & &\cmark\\
			\href{https://dataverse.harvard.edu/dataset.xhtml?persistentId=doi:10.7910/DVN/LEJUQZ}{UNVote}~\cite{poursafaeitowards,DVN/LEJUQZ_2009}	&Politics	&201	&1,035,742	&5,152.95 &	25.6& &\cmark\\

			\bottomrule
		\end{tabular}
		
	\end{table}
	
	\textbf{Datasets.}
	We choose a large set of popular temporal graph datasets from various domains~\cite{kumar2019predicting,jin2022neural,poursafaeitowards}, on which we build the benchmark datasets  attached CC BY-NC \href{https://creativecommons.org/licenses/by-nc/4.0/}{license}.
	The statistics of these datasets are summarized in Table \ref{tab:datasets}. Please refer to Appendix~\apc{A} for dataset details.

	\textbf{Benchmark Dataset Construction.}
	\sys standardizes the construction of temporal graph benchmark datasets with two steps:
	{\em node feature initialization} and
	{\em node reindexing}.

	
	
	\begin{figure}[!t]
		\centering
		\begin{subfigure}[b]{0.3\textwidth}
			\centering
			\includegraphics[width=\textwidth]{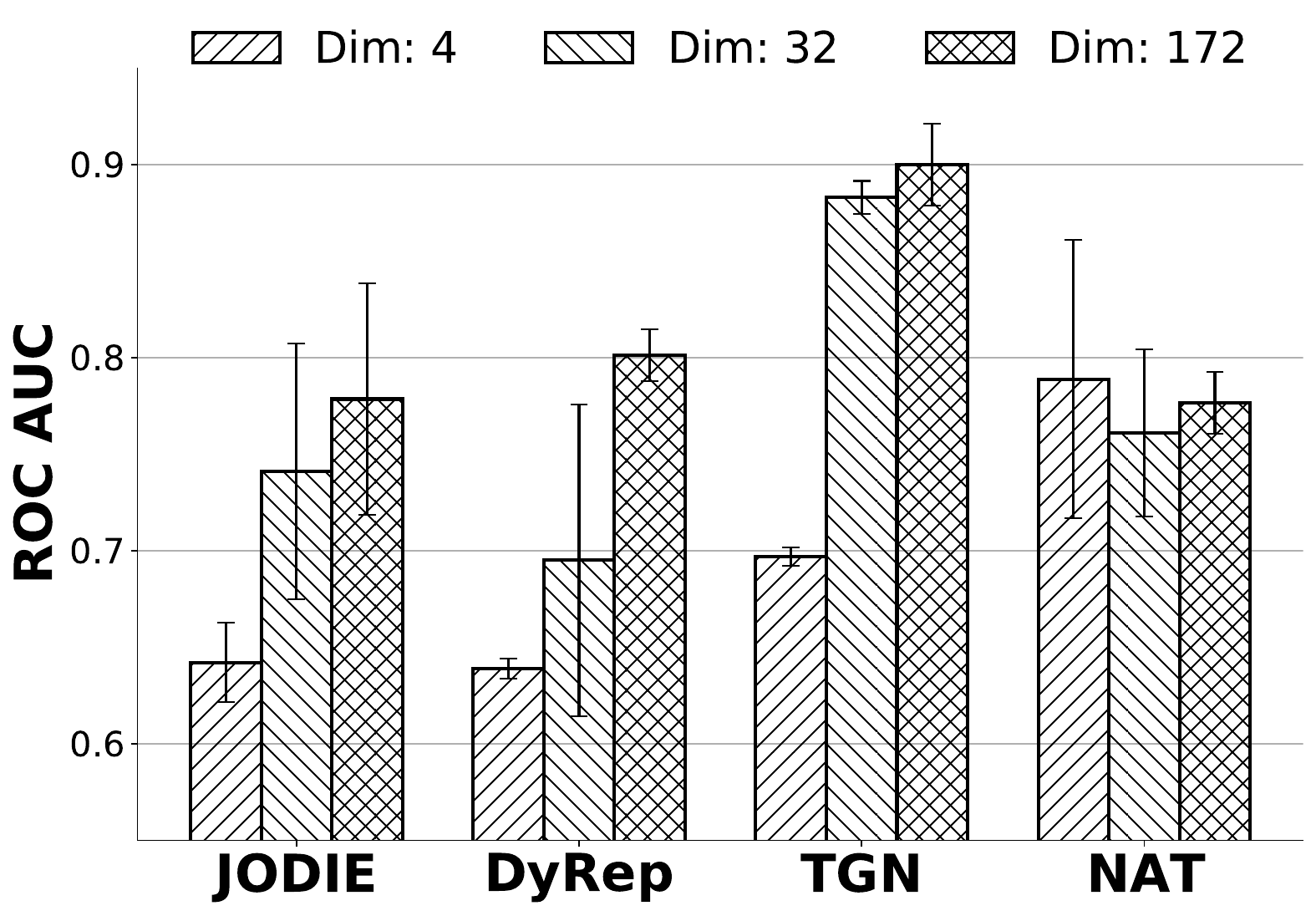}
			\caption{\footnotesize Transductive}
			\label{dimension-Transductive}
		\end{subfigure}
		\begin{subfigure}[b]{0.3\textwidth}
			\centering
			\includegraphics[width=\textwidth]{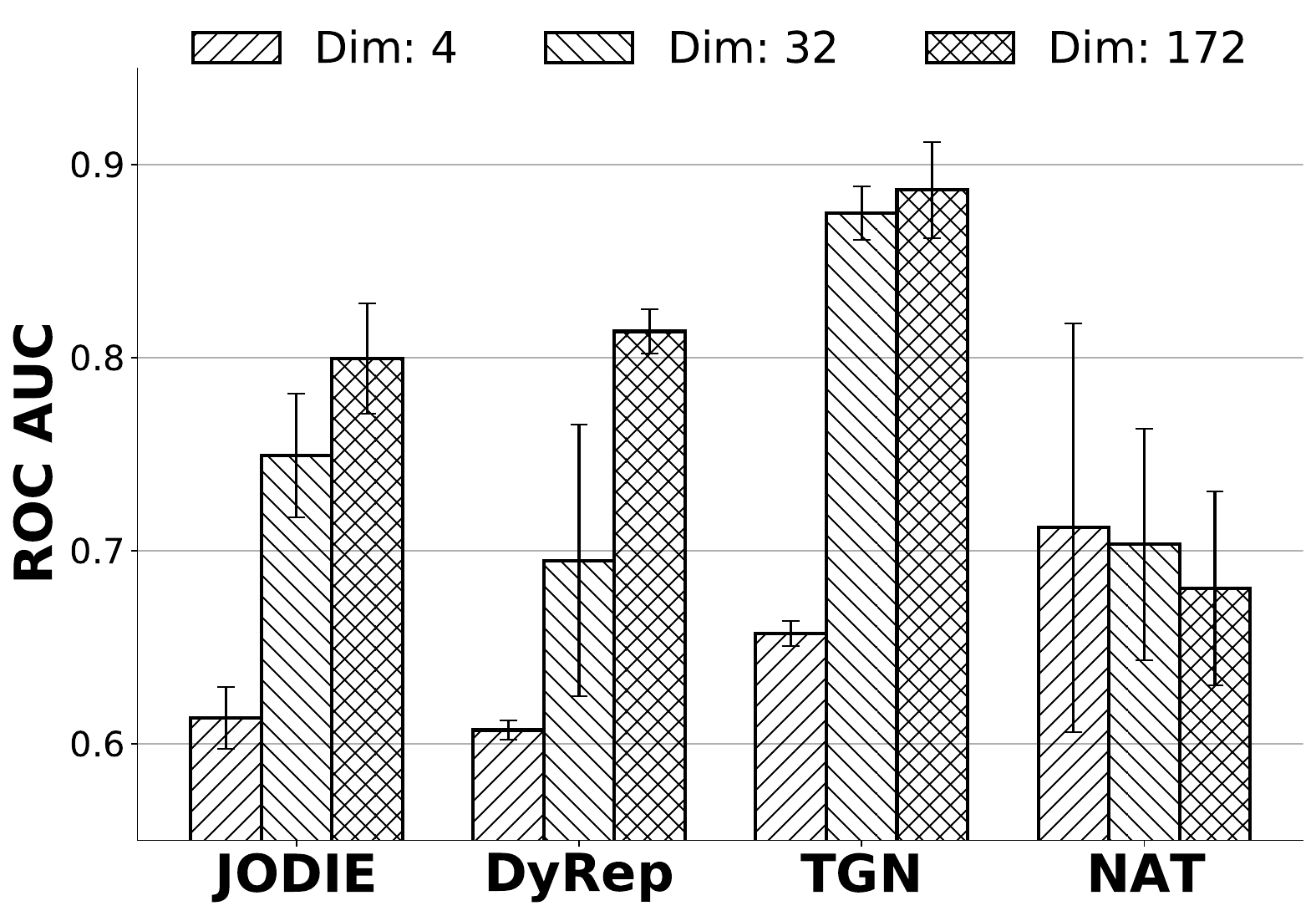}
			\caption{\footnotesize Inductive}
			\label{dimension-Inductive}
		\end{subfigure}
		\vspace{-0.5em}
		\caption{\footnotesize Link prediction results on MOOC dataset with different dimensions for initial node features.}
		\label{fig:dimension}
	\end{figure}

	\textbf{Node Feature Initialization.}
	The prior works choose different dimensions for the initial node features. As the results in Figure~\ref{fig:dimension} show, as the node feature dimension increases from 4 to 172, most TGNN models achieve higher ROC AUC.
	Consequently, to address the discrepancy w.r.t. node feature dimension, we set it to 172 for all the datasets, {\em since it is the most commonly chosen value in prior works}.
	

	\begin{figure}[!t]
		\centering
		\begin{subfigure}[b]{0.4\textwidth}
			\centering
			\includegraphics[width=\textwidth]{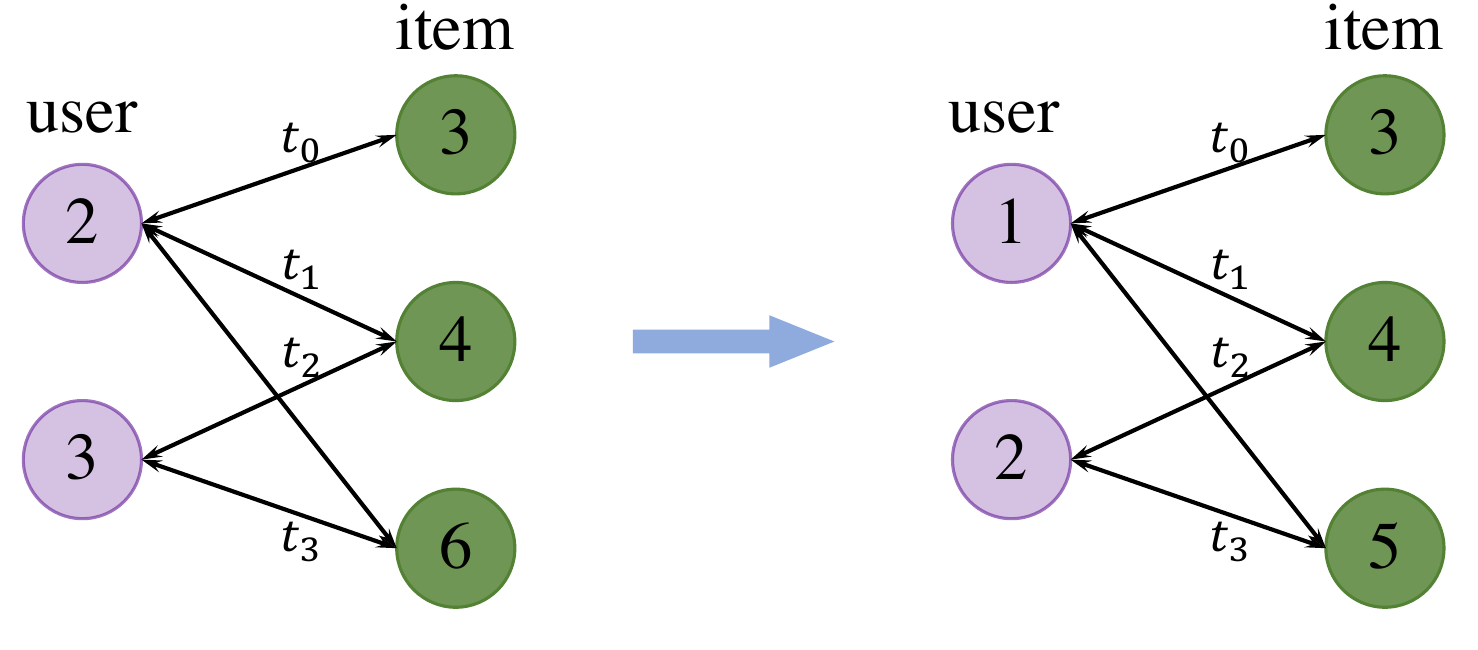}
			\caption{\footnotesize Heterogeneous graph}
			\label{bipartite}
		\end{subfigure}
		\hfill
		\begin{subfigure}[b]{0.4\textwidth}
			\centering
			\includegraphics[width=\textwidth]{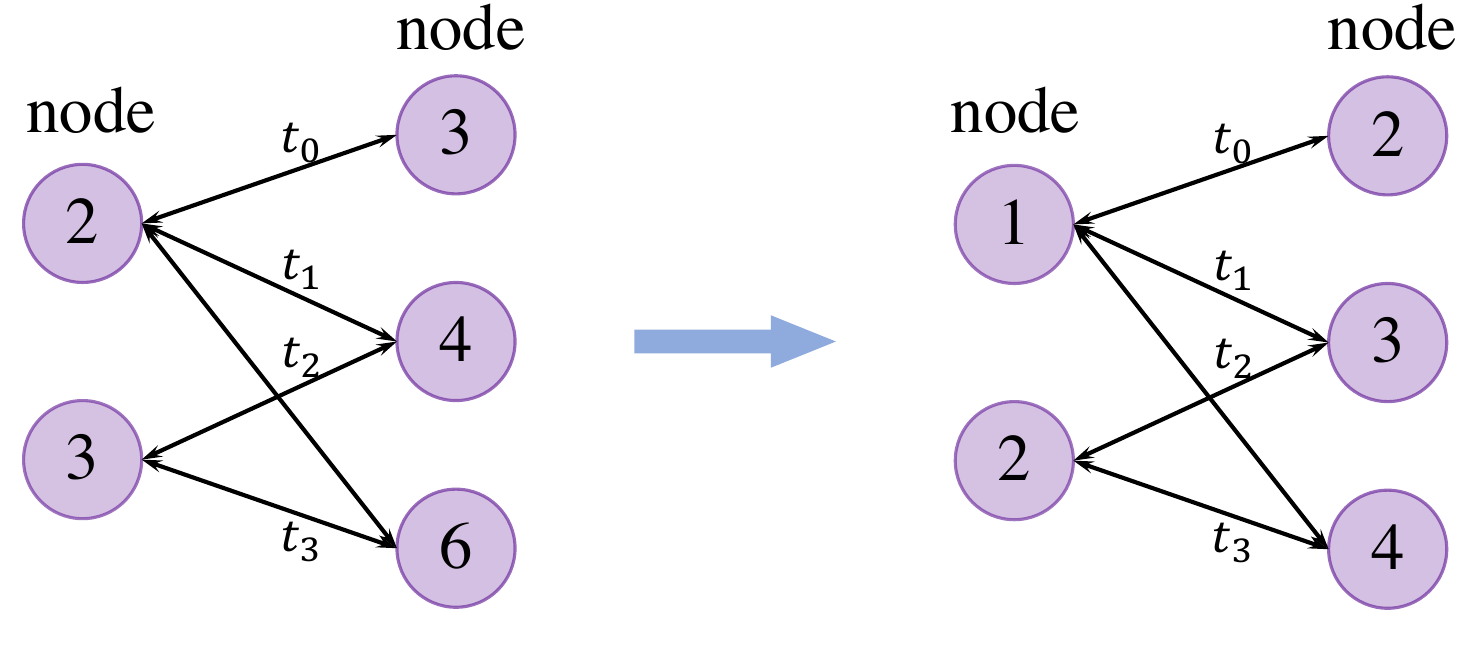}
			\caption{\footnotesize Homogeneous graph}
			\label{non-bipartite}
		\end{subfigure}
		\vspace{-0.5em}
		\caption{\footnotesize Node reindexing of temporal graphs. 
			Node colors represent node types.}
		\label{factorization}
	\end{figure}

	\textbf{Node Reindexing.}
	In some of the original graphs, the node indices are not continuous, and
	the maximal node index is larger than the number of nodes. 
	
	
	As shown in Figure~\ref{bipartite}, for a heterogeneous temporal graph, we propose to reindex the user and item, respectively.
	The user indices are mapped to a continuous range starting from 1.
	Then, the item indices are reindexed in the same manner, starting from the maximal user index.
	As for a homogeneous temporal graph, 
	the user and item nodes belong to the same type.
	Thus, \sys concatenates the original user and item indices, and reindex them together, as shown in Figure \ref{non-bipartite}.
	For the Taobao dataset, the size of node feature matrix can be reduced from $5,162,993\times{172}$ to $82,566\times{172}$, a reduction of 62.53$\times$.


	\subsection{\sys Pipeline}
	\label{sec:pipeline}

	\sys presents a general pipeline for TGNN and standardizes the entire lifecycle.
	There are seven major modules in \sys --- {\em Dataset}, {\em DataLoader}, {\em EdgeSampler}, {\em Evaluator}, {\em Model}, {\em EarlyStopMonitor}, and {\em Leaderboard}.
	Users can also implement their own modules using our library.

	

	\subsubsection{Link Prediction}
	\label{sec:pipeline_lp}

	\begin{figure}[!t]
		\centering
		\includegraphics[width=0.8\textwidth]{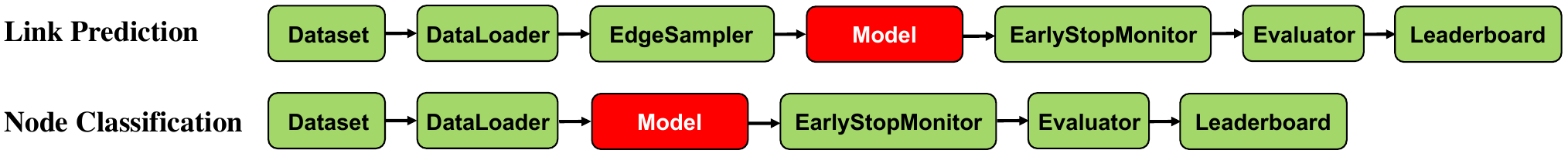}
		\caption{\footnotesize pipelines for link prediction and node classification tasks.}
		\label{fig:lp_pipeline}
	\end{figure}

	Figure~\ref{fig:lp_pipeline} showcases a pipeline for the link prediction task, which consists of all seven modules. 
	
	\begin{itemize}
		
		\item{Dataset.}
		Referring to Section~\ref{subsec:dataset},
		\sys provides fifteen benchmark datasets. 
		The Dataset module is also compatible with user-generated benchmark datasets.
		
		\item{DataLoader.} 
		DataLoader module is responsible for data ingest, preprocessing and splitting.  
		DataLoader standardizes the data splitting under both transductive and inductive settings.
		In transductive link prediction, Dataloader splits the temporal graphs {\em chronologically} into 70\%-15\%-15\% for train, validation and test sets according to edge timestamps. 
		In inductive link prediction, Dataloader performs the same split as the transductive setting, and randomly {\em masks} 10\% nodes as unseen nodes.
		Any edges associated with these unseen nodes are removed from the training set.
		To reflect different inductive scenarios,
		DataLoader further generates three inductive test sets from the transductive test dataset, by {\em filtering edges} in different manners:
		\textbf{Inductive} selects edges with at least one unseen node. 
		\textbf{Inductive New-Old} selects edges between a seen node and an unseen node.
		\textbf{Inductive New-New} selects edges between two unseen nodes. 
		In fact, \textbf{Inductive New-Old} and \textbf{Inductive New-New} together form \textbf{Inductive}; that is,
		$\textbf{Inductive} = \textbf{New-Old} \cup \textbf{New-New}$.

		
		
		
		

		\item{EdgeSampler.} 
		Since link prediction is a self-supervised task, EdgeSampler with a seed is used to randomly sample a set of negative edges.
		
		\item{Model.}
		A TGNN model is created to perform training.
		\sys has implemented seven TGNN models (see Section \ref{sec:models}).
		The users can also implement their own models using our library.
		
		\item{EarlyStopMonitor.} 
		\sys provides a unified EarlyStopMonitor to improve training efficiency and save resources.
		We invoke early stopping with a patience of 3 and $10^{-3}$ tolerance.
		
		\item{Evaluator.}
		Different evaluation metrics are available, including Area Under the Receiver Operating Characteristic Curve (ROC AUC) and Average Precision (AP).
		
		\item{Leaderboard.}
		The results of the pipeline are pushed to the leaderboard, where the users can see the overall performance in the literature.
		
	\end{itemize}
	
	\subsubsection{Node Classification}
	\label{sec:pipeline_nc}
	
	
	The pipeline for node classification consists of six modules, excluding the EdgeSampler.

	\begin{itemize}
		\item{Dataset.} 
		Due to the presence of node labels, only three datasets (Reddit, Wikipedia, and MOOC) are available for node classification task.
		
		\item{DataLoader.} 
		The DataLoader module sorts edges and splits the input dataset (70\%-15\%-15\%) according to edge timestamps.
		
		\item{Model.}
		Some baseline models, such as CAWN, NeurTW, and NAT, have not implemented the node classification task.
		In \sys, we have made a lot of efforts to implement the node classification task for all seven TGNN models.
		
		\item{EarlyStopMonitor, Evaluator \& Leaderboard.}
		These modules are the same as link prediction.
		
	\end{itemize}

	

	\section{Experiments}
	\label{sec:Experiments}
	

	\subsection{Experimental setup}
	\label{subsec:Experimentalsetup}
	
	\textbf{Environment.}
	All experiments were run on an Ubuntu 20.04 server with two 32-core 2.90 GHz Xeon Platinum 8375C CPUs, 256 GB of RAM, and four Nvidia 4090 24 GB GPUs.
	
	\textbf{Datasets.}
	We evaluate fifteen datasets.
	Please refer to Section~\ref{subsec:dataset}, Appendix~\apc{A} and~\apc{B} for details.

	
	
	\textbf{Model Implementations.}
	It is nontrivial to implement the chosen TGNN models since there are inconsistencies between their original implementations. Please refer to Appendix~\apc{C} for details.
	
	\textbf{Tasks.}
	We run both link prediction task and node classification task.
	For link prediction task, we run transductive, inductive, inductive New-Old, and inductive New-New settings (see Section~\ref{sec:pipeline_lp}). 
	For node classification task, we run the traditional transductive setting as introduced in Section~\ref{sec:pipeline_nc}.
	
	\textbf{Evaluation Metrics.} 
	We use the Evaluator module, choosing AUC and AP for the link prediction task, and AUC for the node classification task, following the prior works~\cite{kumar2019predicting,trivedi2019dyrep,rossi2020temporal, xu2020inductive, wanginductive, jin2022neural, luo2022neighborhood}.
	In addition, we report efficiency metrics, as shown in Table~\ref{tab:lp-efficiency}.
	

	\textbf{Protocol.}
	We run each job three times (unless timed out) and report the mean and standard deviation.
	We use an EarlyStopMonitor with a patience of 3 and tolerance of $10^{-3}$, and set a timeout (48 hours).
	We choose binary cross entropy loss and Adam optimizer~\cite{kingma2014adam} to train the models. For Adam, we set learning rate to $10^{-4}$ and use default values for other hyperparameters.


	\begin{table}[!t]
		\tiny
		\setlength{\tabcolsep}{4.5pt}
		\centering
		\caption{\footnotesize ROC AUC results on the link prediction task. 
			"*" denotes that the model encounters runtime error; 
			"---" denotes timeout after 48 hours. 
			The best and second-best results are highlighted as \fv{bold red} and \sv{underlined blue}.
			Some standard deviations are zero because we terminate those models that can only run one epoch within 2 days.
			We do not highlight the second-best if the gap is $>0.05$ compared with the best result.}
		\label{tab:lp-auc}
		\begin{tabular}{l|lllllll}
			\toprule
			&\multicolumn{7}{|c}{\textbf{Transductive}}\\
			\midrule
			\diagbox{Dataset}{Model} &JODIE&DyRep&TGN&TGAT&CAWN&NeurTW&NAT\\
			\midrule
			Reddit & 0.9760 ± 0.0006 & 0.9803 ± 0.0005 & \sv{0.9871 ± 0.0001} & 0.981 ± 0.0002 & \fv{0.9889 ± 0.0002
			} & 0.9841 ± 0.0016 & 0.9854 ± 0.002  \\ 
			Wikipedia & 0.9505 ± 0.0032 & 0.9426 ± 0.0007 & 0.9846 ± 0.0003 & 0.9509 ± 0.0017 & \sv{0.9889 ± 0.0} & \fv{0.9912 ± 0.0001} & 0.9786 ± 0.0035  \\ 
			MOOC & 0.7899 ± 0.0208 & 0.8243 ± 0.0323 & \sv{0.8999 ± 0.0213} & 0.7391 ± 0.0056 & \fv{0.9459 ± 0.0008} & 0.8071 ± 0.0193 & 0.7568 ± 0.0305  \\ 
			LastFM & 0.6766 ± 0.0590 & 0.6793 ± 0.0553 & 0.7743 ± 0.0256 & 0.5094 ± 0.0071 & \fv{0.8746 ± 0.0013} & 0.839 ± 0.0 & \sv{0.8536 ± 0.0027}  \\ 
			Enron & 0.8293 ± 0.0148 & 0.7986 ± 0.0358 & 0.8621 ± 0.0173 & 0.6161 ± 0.0214 & \sv{0.9159 ± 0.0032} & 0.8956 ± 0.0045 & \fv{0.9212 ± 0.0029}  \\ 
			SocialEvo & 0.8666 ± 0.0233 & 0.9020 ± 0.0026 & \fv{0.952 ± 0.0003} & 0.7851 ± 0.0047 & \sv{0.9337 ± 0.0003} & --- & 0.9202 ± 0.0065  \\ 
			UCI & 0.8786 ± 0.0017 & 0.5086 ± 0.0651 & 0.8875 ± 0.0161 & 0.7998 ± 0.0052 & \sv{0.9189 ± 0.0017} & \fv{0.9670 ± 0.0031} & 0.9076 ± 0.0116  \\ 
			CollegeMsg & 0.5730 ± 0.0690 & 0.5382 ± 0.0058 & 0.8419 ± 0.084 & 0.8084 ± 0.0032 & 0.9156 ± 0.004 & \fv{0.9698 ± 0.0} & 0.9059 ± 0.0122  \\ 
			CanParl & 0.7939 ± 0.0063 & 0.7737 ± 0.0255 & 0.7575 ± 0.0694 & 0.7077 ± 0.0218 & 0.7197 ± 0.0905 & \fv{0.8920 ± 0.0173} & 0.6917 ± 0.0722  \\ 
			Contact & 0.9379 ± 0.0073 & 0.9276 ± 0.0206 & \sv{0.9769 ± 0.0032} & 0.5582 ± 0.009 & 0.9685 ± 0.0028 & \fv{0.984 ± 0.0} & 0.9463 ± 0.021  \\ 
			Flights & 0.9449 ± 0.0073 & 0.8981 ± 0.0056 & \sv{0.9787 ± 0.0025} & 0.9016 ± 0.0027 & \fv{0.9861 ± 0.0002} & 0.9302 ± 0.0 & 0.9747 ± 0.0061  \\ 
			UNTrade & 0.6786 ± 0.0103 & 0.6377 ± 0.0032 & 0.6543 ± 0.01 & * & \sv{0.7511 ± 0.0012} & 0.5924 ± 0.0368 & \fv{0.783 ± 0.0472}  \\ 
			USLegis & 0.8278 ± 0.0024 & 0.7425 ± 0.0374 & 0.8137 ± 0.002 & 0.7738 ± 0.0062 & \sv{0.9643 ± 0.0043} & \fv{0.9715 ± 0.0009} & 0.782 ± 0.0261  \\ 
			UNVote & 0.6523 ± 0.0082 & 0.6236 ± 0.0305 & \fv{0.7176 ± 0.0109} & 0.5134 ± 0.0026 & 0.6037 ± 0.0019 & 0.5871 ± 0.0 & \sv{0.6776 ± 0.0411} \\ 
			Taobao & 0.8405 ± 0.0006 & 0.8409 ± 0.0012 & 0.8654 ± 0.0005 & 0.5396 ± 0.009 & 0.7708 ± 0.0026 & \sv{0.8759 ± 0.0009} & \fv{0.8937 ± 0.0015} \\ 
			\midrule
			&\multicolumn{7}{|c}{\textbf{Inductive}}  \\
			\midrule
			Reddit & 0.9514 ± 0.0045 & 0.9583 ± 0.0004 & 0.976 ± 0.0002 & 0.9651 ± 0.0002 & \sv{0.9868 ± 0.0002} & 0.9802 ± 0.0011 & \fv{0.9906 ± 0.0034}  \\ 
			Wikipedia & 0.9305 ± 0.0020 & 0.9099 ± 0.0031 & 0.9781 ± 0.0005 & 0.9343 ± 0.0031 & 0.989 ± 0.0003 & \sv{0.9904 ± 0.0002} & \fv{0.9962 ± 0.0026}  \\ 
			MOOC & 0.7779 ± 0.0575 & 0.8269 ± 0.0182 & 0.8869 ± 0.0249 & 0.737 ± 0.006 & \fv{0.9481 ± 0.0006} & 0.8045 ± 0.0224 & 0.7325 ± 0.0433  \\ 
			LastFM & 0.8011 ± 0.0344 & 0.7990 ± 0.0444 & 0.8284 ± 0.0142 & 0.5196 ± 0.0144 & \sv{0.9082 ± 0.0015} & 0.884 ± 0.0 & \fv{0.9139 ± 0.0038}  \\ 
			Enron & 0.8038 ± 0.0223 & 0.7120 ± 0.0600 & 0.8159 ± 0.0231 & 0.5529 ± 0.015 & \sv{0.9162 ± 0.0016} & 0.9051 ± 0.0027 & \fv{0.952 ± 0.0057}  \\ 
			SocialEvo & 0.8963 ± 0.0228 & 0.9158 ± 0.0039 & \sv{0.9244 ± 0.0084} & 0.6748 ± 0.0021 & \fv{0.9298 ± 0.0002} & --- & 0.896 ± 0.0164  \\ 
			UCI & 0.7517 ± 0.0059 & 0.4297 ± 0.0428 & 0.8083 ± 0.0237 & 0.7024 ± 0.0048 & 0.9177 ± 0.0015 & \fv{0.9686 ± 0.0031} & \sv{0.9622 ± 0.0167}  \\ 
			CollegeMsg & 0.5097 ± 0.0306 & 0.4838 ± 0.0116 & 0.777 ± 0.0522 & 0.715 ± 0.0007 & 0.9163 ± 0.0038 & \fv{0.9726 ± 0.0002} & \sv{0.9603 ± 0.0173}  \\ 
			CanParl & 0.5012 ± 0.0155 & 0.5532 ± 0.0088 & 0.5727 ± 0.0268 & 0.5802 ± 0.0069 & 0.7154 ± 0.0967 & \fv{0.8871 ± 0.0139} & 0.6214 ± 0.0734  \\ 
			Contact & 0.9358 ± 0.0025 & 0.8650 ± 0.0431 & 0.952 ± 0.0056 & 0.5571 ± 0.0047 & \sv{0.9691 ± 0.0031} & \fv{0.9842 ± 0.0} & 0.9466 ± 0.0127  \\ 
			Flights & 0.9218 ± 0.0094 & 0.8689 ± 0.0128 & 0.9519 ± 0.0043 & 0.8321 ± 0.0041 & \fv{0.9834 ± 0.0001} & 0.9158 ± 0.0 & \sv{0.9827 ± 0.003}  \\ 
			UNTrade & 0.6727 ± 0.0132 & 0.6467 ± 0.0112 & 0.5977 ± 0.014 & * & \fv{0.7398 ± 0.0007} & 0.5915 ± 0.0328 & 0.6475 ± 0.0664  \\ 
			USLegis & 0.5840 ± 0.0129 & 0.5980 ± 0.0097 & 0.6128 ± 0.0046 & 0.5568 ± 0.0078 & \sv{0.9665 ± 0.0032} & \fv{0.9708 ± 0.0009} & 0.7453 ± 0.0286  \\ 
			UNVote & 0.5121 ± 0.0005 & 0.4993 ± 0.0103 & 0.5881 ± 0.0118 & 0.477 ± 0.0047 & 0.5911 ± 0.0006 & 0.586 ± 0.0 & \fv{0.779 ± 0.0082} \\
			Taobao & 0.701 ± 0.0013 & 0.7026 ± 0.0006 & 0.7017 ± 0.0026 & 0.5261 ± 0.0119 & 0.7737 ± 0.0027 & 0.8843 ± 0.0016 & \fv{0.9992 ± 0.0002} \\ 
			\midrule
			&\multicolumn{7}{|c}{\textbf{Inductive New-Old}}  \\
			\midrule
			Reddit & 0.9488 ± 0.0043 & 0.9549 ± 0.0029 & 0.9742 ± 0.0004 & 0.9639 ± 0.0004 & \sv{0.9848 ± 0.0002} & 0.9789 ± 0.0017 & \fv{0.9949 ± 0.0017}  \\ 
			Wikipedia & 0.9084 ± 0.0043 & 0.8821 ± 0.0031 & 0.9703 ± 0.0008 & 0.9178 ± 0.0023 & \sv{0.9886 ± 0.0002} & 0.9878 ± 0.0002 & \fv{0.9963 ± 0.0021}  \\ 
			MOOC & 0.7910 ± 0.0475 & 0.8274 ± 0.0132 & 0.8808 ± 0.0326 & 0.7438 ± 0.0063 & \fv{0.949 ± 0.0016} & 0.8052 ± 0.0244 & 0.7487 ± 0.0459  \\ 
			LastFM & 0.7305 ± 0.0051 & 0.6980 ± 0.0364 & 0.763 ± 0.0231 & 0.5189 ± 0.003 & \sv{0.8678 ± 0.0030} & 0.8311 ± 0.0 & \fv{0.9144 ± 0.0013}  \\ 
			Enron & 0.7859 ± 0.0134 & 0.6915 ± 0.0650 & 0.8100 ± 0.0204 & 0.5589 ± 0.0235 & \sv{0.9185 ± 0.003} & 0.9007 ± 0.0039 & \fv{0.9491 ± 0.0079}  \\ 
			SocialEvo & 0.8953 ± 0.0303 & \sv{0.9182 ± 0.0050} & \fv{0.9257 ± 0.0086} & 0.684 ± 0.0034 & 0.9155 ± 0.0002 & --- & 0.8793 ± 0.0318  \\ 
			UCI & 0.7139 ± 0.0112 & 0.4259 ± 0.0397 & 0.8015 ± 0.0269 & 0.6842 ± 0.0078 & 0.9176 ± 0.0028 & \sv{0.9696 ± 0.0039} & \fv{0.9748 ± 0.0163}  \\ 
			CollegeMsg & 0.5168 ± 0.0360 & 0.4808 ± 0.0279 & 0.7725 ± 0.0365 & 0.7012 ± 0.005 & 0.9166 ± 0.0032 & \sv{0.968 ± 0.0018} & \fv{0.9725 ± 0.0189}  \\ 
			CanParl & 0.5078 ± 0.0005 & 0.5393 ± 0.0204 & 0.5691 ± 0.0223 & 0.5724 ± 0.0063 &  0.7231 ± 0.085 & \fv{0.8847 ± 0.0102} & 0.6277 ± 0.0811  \\ 
			Contact & 0.9345 ± 0.0027 & 0.8574 ± 0.0454 & 0.9527 ± 0.0052 & 0.556 ± 0.0039 & \sv{0.9691 ± 0.0028} & \fv{0.9841 ± 0.0} & 0.9351 ± 0.0202  \\ 
			Flights & 0.9172 ± 0.0114 & 0.8650 ± 0.0126 & 0.9503 ± 0.0043 & 0.8285 ± 0.0038 & \sv{0.9828 ± 0.0002} & 0.9127 ± 0.0 & \fv{0.986 ± 0.0034}  \\ 
			UNTrade & 0.6650 ± 0.0106 & 0.6306 ± 0.0139 & 0.5959 ± 0.0171 & * & \fv{0.7413 ± 0.001} & 0.5965 ± 0.0371 & 0.5812 ± 0.0957  \\ 
			USLegis & 0.5801 ± 0.0213 & 0.5673 ± 0.0098 & 0.5741 ± 0.0148 & 0.5596 ± 0.0092 & \sv{0.9672 ± 0.0029} & \fv{0.9682 ± 0.0018} & 0.531 ± 0.1  \\ 
			UNVote & 0.5208 ± 0.0075 & 0.5023 ± 0.0204 & 0.5889 ± 0.0106 & 0.4787 ± 0.0033 & 0.5933 ± 0.0007 & 0.5878 ± 0.0 & \fv{0.7789 ± 0.0192} \\
			Taobao & 0.6988 ± 0.0024 & 0.6992 ± 0.0001 & 0.7025 ± 0.0038 & 0.5266 ± 0.0239 & 0.7574 ± 0.0032 & 0.8617 ± 0.0032 & \fv{0.9997 ± 0.0001} \\
			\midrule
			&\multicolumn{7}{|c}{\textbf{Inductive New-New}}  \\
			\midrule
			Reddit & 0.9381 ± 0.0090 & 0.9525 ± 0.0050 & 0.9811 ± 0.0004 & 0.9597 ± 0.0043 & \sv{0.9952 ± 0.0016} & 0.9875 ± 0.0004 & \fv{0.9954 ± 0.0011}  \\ 
			Wikipedia & 0.9349 ± 0.0051 & 0.9261 ± 0.0030 & 0.9858 ± 0.0007 & 0.958 ± 0.0039 & 0.9934 ± 0.0005 & \sv{0.996 ± 0.0001} & \fv{0.9984 ± 0.0008}  \\ 
			MOOC & 0.7065 ± 0.0165 & 0.7217 ± 0.0178 & 0.8762 ± 0.0038 & 0.7403 ± 0.0057 & \fv{0.9422 ± 0.0003} & 0.8048 ± 0.0087 & 0.6562 ± 0.0287  \\ 
			LastFM & 0.8852 ± 0.0090 & 0.8683 ± 0.0160 & 0.8754 ± 0.0011 & 0.5092 ± 0.0333 & \sv{0.9697 ± 0.0002} & 0.9628 ± 0.0 & \fv{0.9743 ± 0.0015}  \\ 
			Enron & 0.6800 ± 0.0017 & 0.6571 ± 0.0521 & 0.7644 ± 0.0179 & 0.531 ± 0.0179 & \sv{0.9609 ± 0.0051} & 0.9387 ± 0.0001 & \fv{0.9687 ± 0.0051}  \\ 
			SocialEvo & 0.6484 ± 0.0491 & 0.7740 ± 0.0215 & 0.8791 ± 0.0045 & 0.4659 ± 0.0068 & \fv{0.9318 ± 0.0003} & --- & \sv{0.9275 ± 0.0471}  \\ 
			UCI & 0.6393 ± 0.0158 & 0.4771 ± 0.0100 & 0.8051 ± 0.021 & 0.768 ± 0.0041 & 0.9245 ± 0.0027 & \fv{0.9716 ± 0.0016} & \sv{0.9472 ± 0.0262}  \\ 
			CollegeMsg & 0.5320 ± 0.0269 & 0.5269 ± 0.0049 & 0.7969 ± 0.0111 & 0.7832 ± 0.0026 & 0.9304 ± 0.0024 & \fv{0.9762 ± 0.0008} & \sv{0.9404 ± 0.0371}  \\ 
			CanParl & 0.4347 ± 0.0090 & 0.4430 ± 0.0068 & 0.5625 ± 0.0396 & 0.5955 ± 0.0074 & 0.7005 ± 0.1241 & \fv{0.8882 ± 0.0045} & 0.5685 ± 0.0326  \\ 
			Contact & 0.7531 ± 0.0059 & 0.6602 ± 0.0395 & 0.9118 ± 0.0053 & 0.5449 ± 0.0056 & \sv{0.9652 ± 0.0012} & \fv{0.982 ± 0.0} & 0.9495 ± 0.0034  \\ 
			Flights & 0.9303 ± 0.0083 & 0.8900 ± 0.0266 & 0.9652 ± 0.0022 & 0.857 ± 0.0056 & \sv{0.9873 ± 0.0009} & 0.9411 ± 0.0 & \fv{0.9905 ± 0.0014}  \\ 
			UNTrade & 0.5922 ± 0.0085 & 0.5362 ± 0.0147 & 0.5068 ± 0.0061 & * & \fv{0.7458 ± 0.0081} & 0.5938 ± 0.0600 & 0.6876 ± 0.0177  \\ 
			USLegis & 0.5390 ± 0.0075 & 0.5640 ± 0.0192 & 0.5626 ± 0.0195 & 0.5324 ± 0.0294 & \sv{0.9738 ± 0.0058} & \fv{0.9787 ± 0.0004} & 0.8897 ± 0.0225  \\ 
			UNVote & 0.4913 ± 0.0203 & 0.4728 ± 0.0033 & 0.5663 ± 0.0093 & 0.5 ± 0.006 & 0.5775 ± 0.0022 & 0.5669 ± 0.0 & \fv{0.7198 ± 0.0748} \\ 
			Taobao & 0.7169 ± 0.0013 & 0.717 ± 0.0011 & 0.7082 ± 0.001 & 0.5226 ± 0.0054 & 0.7847 ± 0.0148 & 0.9083 ± 0.0013 &  \fv{0.9996 ± 0.0001} \\ 
			\bottomrule
		\end{tabular}
	\end{table}

	\subsection{Link Prediction Task}
	\label{subsec:lp-performance}
	
	\textbf{Model Performance.}
	We run the link prediction task on 7 TGNN models and 15 datasets under different settings.
	In Table~\ref{tab:lp-auc}, we show the AUC results and highlight the best and second-best numbers for each job.
	Please refer to Appendix~\apc{D.1} for the results of average precision (AP).
	The overall performance is similar to that of AUC.
	
	For the \textbf{transductive} setting, we find that no model can perform the best consistently across different datasets.
	This is not surprising considering the dynamic nature and diverse structures of temporal graphs.
	In a nutshell, CAWN obtains the best or second-best results on 10 datasets out of 15,
	followed by NeurTW (7 out of 15), NAT (5 out of 15), and TGN (6 out of 15).
	The outstanding performance of CAWN and NeurTW comes from their temporal walk mechanism that perceives structural information.
	This also proves that introducing graph topology in TGNN models can effectively improve the model quality.
	Specially, NeurTW significantly outperforms other baselines on the CanParl dataset.
	NeurTW introduces a continuous-time operation that can depict evolution trajectory, which 
	is potentially suitable for CanParl with a large time granularity (per year)~\cite{jin2022neural,poursafaeitowards}.
	The source code of TGAT reports error on the UNTrade dataset because it may not find suitable neighbors within some given time intervals.
	This error also occurs using other implementations~\cite{poursafaeitowards}.

	When it comes to the \textbf{inductive} setting,
	we observe different results.
	CAWN ranks top-2 on 9 datasets, followed by NeurTW on 6 and NAT on 9.
	These methods reveal better generalization capability under the inductive setting since they consider local structures, either by temporal walk or joint neighborhood.
	In contrast, methods adopting attention and RNN (e.g., TGAT and TGN) suffer performance degradation when the features of test data are not used during the training.
	NeurTW exceeds other models on CanParl, UNTrade, and UNVote datasets.
	As we have discussed, NeurTW works well on these large-granularity datasets due to its continuous-time encoder.
	For the New-Old and New-New scenarios, we observe similar results.
	Specifically, NAT achieves top-2 performance on 10 datasets under the New-New scenario, verifying that 
	learning joint neighborhood benefits node representations.
	CAWN, NeurTW, and NAT still perform well under the inductive \textbf{New-New} setting due to their structure-aware techniques. CAWN and NeurTW are both based on motifs and index anonymization operation \cite{wanginductive, jin2022neural}. NeurTW additionally constructs  neural ordinary
	differential equations (NODEs). NAT relies on joint neighborhood features based on  a dedicated
	data structure termed \textit{N-caches}~\cite{luo2022neighborhood}.

	
	
	\begin{table}[!t]
		\tiny
		\setlength{\tabcolsep}{2pt}
		\centering
		\caption{\footnotesize Model efficiency on the link prediction task. 
			We report seconds per epoch as \textbf{Runtime}, 
			the averaged number of epochs for convergence before early stopping as \textbf{Epoch}, the maximum RAM usage as \textbf{RAM}, and the maximum GPU memory usage  as \textbf{GPU Memory}, respectively.
			Refer to Table~\ref{tab:lp-auc} for the meaning of "*" and "---".
			"x" indicates that the model cannot converge within 48 hours.}
		\label{tab:lp-efficiency}
		\begin{tabular}{l|rrrrrrr|rrrrrrr}
			\toprule
			&\multicolumn{7}{|c|}{\textbf{Runtime} (second)}&\multicolumn{7}{|c}{\textbf{Epoch}}\\
			\midrule
			\diagbox{Dataset}{Model} &JODIE&DyRep&TGN&TGAT&CAWN&NeurTW&NAT&JODIE&DyRep&TGN&TGAT&CAWN&NeurTW&NAT\\
			\midrule
			Reddit & 140.20  & 134.44  & \sv{61.63}  & 515.64  & 1,701.72  & 27,108.06  & \fv{27.05}  & 19 & 20 & 21 & \fv{9} & x & x & \sv{12}  \\ 
			Wikipedia & \sv{9.37}  & 10.02  & 12.17  & 109.24  & 270.83  & 4,388.59  & \fv{6.15}  & 30 & 32 & 34 & 13 & \sv{8} & 12 & \fv{6}  \\ 
			MOOC & \sv{30.19}  & 33.54  & 41.48  & 256.26  & 1,913.38  & 13,497.27  & \fv{16.48}  & 9 & 18 & 17 & \sv{5} & 10 & x & \fv{4}  \\ 
			LastFM & \fv{29.42}  & \sv{39.61}  & 45.98  & 882.36  & 5,527.12  & 51,007.30  & 51.95  & 9 & 12 & 14 & \sv{6} & x & x & \fv{4}  \\ 
			Enron & \fv{2.41}  & \sv{3.45}  & 4.13  & 92.24  & 398.38  & 10,896.81  & 5.70  & 15 & 15 & 19 & 9 & \sv{8} & x & \fv{6}  \\ 
			SocialEvo & \fv{27.77}  & \sv{42.57}  & 51.35  & 1,544.52  & 12,292.31  & --- & 84.72  & 23 & 37 & 14 & 24 & \sv{10} & x & \fv{7}  \\ 
			UCI & \fv{1.89}  & \sv{2.49}  & 2.72  & 57.49  & 121.41  & 3,801.79  & 2.90  & 15 & \fv{6} & 20 & \sv{7} & 11 & 18 & 10  \\ 
			CollegeMsg & \fv{1.87}  & \sv{2.42}  & 2.85  & 45.80  & 111.96  & 2,097.39  & 2.83  & \sv{7} & \fv{6} & 22 & \sv{7} & 12 & 24 & 10  \\ 
			CanParl & \fv{1.96}  & \sv{2.44}  & 3.03  & 46.28  & 234.65  & 4,566.24  & 3.79  & \fv{6} & \sv{7} & \sv{7} & \sv{7} & 13 & x & \fv{6}  \\ 
			Contact & \fv{41.99}  & \sv{58.20}  & 69.60  & 1,645.10  & 12,100.94  & 114,274.38  & 96.35  & 12 & 11 & 14 & \sv{8} & x & x & \fv{4}  \\ 
			Flights & \sv{180.80}  & 197.61  & 262.51  & 1,195.30  & 12,105.70  & 143,731.99  & \fv{76.91}  & 12 & 8 & 12 & \sv{7} & x & x & \fv{5}  \\ 
			UNTrade & \fv{7.89}  & \sv{11.75}  & 14.17  & * & 1,860.84  & 39,402.43  & 21.57  & 15 & 9 & \sv{6} & * & \fv{3} & x & 18  \\ 
			USLegis & \fv{2.15}  & \sv{2.73}  & 2.93  & 36.69  & 220.41  & 3,208.23  & 3.16  & 13 & 12 & 15 & \sv{8} & 17 & \fv{5} & 12  \\ 
			UNVote & \fv{22.31}  & \sv{25.53}  & 28.97  & 686.15  & 6,414.90  &  88,939.57  & 40.51  & 26 & \sv{9} & 18 & \fv{4} & x & x & \sv{9} \\
			Taobao & 32.17  & 38.03  & 34.51  & \sv{29.15}  & 135.04  & 1,156.96  & \fv{4.12}  & 10 & 9 & 9 & 7 & 9 & \sv{6} & \fv{4} \\ 
			\midrule
			&\multicolumn{7}{|c|}{\textbf{RAM} (GB)}&\multicolumn{7}{|c}{\textbf{GPU Memory} (GB)}\\
			\midrule
			Reddit & \sv{3.4} & \sv{3.4} & \sv{3.4} & 3.6 & 30.8 & 10.1 & \fv{3.3} & \sv{2.8} & \sv{2.8} & 3.0 & 4.4 & 3.5 & \sv{2.8} & \fv{2.2}  \\ 
			Wikipedia & \sv{2.7} & \sv{2.7} & \fv{2.6} & \sv{2.7} & 14.4 & 9.8 & \fv{2.6} & \sv{1.8} & \sv{1.8} & 1.9 & 3.6 & 3 & 2.1 & \fv{1.3}  \\ 
			MOOC & \sv{2.6} & \sv{2.6} & \sv{2.6} & \sv{2.6} & 11.5 & 9.1 & \fv{2.4} & 2.1 & 2.1 & 2.1 & 2.8 & \sv{1.5} & 1.6 & \fv{1.3}  \\ 
			LastFM & \sv{2.6} & \sv{2.6} & \sv{2.6} & 3.1 & 20.5 & 9.8 & \fv{2.5} & \sv{1.3} & 1.4 & 1.4 & 2.6 & 1.5 & 1.6 & \fv{1.1}  \\ 
			Enron & \fv{2.4} & \fv{2.4} & \fv{2.4} & \sv{2.5} & 15.8 & 8.1 & \fv{2.4} & \fv{1.3} & \sv{1.4} & \sv{1.4} & 3.0 & 2.3 & 1.7 & \fv{1.3}  \\ 
			SocialEvo & \sv{2.7} & \sv{2.7} & \sv{2.7} & 3.6 & 13.4 & 6.4 & \fv{2.6} & \fv{1.1} & \sv{1.2} & \sv{1.2} & 2.6 & 2.0 & 1.5 & \sv{1.2}  \\ 
			UCI & \sv{2.5} & \fv{2.4} & \fv{2.4} & \sv{2.5} & 11.2 & 15.2 & \fv{2.4} & \sv{1.5} & 1.6 & 1.6 & 3.2 & 2 & 1.9 & \fv{1.4}  \\ 
			CollegeMsg & \fv{2.5} & \fv{2.5} & \fv{2.5} & \fv{2.5} & \sv{11} & 14.8 & \fv{2.5} & \sv{1.4} & 1.7 & 1.5 & 3.5 & 2.5 & 1.8 & \fv{1.3}  \\ 
			CanParl & \fv{2.4} & \fv{2.4} & \fv{2.4} & \fv{2.4} & \sv{8.5} & 8.8 & \fv{2.4} & \fv{1.3} & \fv{1.3} & \sv{1.4} & 2.8 & 1.7 & 1.6 & \fv{1.3}  \\ 
			Contact & \sv{2.8} & \sv{2.8} & \sv{2.8} & 3.7 & 13.2 & 10.1 & \fv{2.6} & \sv{1.3} & \sv{1.3} & 1.4 & 2.7 & 2.1 & 1.6 & \fv{1.1}  \\ 
			Flights & \sv{2.8} & \sv{2.8} & \sv{2.8} & 3.5 & 9.9 & 10.2 & \fv{2.6} & 2.2 & 2.6 & 2.9 & 2.8 & 2.3 & \sv{1.7} & \fv{1.3}  \\ 
			UNTrade & \sv{2.5} & \sv{2.5} & \sv{2.5} & * & 5.4 & 8.7 & \fv{2.4} & \fv{1.1} & \sv{1.2} & \sv{1.2} & * & 2.1 & 1.6 & \fv{1.1}  \\ 
			USLegis & \fv{2.4} & \fv{2.4} & \fv{2.4} & \fv{2.4} & 8.8 & \sv{5.9} & \fv{2.4} & \sv{1.4} & \sv{1.4} & 1.5 & 2.8 & 2.4 & 1.6 & \fv{1.3}  \\ 
			UNVote & \fv{2.5} & \fv{2.5} & \fv{2.5} & \sv{3} & 17.9 & 6.4 & \fv{2.5} & \fv{1.1} & \sv{1.2} & \sv{1.2} & 2.6 & 2.0 & 1.5 & \fv{1.1}  \\ 
			Taobao & 3 & 2.9 & 3 & \fv{2.6} & 3.5 & \sv{2.8} & \fv{2.6} & 7.5 & 7.7 & 8.4 & 2.9 & 2.7 & \sv{1.6} & \fv{1.4} \\
			\bottomrule
		\end{tabular}
		
	\end{table}

	\textbf{Model Efficiency.}
	Since many real-world graphs are extremely large, we believe efficiency is a vital issue for TGNNs in practice.
	We thereby compare the efficiency of the evaluated models, and
	present the results in Table~\ref{tab:lp-efficiency}.
	In terms of runtime per epoch, JODIE, DyRep, TGN and TGAT are faster, as their message passing and memory updater operators are efficient.
	In contrast, CAWN and NeurTW can obtain more accurate models, but are much slower (sometimes encountering timeouts) due to their inefficient temporal walk operations.
	NAT is relatively faster than temporal walk-based methods through caching and parallelism optimizations, {\em achieving a good trade-off between model quality and efficiency}.
	Regarding the consumed epochs till convergence,
	TGAT and NAT converge relatively faster, while CAWN and NeurTW often cannot converge in a reasonable time.
	\textbf{RAM} results illustrate that CAWN and NeurTW consume more memory than other models due to the complex sampler.
	As for \textbf{GPU Memory}, most models consume 1GB - 3GB. 
	TGAT requires the highest GPU memory on almost all datasets, since it stacks multiple attention layers.
	The memory usages of JODIE, DyRep, and TGN on Taobao are much higher than those of other models.
	Taobao has the maximal nodes, so that
	the \textit{Memory} module consumes much larger GPU memory. 
	We also evaluate GPU utilization (please refer to Appendix~\apc{D.2}) and the results are similar.

	\subsection{Node Classification Task}
	
	\textbf{Model Performance.}
	We evaluate Reddit, Wikipedia, and MOOC datasets since they have two classes of node labels.
	The test AUC results for the node classification task are shown in Table \ref{tab:nc-auc}.
	In these three datasets, the node labels are highly imbalanced and may change over time, yielding relatively lower AUC than the link prediction task. 
	TGN achieves the best result on Wikipedia dataset, verifying the effectiveness of temporal memory on node classification task.
	TGAT obtains the best or second-best results on two datasets.
	TGAT adopts temporal attention to explicitly capture the evolution of node embeddings, and therefore can potentially learn better knowledge of nodes.
	CAWN and NeurTW perform better on MOOC, since MOOC is relatively denser and
	the temporal walk mechanism can effectively perceive local structures.
	NAT performs poorly on the node classification task. 
	The node classification task does not rely on structural features as much as the link prediction task, so that the joint neighborhood mechanism may be less effective.

	
	\begin{table}[!t]
		\tiny
		\setlength{\tabcolsep}{4pt}
		\caption{\footnotesize ROC AUC results for the node classification task. 
			The top-2 results are highlighted as \fv{bold red} and \sv{underlined blue}.}
		\label{tab:nc-auc}
		\centering
		\begin{tabular}{l|lllllll}
			\toprule
			\diagbox{Dataset}{Model} &JODIE&DyRep&TGN&TGAT&CAWN&NeurTW&NAT\\
			\midrule
			Reddit  & 0.6033 ± 0.0173 & 0.4988 ± 0.0066 & 0.6216 ± 0.007 & \sv{0.6252 ± 0.0057} & \fv{0.6502 ± 0.0252} & 0.6054 ± 0.0352 & 0.4746 ± 0.0207  \\ \midrule
			Wikipedia  & 0.8527 ± 0.002 & 0.8276 ± 0.008 & \fv{0.8831 ± 0.0009} & \sv{0.8603 ± 0.0051} & 0.8586 ± 0.0030 & 0.8470 ± 0.0233 & 0.5417 ± 0.0474  \\ \midrule
			MOOC & 0.6774 ± 0.0066 & 0.6604 ± 0.0037 & 0.626 ± 0.0062 & 0.6705 ± 0.005 & \sv{0.7271 ± 0.0016} & \fv{0.7719 ± 0.0073} & 0.5175 ± 0.0143 \\ 
			\bottomrule
		\end{tabular}
	\end{table}

	\textbf{Model Efficiency.}
	Due to the space constraint, we report the efficiency performance of node classification task in Appendix~\apc{D.3}.
	Overall, we observe similar results as the link prediction task.

	\subsection{Discussion and Visions}
	
	\textbf{Link Prediction.} 
	\underline{Temporal walk}-based models retrieve a set of motifs to learn \underline{structural information}, and therefore {\em benefit future edge prediction}. 
	NAT~\cite{luo2022neighborhood} achieves a {\em trade-off} between statistical performance and efficiency, owing to its \underline{joint-neighborhood} learning paradigm.
	CAWN~\cite{wanginductive} and NeurTW~\cite{jin2022neural} perform well on the link prediction task and are both based on motifs and index anonymization operation. However, NeurTW~\cite{jin2022neural} additionally constructs  neural ordinary
	differential equations (NODEs)  for embedding extracted motifs over time intervals to capture dynamics.

	\textbf{Node Classification.}
	TGN, TGAT, CAWN and NeurTW obtain top-2 results on at least one dataset, verifying that
	node memory, graph attention, and temporal walk strategies are {\em effective for node classification}.
	These three techniques perform \underline{message passing} for each node using all of its neighborhoods, and can learn better node representations. 
	
	\textbf{Limitation and Future Direction.} It is hard for traditional GNNs to capture the dynamic evolution of temporal graphs. Motifs suffer the complex random walk and sampler. Joint-neighborhoods overly focused on joint neighborhoods and performs poorly on node classification task. Inspired by the above analysis, we can conclude that the future directions of TGNN models are more focused on mining the temporal structure of the temporal graph and increasing the model's structure-aware ability by jointing motifs~\cite{wanginductive, jin2022neural} and joint-neighborhood~\cite{luo2022neighborhood}.
	
	\textbf{Our Improvement.}
	Here, we introduce TeMP in the Appendix, a novel approach that incorporates GNN aggregation and temporal structure.
	According to our empirical results, TeMP effectively balances model performance and efficiency, and performs well on both link prediction and node classification tasks.
	Please refer to \href{https://openreview.net/attachment?id=rnZm2vQq31&name=supplementary_material}{Appendix}~\apc{E} for details.

	
	\section{Related work}
	
	
	Early works \cite{xu2019generative,pareja2020evolvegcn,hajiramezanali2019variational} treat the temporal graph as a sequence of snapshots, then encode the snapshots utilizing static GNNs, and capture temporal evolution patterns through sequence analysis methods. 
	
	Recent works~\cite{kumar2019predicting,trivedi2019dyrep,rossi2020temporal,xu2020inductive,wanginductive,jin2022neural, luo2022neighborhood, nguyen2018continuous, tian2021streaming,zhou2022tgl, souza2022provably,liu2022neural}
	tend to choose a continuous-time abstraction, treat the input as
	interaction streams, thus preserving dynamic evolution in temporal graphs.
	Some models update the representation using the node that is currently interacting~\cite{rossi2020temporal};
	while others update the representation using the historical neighbor nodes~\cite{trivedi2019dyrep,xu2020inductive}, and encode the temporal structural pattern.
	In addition, a series of motif-based models ~\cite{wanginductive,jin2022neural} aim at capturing high-order connectivity.
	These models first sample temporal random walks starting from the target node and obtain motifs which reflect the spatio-temporal information around the target node.
	
	However, to our best knowledge, there is no benchmark that conducts a standardized comparison for TGNN models.
	A few works~\cite{poursafaeitowards} have made preliminary attempts, but they have not provided standard datasets and unified evaluation pipelines.
	In this work, we try to build a comprehensive benchmark framework for TGNNs and standardize the entire lifecycle.

	\section{Conclusion}
	\label{sec:conclusion}
	
	In this work, we propose \sys, a general benchmark for temporal graph neural networks.
	In the \sys, we construct a series of benchmark datasets and design a unified pipeline. 
	We compare the effectiveness and efficiency of different TGNN models fairly across different tasks, settings, and metrics.
	Through an in-depth analysis of the empirical results, we draw some valuable insights for future research.
	In future work, we will continuously update our benchmark, corresponding to newly emerged TGNN models.
	Contributions and issues from the community are eagerly welcomed, with which
	we can together push forward the TGNN research.

	
	\bibliographystyle{unsrtnat}
	\bibliography{reference}

\begin{thebibliography}{34}
\providecommand{\natexlab}[1]{#1}
\providecommand{\url}[1]{\texttt{#1}}
\expandafter\ifx\csname urlstyle\endcsname\relax
  \providecommand{\doi}[1]{doi: #1}\else
  \providecommand{\doi}{doi: \begingroup \urlstyle{rm}\Url}\fi

\bibitem[Kumar et~al.(2019)Kumar, Zhang, and Leskovec]{kumar2019predicting}
Srijan Kumar, Xikun Zhang, and Jure Leskovec.
\newblock Predicting dynamic embedding trajectory in temporal interaction
  networks.
\newblock In \emph{Proceedings of the 25th ACM SIGKDD International Conference
  on Knowledge Discovery and Data Mining}, pages 1269--1278, 2019.

\bibitem[Trivedi et~al.(2019)Trivedi, Farajtabar, Biswal, and
  Zha]{trivedi2019dyrep}
Rakshit Trivedi, Mehrdad Farajtabar, Prasenjeet Biswal, and Hongyuan Zha.
\newblock Dyrep: Learning representations over dynamic graphs.
\newblock In \emph{International Conference on Learning Representations}, 2019.

\bibitem[Rossi et~al.(2020)Rossi, Chamberlain, Frasca, Eynard, Monti, and
  Bronstein]{rossi2020temporal}
Emanuele Rossi, Ben Chamberlain, Fabrizio Frasca, Davide Eynard, Federico
  Monti, and Michael Bronstein.
\newblock Temporal graph networks for deep learning on dynamic graphs.
\newblock \emph{arXiv preprint arXiv:2006.10637}, 2020.

\bibitem[Xu et~al.(2020)Xu, Ruan, Korpeoglu, Kumar, and Achan]{xu2020inductive}
Da~Xu, Chuanwei Ruan, Evren Korpeoglu, Sushant Kumar, and Kannan Achan.
\newblock Inductive representation learning on temporal graphs.
\newblock In \emph{International Conference on Learning Representations}, 2020.

\bibitem[Wang et~al.(2021)Wang, Chang, Liu, Leskovec, and Li]{wanginductive}
Yanbang Wang, Yen-Yu Chang, Yunyu Liu, Jure Leskovec, and Pan Li.
\newblock Inductive representation learning in temporal networks via causal
  anonymous walks.
\newblock In \emph{International Conference on Learning Representations}, 2021.

\bibitem[Jin et~al.(2022)Jin, Li, and Pan]{jin2022neural}
Ming Jin, Yuan-Fang Li, and Shirui Pan.
\newblock Neural temporal walks: Motif-aware representation learning on
  continuous-time dynamic graphs.
\newblock In \emph{Advances in Neural Information Processing Systems}, 2022.

\bibitem[Luo and Li(2022)]{luo2022neighborhood}
Yuhong Luo and Pan Li.
\newblock Neighborhood-aware scalable temporal network representation learning.
\newblock In \emph{The First Learning on Graphs Conference}, 2022.

\bibitem[Poursafaei et~al.(2022)Poursafaei, Huang, Pelrine, and
  Rabbany]{poursafaeitowards}
Farimah Poursafaei, Andy Huang, Kellin Pelrine, and Reihaneh Rabbany.
\newblock Towards better evaluation for dynamic link prediction.
\newblock In \emph{Advances in Neural Information Processing Systems Datasets
  and Benchmarks Track}, 2022.

\bibitem[data dump()]{url_reddit}
Reddit data dump.
\newblock \url{http://files.pushshift.io/reddit/}.

\bibitem[cup 2015()]{url_mooc}
Kdd cup 2015.
\newblock \url{https://biendata.com/competition/kddcup2015/data/}.

\bibitem[Hidasi and Tikk(2012)]{hidasi2012fast}
Bal{\'a}zs Hidasi and Domonkos Tikk.
\newblock Fast als-based tensor factorization for context-aware recommendation
  from implicit feedback.
\newblock In \emph{The European Conference on Machine Learning and Principles
  and Practice of Knowledge Discovery in Databases (ECML PKDD)}, pages 67--82.
  Springer, 2012.

\bibitem[Zhu et~al.(2018)Zhu, Li, Zhang, Li, He, Li, and Gai]{zhu2018learning}
Han Zhu, Xiang Li, Pengye Zhang, Guozheng Li, Jie He, Han Li, and Kun Gai.
\newblock Learning tree-based deep model for recommender systems.
\newblock In \emph{Proceedings of the 24th ACM SIGKDD International Conference
  on Knowledge Discovery and Data Mining}, pages 1079--1088, 2018.

\bibitem[email dataset()]{url_enron}
Enron email dataset.
\newblock \url{http://www.cs.cmu.edu/~enron/}.

\bibitem[Leskovec and Krevl(2014)]{leskovec2014snap}
Jure Leskovec and Andrej Krevl.
\newblock Snap datasets: Stanford large network dataset collection, 2014.

\bibitem[Madan et~al.(2011)Madan, Cebrian, Moturu, Farrahi,
  et~al.]{madan2011sensing}
Anmol Madan, Manuel Cebrian, Sai Moturu, Katayoun Farrahi, et~al.
\newblock Sensing the "health state" of a community.
\newblock \emph{IEEE Pervasive Computing}, 11\penalty0 (4):\penalty0 36--45,
  2011.

\bibitem[Opsahl and Panzarasa(2009)]{opsahl2009clustering}
Tore Opsahl and Pietro Panzarasa.
\newblock Clustering in weighted networks.
\newblock \emph{Social networks}, 31\penalty0 (2):\penalty0 155--163, 2009.

\bibitem[Panzarasa et~al.(2009)Panzarasa, Opsahl, and
  Carley]{panzarasa2009patterns}
Pietro Panzarasa, Tore Opsahl, and Kathleen~M Carley.
\newblock Patterns and dynamics of users' behavior and interaction: Network
  analysis of an online community.
\newblock \emph{Journal of the American Society for Information Science and
  Technology}, 60\penalty0 (5):\penalty0 911--932, 2009.

\bibitem[Huang et~al.(2020)Huang, Hitti, Rabusseau, and
  Rabbany]{huang2020laplacian}
Shenyang Huang, Yasmeen Hitti, Guillaume Rabusseau, and Reihaneh Rabbany.
\newblock Laplacian change point detection for dynamic graphs.
\newblock In \emph{Proceedings of the 26th ACM SIGKDD International Conference
  on Knowledge Discovery and Data Mining}, pages 349--358, 2020.

\bibitem[Sapiezynski et~al.(2019)Sapiezynski, Stopczynski, Lassen, and
  Lehmann]{sapiezynski2019interaction}
Piotr Sapiezynski, Arkadiusz Stopczynski, David~Dreyer Lassen, and Sune
  Lehmann.
\newblock Interaction data from the copenhagen networks study.
\newblock \emph{Scientific Data}, 6\penalty0 (1):\penalty0 315, 2019.

\bibitem[Sch{\"a}fer et~al.(2014)Sch{\"a}fer, Strohmeier, Lenders, Martinovic,
  and Wilhelm]{schafer2014bringing}
Matthias Sch{\"a}fer, Martin Strohmeier, Vincent Lenders, Ivan Martinovic, and
  Matthias Wilhelm.
\newblock Bringing up opensky: A large-scale ads-b sensor network for research.
\newblock In \emph{IPSN-14 Proceedings of the 13th International Symposium on
  Information Processing in Sensor Networks}, pages 83--94. IEEE, 2014.

\bibitem[MacDonald et~al.(2015)MacDonald, Brauman, Sun, Carlson, Cassidy,
  Gerber, and West]{macdonald2015rethinking}
Graham~K MacDonald, Kate~A Brauman, Shipeng Sun, Kimberly~M Carlson, Emily~S
  Cassidy, James~S Gerber, and Paul~C West.
\newblock Rethinking agricultural trade relationships in an era of
  globalization.
\newblock \emph{BioScience}, 65\penalty0 (3):\penalty0 275--289, 2015.

\bibitem[Voeten et~al.(2009)Voeten, Strezhnev, and Bailey]{DVN/LEJUQZ_2009}
Erik Voeten, Anton Strezhnev, and Michael Bailey.
\newblock {United Nations General Assembly Voting Data}, 2009.
\newblock URL \url{https://doi.org/10.7910/DVN/LEJUQZ}.

\bibitem[Kingma and Ba(2014)]{kingma2014adam}
Diederik~P Kingma and Jimmy Ba.
\newblock Adam: A method for stochastic optimization.
\newblock \emph{arXiv preprint arXiv:1412.6980}, 2014.

\bibitem[Xu et~al.(2019)Xu, Ruan, Motwani, Korpeoglu, Kumar, and
  Achan]{xu2019generative}
Da~Xu, Chuanwei Ruan, Kamiya Motwani, Evren Korpeoglu, Sushant Kumar, and
  Kannan Achan.
\newblock Generative graph convolutional network for growing graphs.
\newblock In \emph{ICASSP 2019 IEEE International Conference on Acoustics,
  Speech and Signal Processing}, pages 3167--3171. IEEE, 2019.

\bibitem[Pareja et~al.(2020)Pareja, Domeniconi, Chen, Ma, Suzumura, Kanezashi,
  Kaler, Schardl, and Leiserson]{pareja2020evolvegcn}
Aldo Pareja, Giacomo Domeniconi, Jie Chen, Tengfei Ma, Toyotaro Suzumura,
  Hiroki Kanezashi, Tim Kaler, Tao Schardl, and Charles Leiserson.
\newblock Evolvegcn: Evolving graph convolutional networks for dynamic graphs.
\newblock In \emph{Proceedings of the AAAI Conference on Artificial
  Intelligence}, volume~34, pages 5363--5370, 2020.

\bibitem[Hajiramezanali et~al.(2019)Hajiramezanali, Hasanzadeh, Narayanan,
  Duffield, Zhou, and Qian]{hajiramezanali2019variational}
Ehsan Hajiramezanali, Arman Hasanzadeh, Krishna Narayanan, Nick Duffield,
  Mingyuan Zhou, and Xiaoning Qian.
\newblock Variational graph recurrent neural networks.
\newblock \emph{Advances in Neural Information Processing Systems}, 32, 2019.

\bibitem[Nguyen et~al.(2018)Nguyen, Lee, Rossi, Ahmed, Koh, and
  Kim]{nguyen2018continuous}
Giang~Hoang Nguyen, John~Boaz Lee, Ryan~A Rossi, Nesreen~K Ahmed, Eunyee Koh,
  and Sungchul Kim.
\newblock Continuous-time dynamic network embeddings.
\newblock In \emph{Companion Proceedings of the Web conference}, pages
  969--976, 2018.

\bibitem[Tian et~al.(2021)Tian, Xiong, and Shi]{tian2021streaming}
Sheng Tian, Tao Xiong, and Leilei Shi.
\newblock Streaming dynamic graph neural networks for continuous-time temporal
  graph modeling.
\newblock In \emph{2021 IEEE International Conference on Data Mining (ICDM)},
  pages 1361--1366, 2021.

\bibitem[Zhou et~al.(2022)Zhou, Zheng, Nisa, Ioannidis, Song, and
  Karypis]{zhou2022tgl}
Hongkuan Zhou, Da~Zheng, Israt Nisa, Vasileios Ioannidis, Xiang Song, and
  George Karypis.
\newblock Tgl: a general framework for temporal gnn training on billion-scale
  graphs.
\newblock \emph{Proceedings of the VLDB Endowment}, 15\penalty0 (8):\penalty0
  1572--1580, 2022.

\bibitem[Souza et~al.(2022)Souza, Mesquita, Kaski, and Garg]{souza2022provably}
Amauri Souza, Diego Mesquita, Samuel Kaski, and Vikas Garg.
\newblock Provably expressive temporal graph networks.
\newblock \emph{Advances in Neural Information Processing Systems},
  35:\penalty0 32257--32269, 2022.

\bibitem[Liu et~al.(2022)Liu, Ma, and Li]{liu2022neural}
Yunyu Liu, Jianzhu Ma, and Pan Li.
\newblock Neural predicting higher-order patterns in temporal networks.
\newblock In \emph{Proceedings of the ACM Web Conference 2022}, pages
  1340--1351, 2022.

\bibitem[Pennebaker et~al.(2001)Pennebaker, Francis, and
  Booth]{pennebaker2001linguistic}
James~W Pennebaker, Martha~E Francis, and Roger~J Booth.
\newblock Linguistic inquiry and word count: Liwc 2001.
\newblock \emph{Mahway: Lawrence Erlbaum Associates}, 71\penalty0
  (2001):\penalty0 2001, 2001.

\bibitem[Jin et~al.(2023)Jin, Lee, Sharma, Ye, Sikka, Divakaran, and
  Kumar]{jin2023predicting}
Yiqiao Jin, Yeon-Chang Lee, Kartik Sharma, Meng Ye, Karan Sikka, Ajay
  Divakaran, and Srijan Kumar.
\newblock Predicting information pathways across online communities.
\newblock \emph{arXiv preprint arXiv:2306.02259}, 2023.

\bibitem[Huang et~al.(2022)Huang, Yang, Wang, Wang, Zhang, Xu, Chen, and
  Vazirgiannis]{huang2022dgraph}
Xuanwen Huang, Yang Yang, Yang Wang, Chunping Wang, Zhisheng Zhang, Jiarong Xu,
  Lei Chen, and Michalis Vazirgiannis.
\newblock Dgraph: A large-scale financial dataset for graph anomaly detection.
\newblock \emph{Advances in Neural Information Processing Systems},
  35:\penalty0 22765--22777, 2022.

\end{thebibliography}
	\section*{Checklist}

	\begin{enumerate}
		
		\item For all authors...
		\begin{enumerate}
			\item Do the main claims made in the abstract and introduction accurately reflect the paper's contributions and scope?
			\answerYes{}
			\item Did you describe the limitations of your work?
			\answerYes{See Section~\ref{subsec:lp-performance}, we fail to implement TGAT on UNTrade dataset.}
			\item Did you discuss any potential negative societal impacts of your work?
			\answerNA{}
			\item Have you read the ethics review guidelines and ensured that your paper conforms to them?
			\answerYes{}
		\end{enumerate}
		
		\item If you are including theoretical results...
		\begin{enumerate}
			\item Did you state the full set of assumptions of all theoretical results?
			\answerNA{}
			\item Did you include complete proofs of all theoretical results?
			\answerNA{}
		\end{enumerate}
		
		\item If you ran experiments (e.g. for benchmarks)...
		\begin{enumerate}
			\item Did you include the code, data, and instructions needed to reproduce the main experimental results (either in the supplemental material or as a URL)?
			\answerYes{See Section~\ref{sec:intro}:  \href{https://drive.google.com/drive/folders/1HKSFGEfxHDlHuQZ6nK4SLCEMFQIOtzpz?usp=sharing}{datasets}, \href{https://pypi.org/project/benchtemp/}{PyPI},  \href{https://my-website-6gnpiaym0891702b-1257259254.tcloudbaseapp.com/}{leaderboard}, and \href{https://github.com/qianghuangwhu/benchtemp}{github project}.}
			\item Did you specify all the training details (e.g., data splits, hyperparameters, how they were chosen)?
			\answerYes{See Section \ref{sec:pipeline}, Section~\ref{subsec:Experimentalsetup}, Appendix~\apc{B}, and Appendix~\apc{C}.}
			\item Did you report error bars (e.g., with respect to the random seed after running experiments multiple times)?
			\answerYes{See Section~\ref{sec:Experiments}, Table~\ref{tab:lp-auc}, and Table~\ref{tab:nc-auc}.}
			\item Did you include the total amount of compute and the type of resources used (e.g., type of GPUs, internal cluster, or cloud provider)?
			\answerYes{See Section~\ref{subsec:Experimentalsetup}.}
		\end{enumerate}
		
		\item If you are using existing assets (e.g., code, data, models) or curating/releasing new assets...
		\begin{enumerate}
			\item If your work uses existing assets, did you cite the creators?
			\answerYes{See Section~\ref{subsec:dataset}.}
			\item Did you mention the license of the assets?
			\answerYes{See Section~\ref{subsec:dataset}, CC BY-NC \href{https://creativecommons.org/licenses/by-nc/4.0/}{license}.}
			\item Did you include any new assets either in the supplemental material or as a URL?
			\answerYes{See Section~\ref{sec:intro}.}
			\item Did you discuss whether and how consent was obtained from people whose data you're using/curating?
			\answerYes{See Section \ref{subsec:dataset}, the datasets are open-sourced  with public license.}
			\item Did you discuss whether the data you are using/curating contains personally identifiable information or offensive content?
			\answerNA{}
		\end{enumerate}
		
		\item If you used crowdsourcing or conducted research with human subjects...
		\begin{enumerate}
			\item Did you include the full text of instructions given to participants and screenshots, if applicable?
			\answerNA{}
			\item Did you describe any potential participant risks, with links to Institutional Review Board (IRB) approvals, if applicable?
			\answerNA{}
			\item Did you include the estimated hourly wage paid to participants and the total amount spent on participant compensation?
			\answerNA{}
		\end{enumerate}
		
	\end{enumerate}

	\appendix
	
	\section{Temporal Graph Datasets}
	\label{appendix:datasets_intro}
	
	As shown in Table \apc{2} of the main paper, we select fifteen benchmark databases from diverse domains. All datasets are publicly available under  \href{https://creativecommons.org/licenses/by-nc/4.0/}{CC BY-NC licence} and can be accessed at \url{https://drive.google.com/drive/folders/1HKSFGEfxHDlHuQZ6nK4SLCEMFQIOtzpz?usp=sharing}.
	Figure~\ref{fig:edge_distribution} shows
	the temporal distribution of edges in the evaluated temporal graphs.
	
	\begin{itemize}
		
		\item \href{http://snap.stanford.edu/jodie/reddit.csv}{Reddit} is a bipartite interaction graph, consisting of one month of posts made by users on subreddits \cite{url_reddit}. Users and subreddits are
		nodes, and egdes are interactions of users writing posts to subreddits. The text of each post is converted into LIWC-feature vector \cite{pennebaker2001linguistic} as an edge feature of length 172.
		This public dataset gives 366 true labels
		among 672,447 interactions, and those true label  are ground-truth labels of banned users from Reddit \cite{kumar2019predicting}. 
		
		\item \href{http://snap.stanford.edu/jodie/wikipedia.csv}{Wikipedia} is a bipartite interaction graph, and  contains one month of edits made by editors. 
		This public dataset selects the 1,000 most edited pages as items and editors who made at least 5 edits as users over a month \cite{kumar2019predicting}. Editors and pages are nodes, and edges are interactions of editors editing on pages. Edge features of length 172 are interaction edits converted into LIWC-feature vectors~\cite{pennebaker2001linguistic}. Wikipedia dataset treats 217 public ground-truth labels of banned users from 157,474 interactions  as positive labels.
		
		\item \href{http://snap.stanford.edu/jodie/mooc.csv}{MOOC} is a bipartite MOOC online network
		of students and online course content units \cite{url_mooc}. 
		Students and courses are nodes, and edges with features of length 4 are interactions of user viewing a video, submitting an answer, etc.
		This public dataset treats 4,066 dropout events out of 411,749 interactions as positive labels~\cite{kumar2019predicting}.
		
		\item \href{http://snap.stanford.edu/jodie/lastfm.csv}{LastFM} is a user-song bipartite network \cite{hidasi2012fast}. 
		Users and songs are nodes, and edges are user-listens-song interactions. 
		This public dataset  includes 1,293,103 interactions between  all 1000 users and the 1000 most listened songs~\cite{kumar2019predicting}.
		
		\item \href{https://www.cs.cmu.edu/~./enron/}{Enron} is an email communication network that collects about half a million emails over several years~\cite{url_enron}. 
		Nodes of the network are email addresses, and edges are email communication between accounts  \cite{leskovec2014snap}.
		
		\item \href{http://realitycommons.media.mit.edu/socialevolution.html}{SocialEvo} is a network in which experiments are conducted to closely track the everyday life of a whole undergraduate dormitory with mobile phones. 
		This public dataset is collected by a cell phone application every six minutes, and contains physical proximity and location between students living in halls of residence. 
		\cite{madan2011sensing}.
		
		\item \href{http://konect.cc/networks/opsahl-ucforum/}{UCI} is a facebook-like social network that contains user posts to forums. 
		Nodes are students (1,899) at University of California, Irvine, and edges are interactions of online messages (59,835) among these users~\cite{opsahl2009clustering}. 
		Each edge has 100 features. 
		
		\item \href{http://snap.stanford.edu/data/CollegeMsg.html}{CollegeMsg} is provided by the SNAP team of Stanford \cite{panzarasa2009patterns}. This dataset is derived from the facebook-like social network introduced in dataset UCI. 
		The SNAP team has parsed it to a temporal network.  Each edge has 172 features.

		\begin{figure}[!t]
			\centering
			\begin{minipage}[b]{0.32\linewidth}
				\centering
				\includegraphics[width=\linewidth]{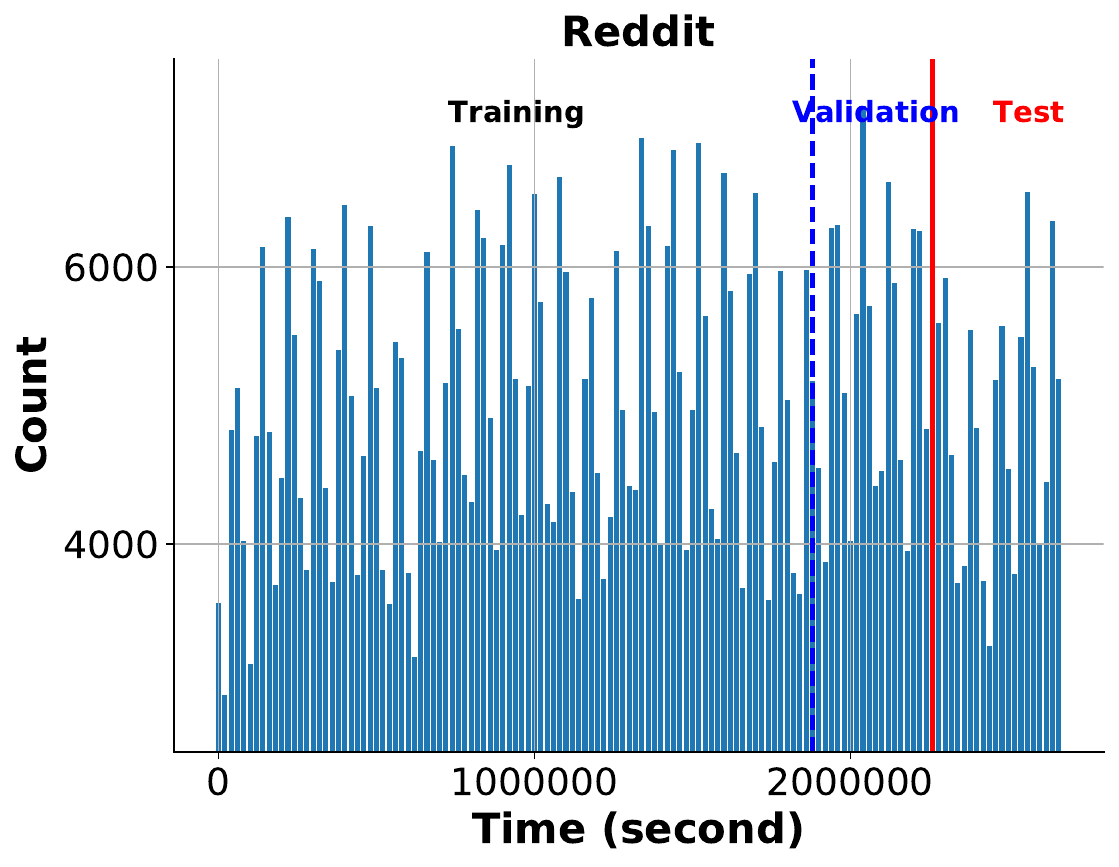}
				\subcaption{Reddit}
			\end{minipage}
			\begin{minipage}[b]{0.32\linewidth}
				\centering
				\includegraphics[width=\linewidth]{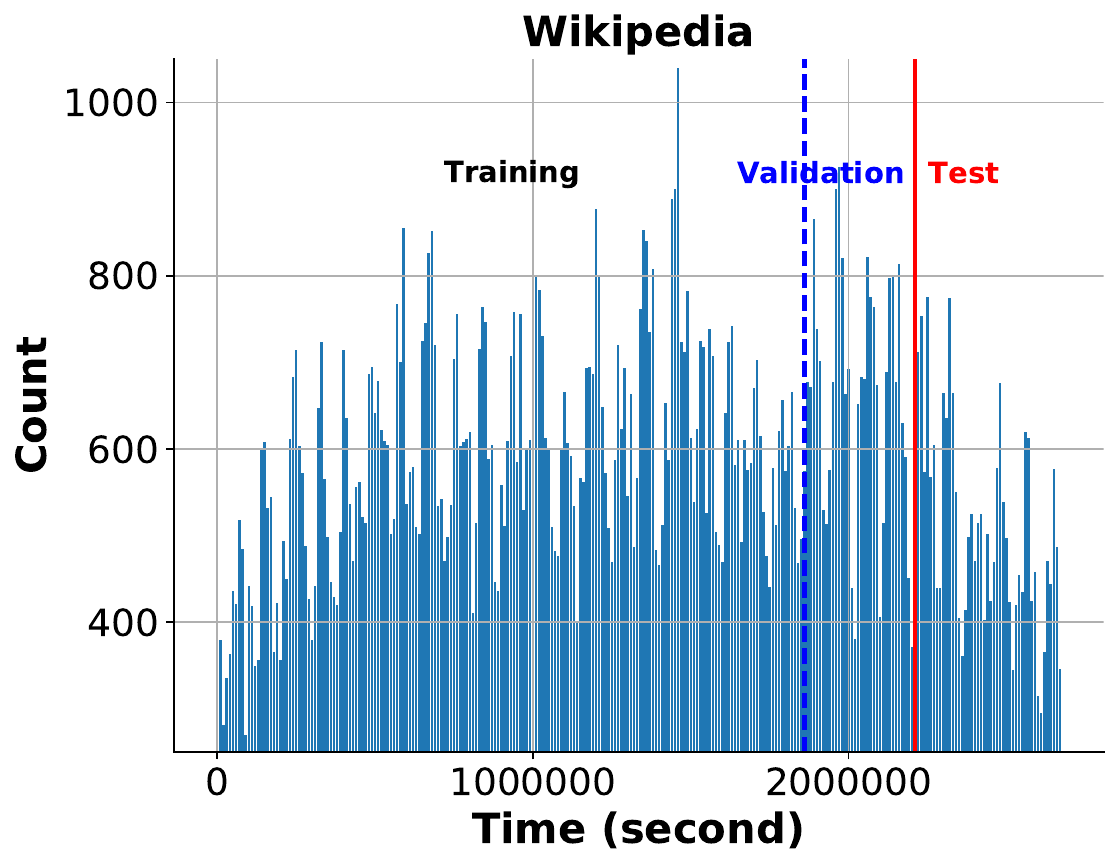}
				\subcaption{Wikipedia}
			\end{minipage}
			\begin{minipage}[b]{0.32\linewidth}
				\centering
				\includegraphics[width=\linewidth]{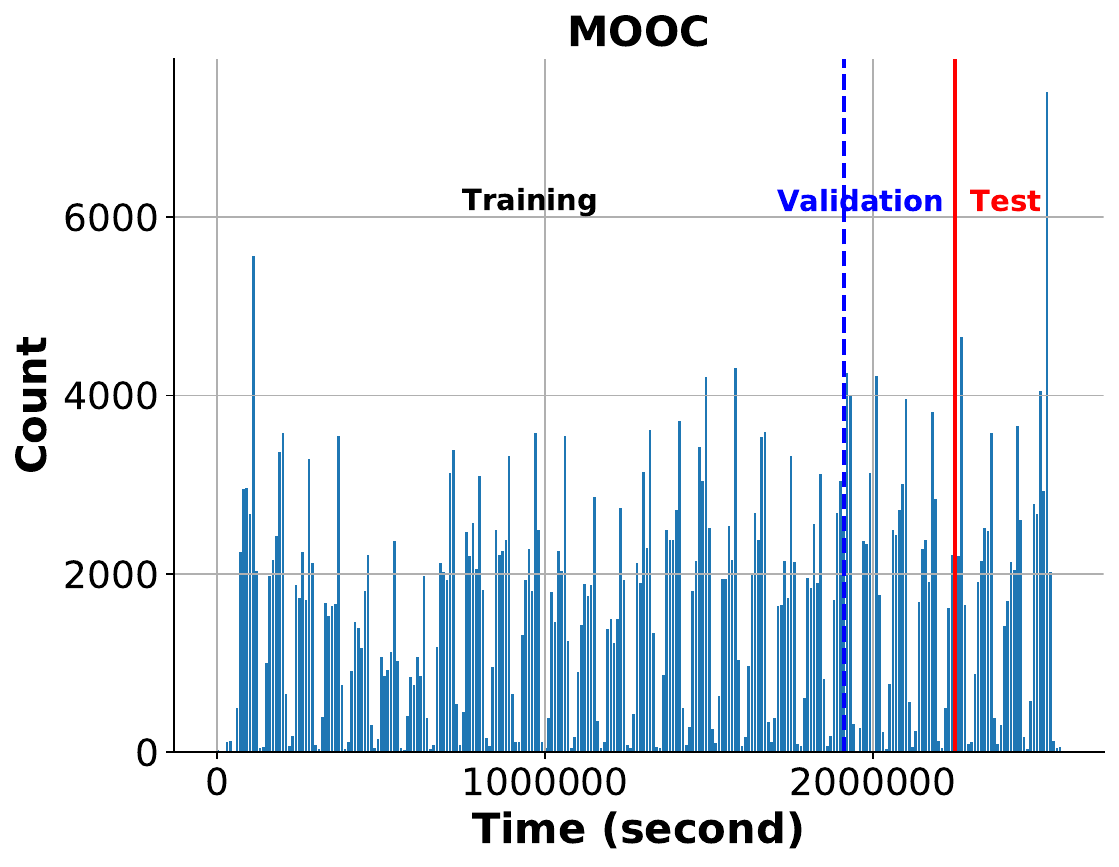}
				\subcaption{MOOC}
			\end{minipage}
			
			\begin{minipage}[b]{0.32\linewidth}
				\centering
				\includegraphics[width=\linewidth]{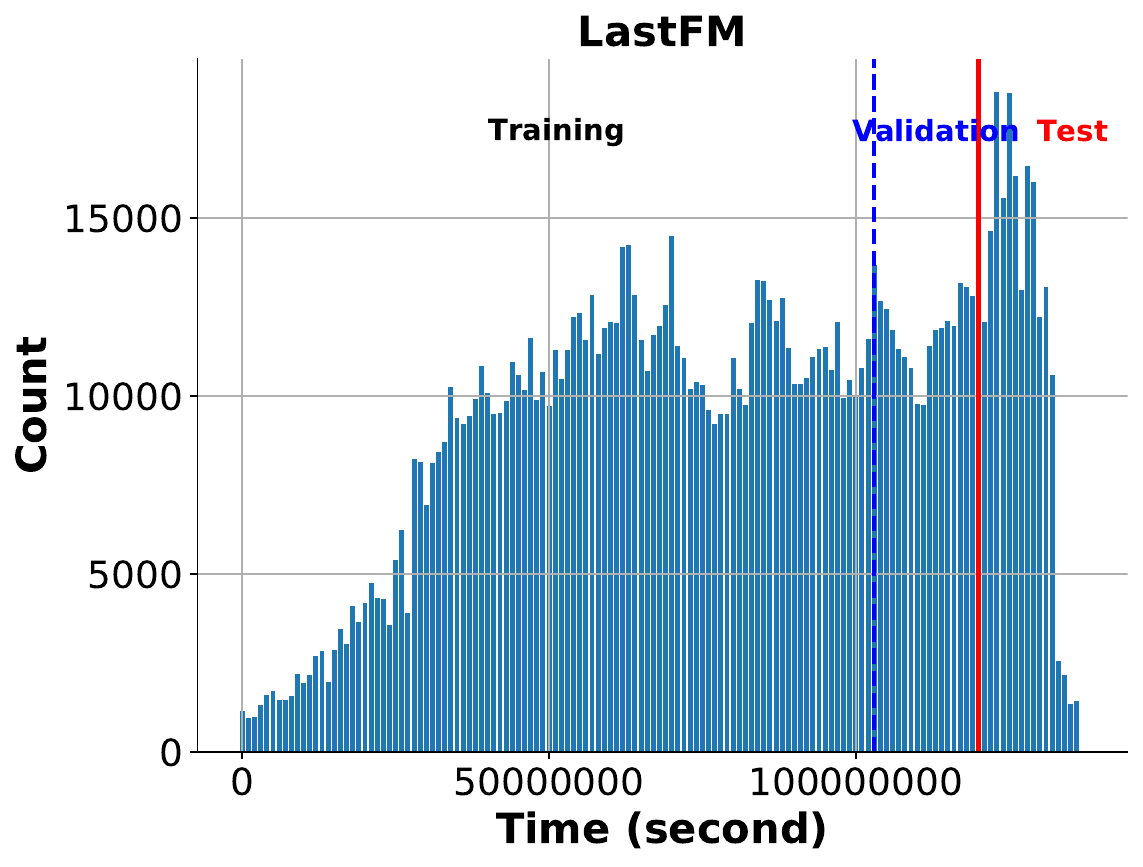}
				\subcaption{LastFM}
			\end{minipage}
			\begin{minipage}[b]{0.32\linewidth}
				\centering
				\includegraphics[width=\linewidth]{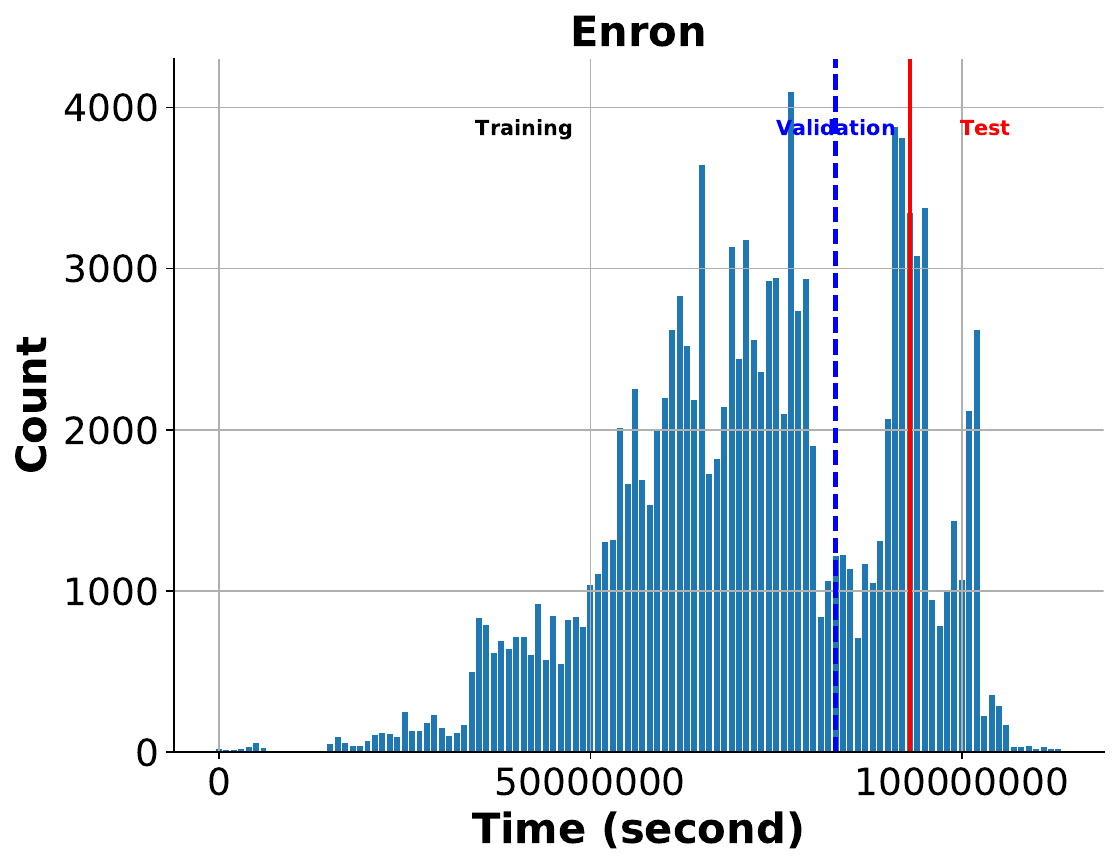}
				\subcaption{Enron}
			\end{minipage}
			\begin{minipage}[b]{0.32\linewidth}
				\centering
				\includegraphics[width=\linewidth]{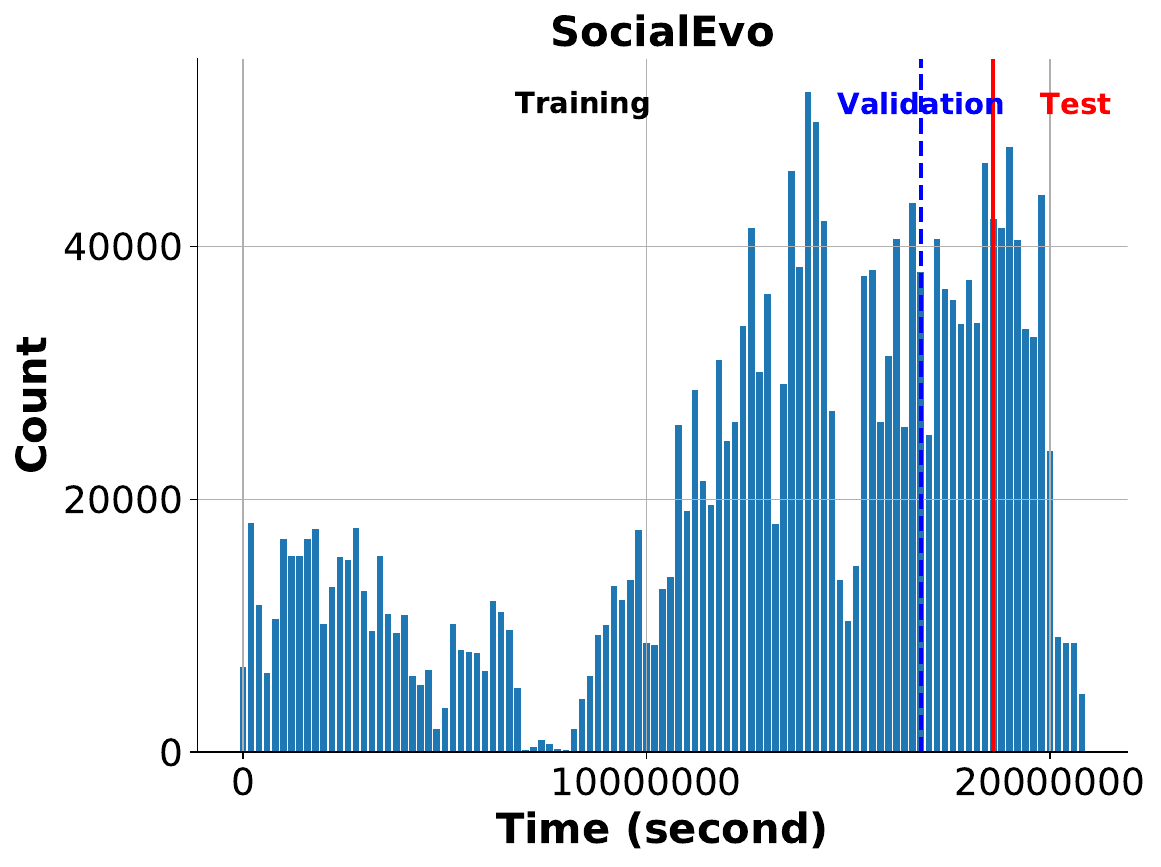}
				\subcaption{SocialEvo}
			\end{minipage}
			
			\begin{minipage}[b]{0.32\linewidth}
				\centering
				\includegraphics[width=\linewidth]{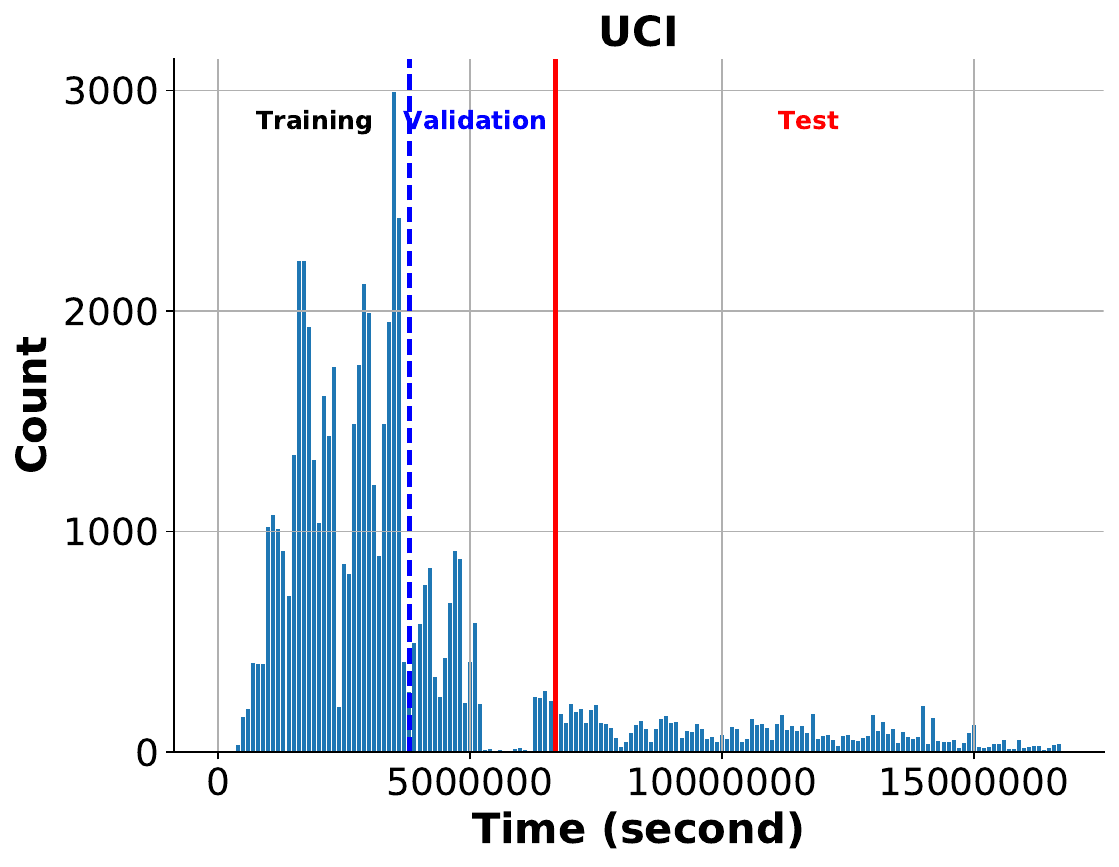}
				\subcaption{UCI}
			\end{minipage}
			\begin{minipage}[b]{0.32\linewidth}
				\centering
				\includegraphics[width=\linewidth]{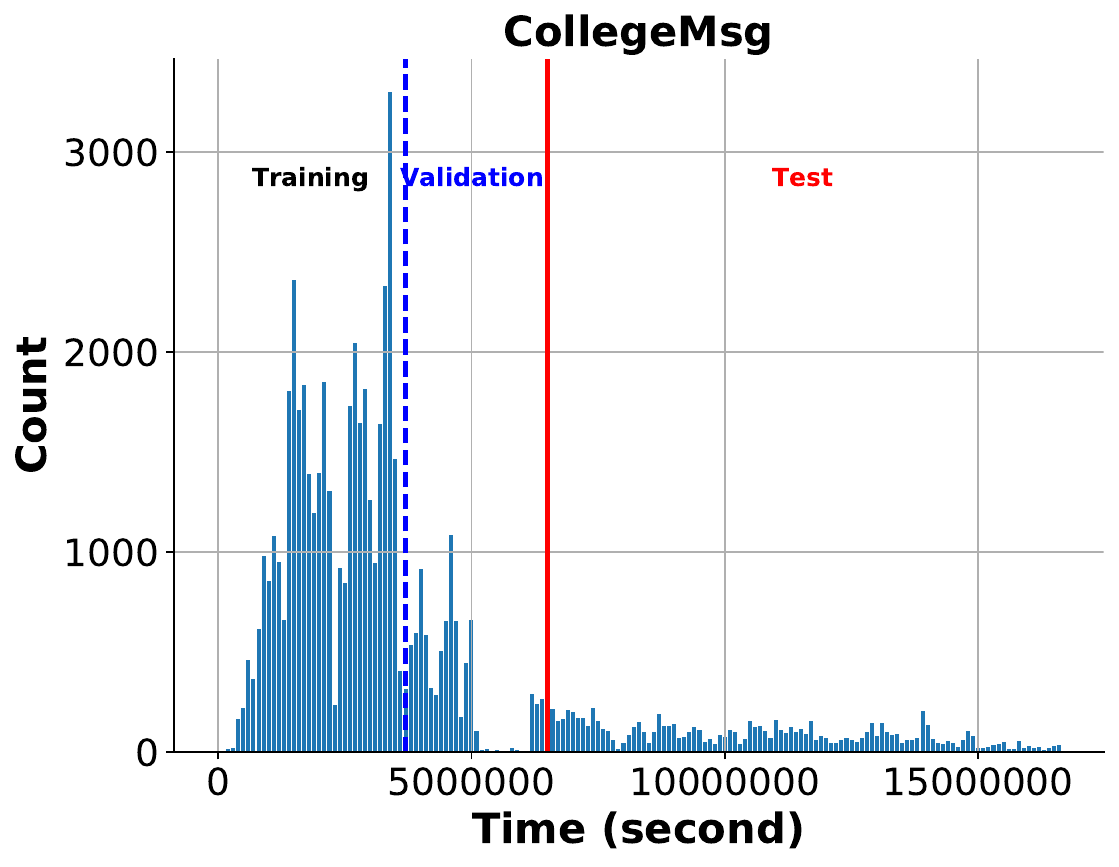}
				\subcaption{CollegeMsg}
			\end{minipage}
			\begin{minipage}[b]{0.32\linewidth}
				\centering
				\includegraphics[width=\linewidth]{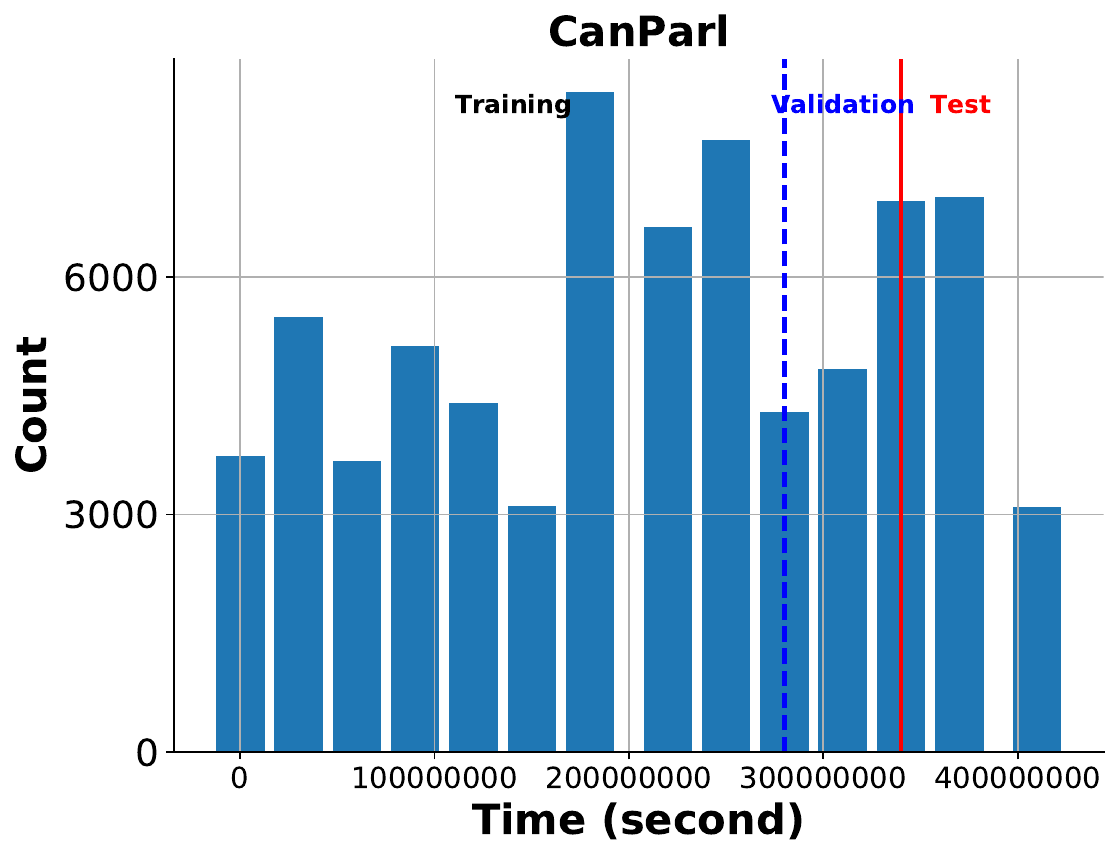}
				\subcaption{CanParl}
			\end{minipage}
			
			\begin{minipage}[b]{0.32\linewidth}
				\centering
				\includegraphics[width=\linewidth]{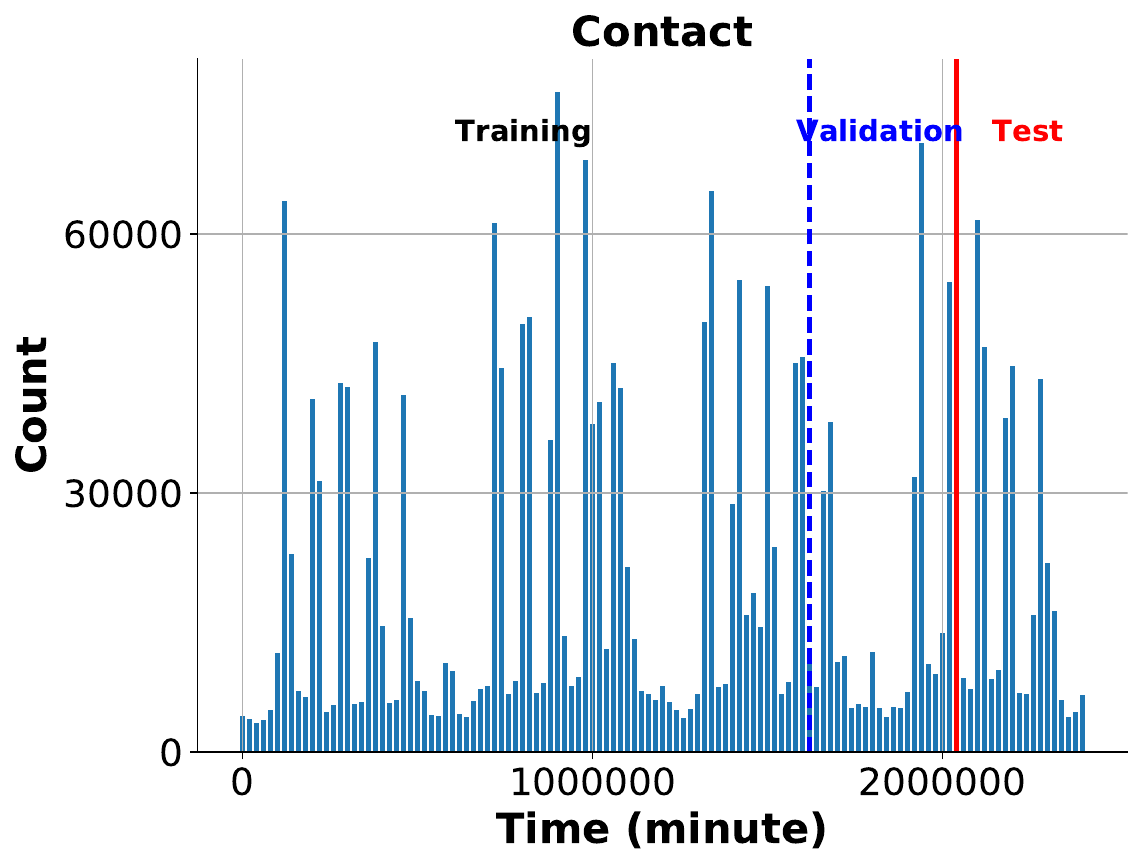}
				\subcaption{Contact}
			\end{minipage}
			\begin{minipage}[b]{0.32\linewidth}
				\centering
				\includegraphics[width=\linewidth]{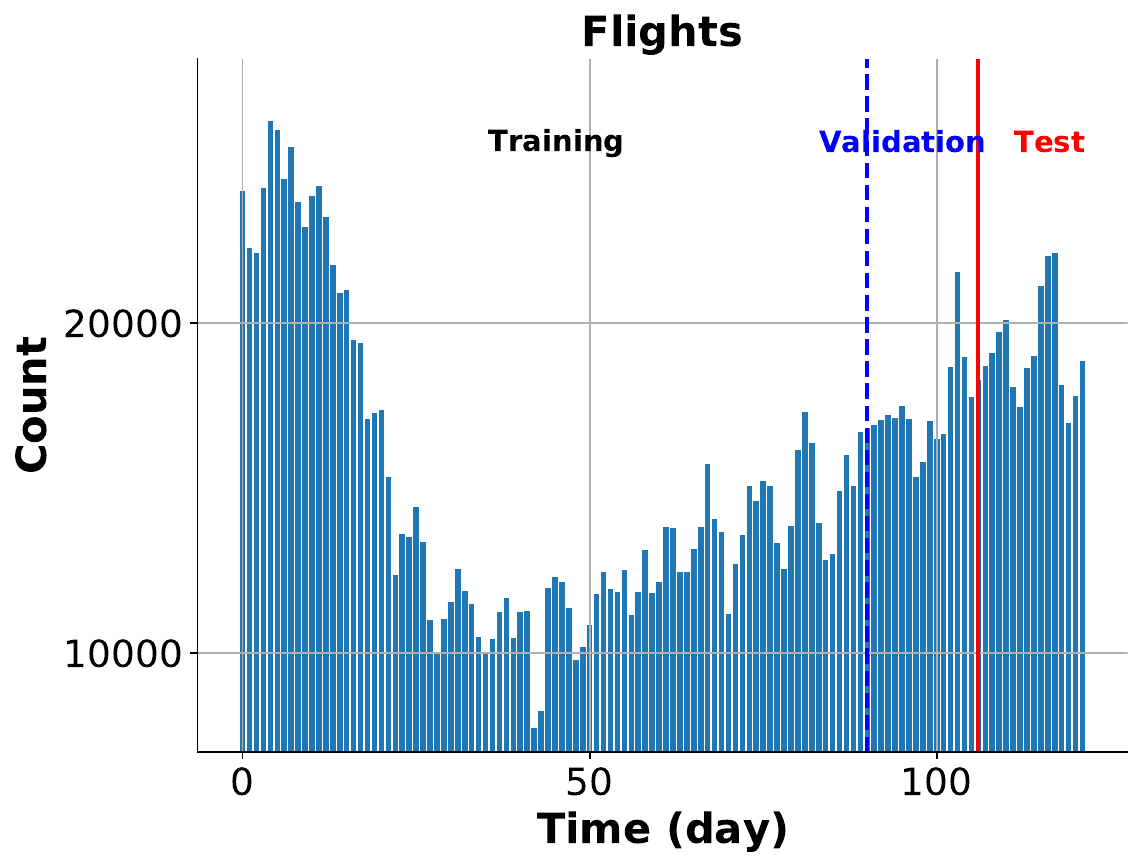}
				\subcaption{Flights}
			\end{minipage}
			\begin{minipage}[b]{0.32\linewidth}
				\centering
				\includegraphics[width=\linewidth]{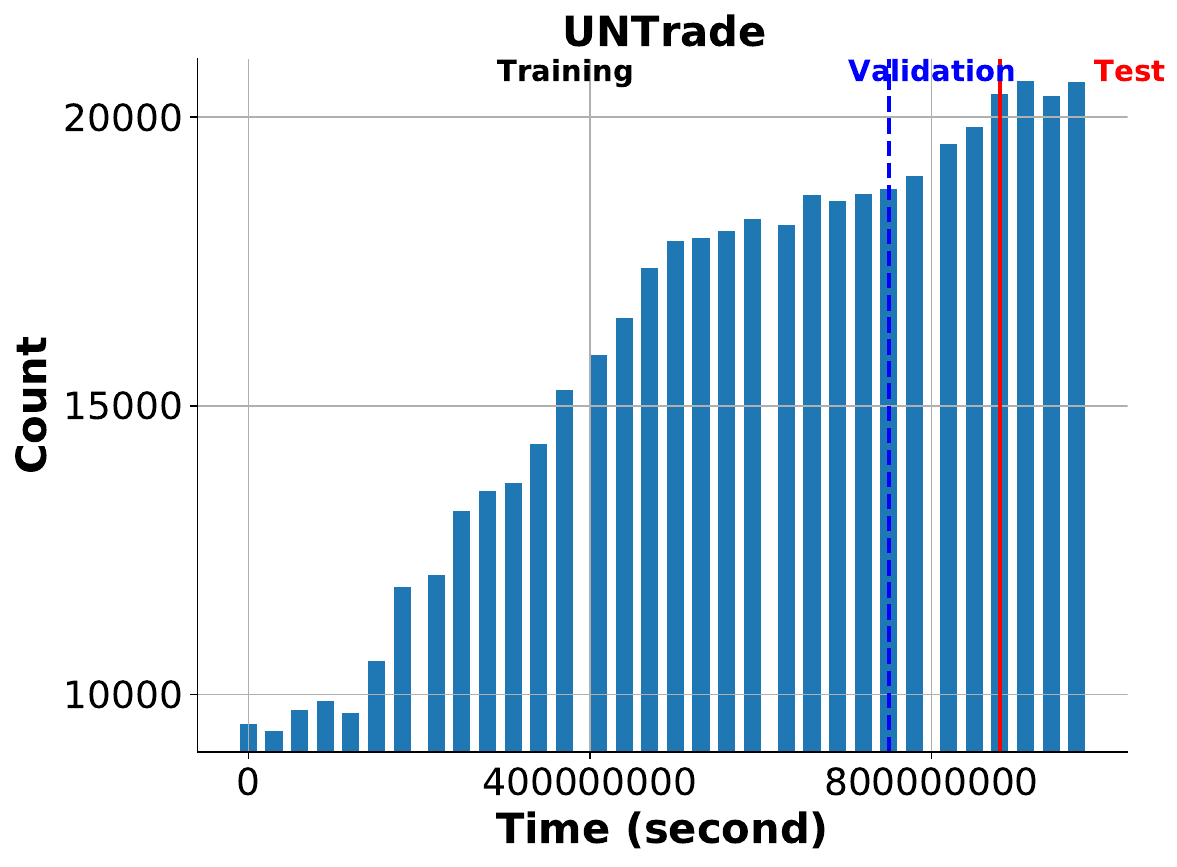}
				\subcaption{UNTrade}
			\end{minipage}

			\begin{minipage}[b]{0.32\linewidth}
				\centering
				\includegraphics[width=\linewidth]{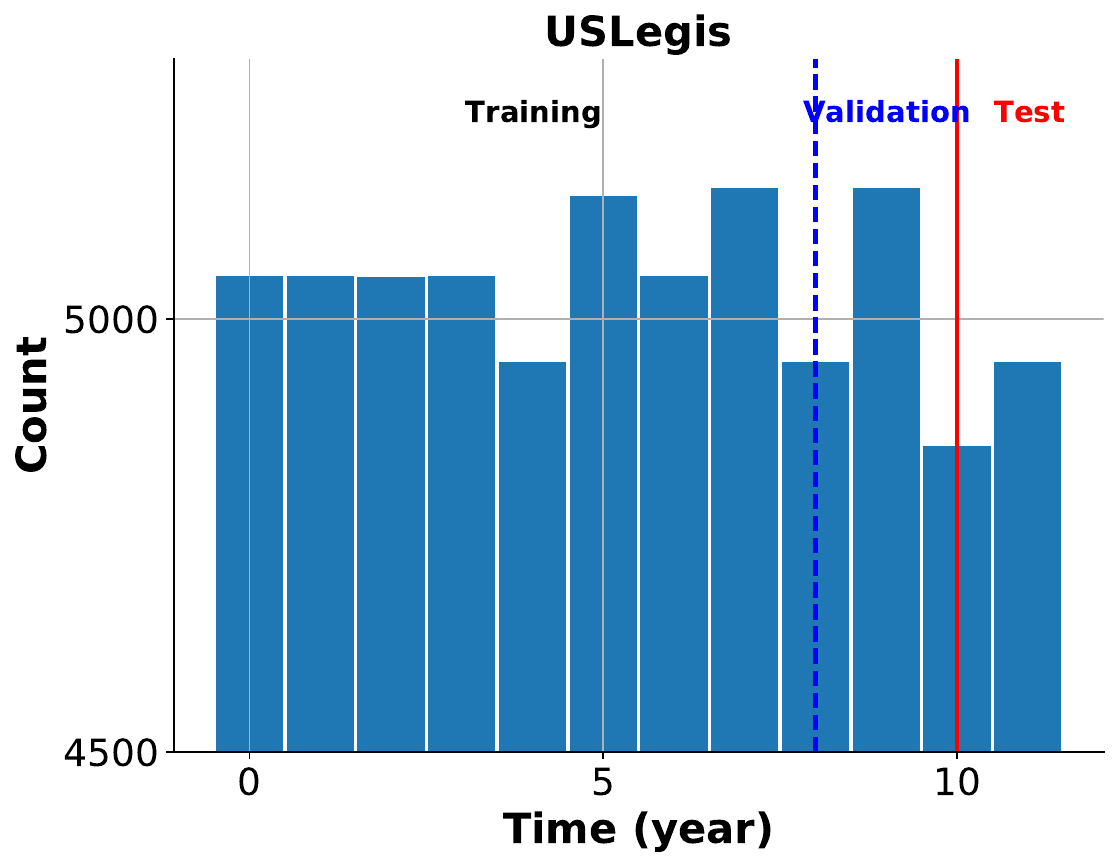}
				\subcaption{USLegis}
			\end{minipage}
			\begin{minipage}[b]{0.32\linewidth}
				\centering
				\includegraphics[width=\linewidth]{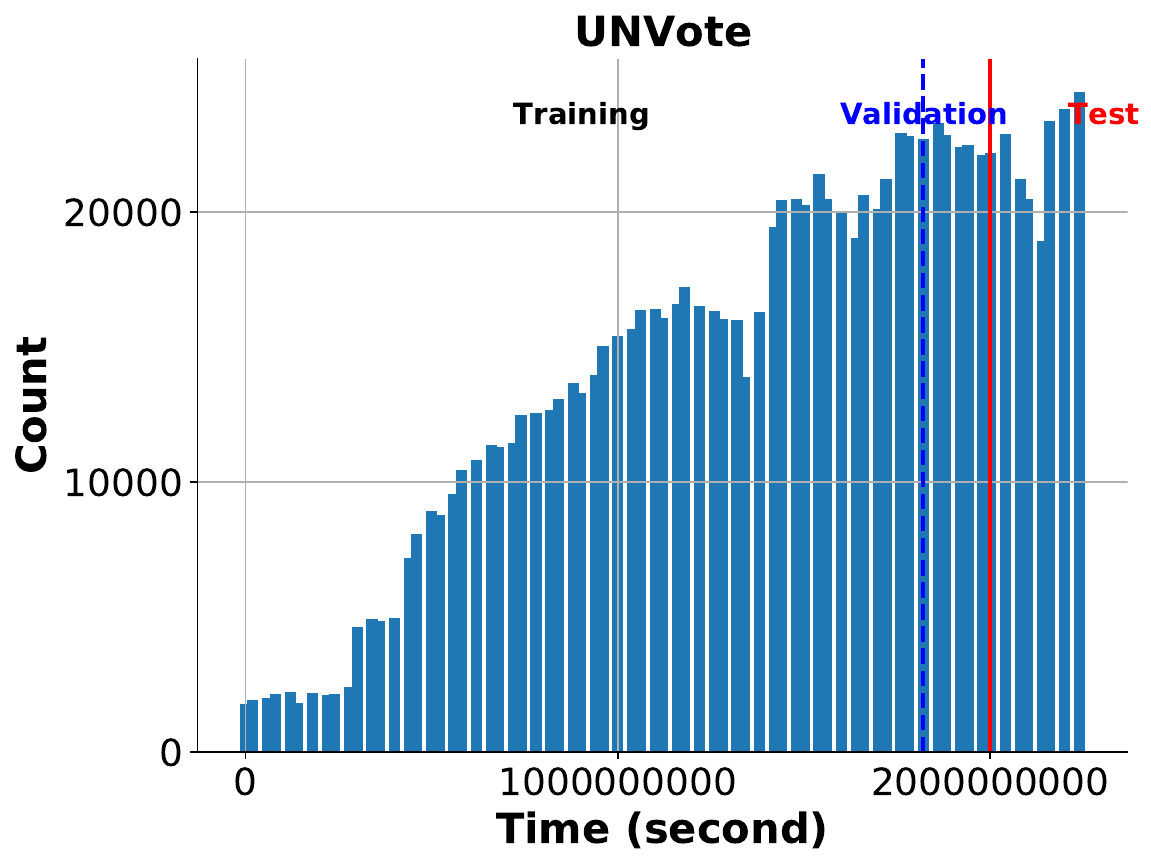}
				\subcaption{UNVote}
			\end{minipage}
			\begin{minipage}[b]{0.32\linewidth}
				\centering
				\includegraphics[width=\linewidth]{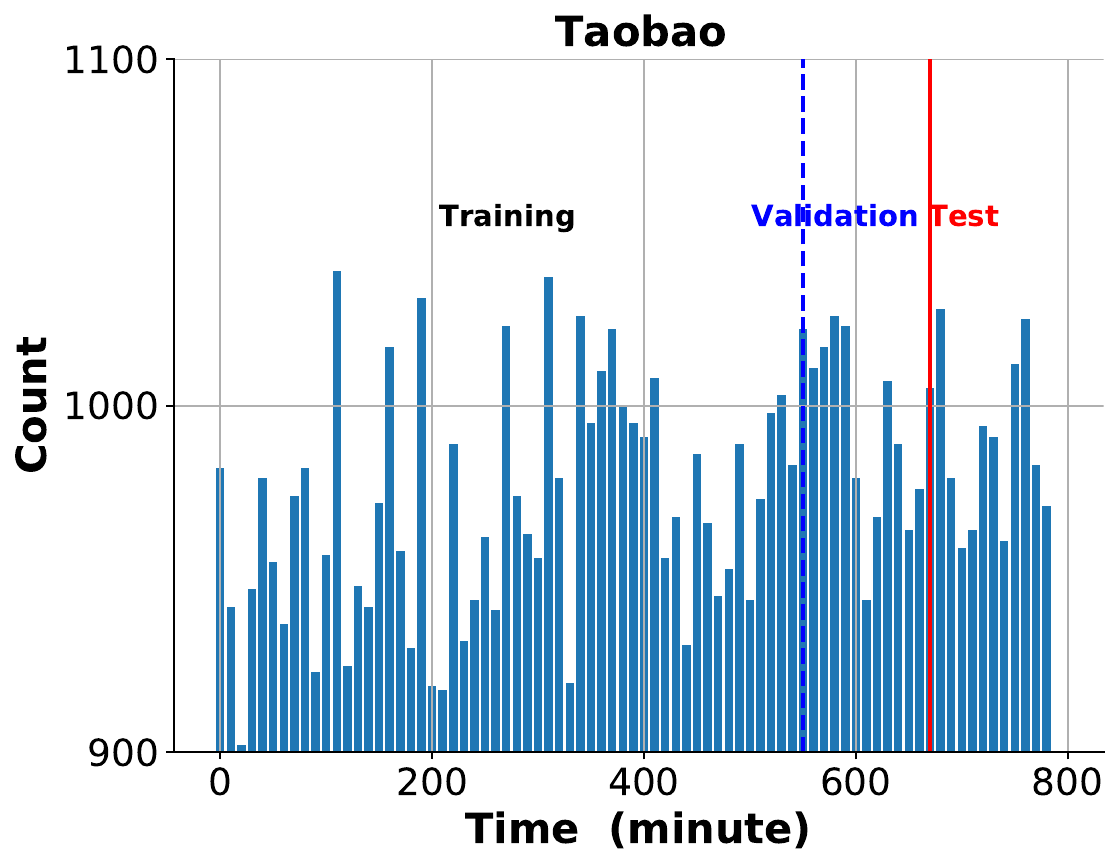}
				\subcaption{Taobao}
			\end{minipage}
			\caption{The temporal distribution of edges of the evaluated temporal graphs.}
			\label{fig:edge_distribution}
			
		\end{figure}

		\item \href{https://github.com/shenyangHuang/LAD}{CanParl} is a Canadian parliament bill voting
		network extracted from open \href{https://openparliament.ca/}{website} 
		\cite{huang2020laplacian}. Nodes are members of parliament (MPs), and edges are the interactions between  MPs from 2006 to 2019.
		
		\item  \href{https://springernature.figshare.com/articles/dataset/Metadata_record_for_Interaction_data_from_the_Copenhagen_Networks_Study/11283407/1}{Contact} is a temporal and weighted network of physical proximity among the participants~\cite{sapiezynski2019interaction}. Nodes are participant and edges are proximity events between the study participants. 
		Edge features indicate the physical proximity between participants~\cite{poursafaeitowards}.
		
		\item \href{https://zenodo.org/record/3974209/#.Yf62HepKguU}{Flights} is 
		a weighted flight network. 
		Nodes are airports, and edges are tracked flights~\cite{schafer2014bringing}. 
		The weights of edges indicate the
		number of flights between two given airports within a day \cite{poursafaeitowards}. 
		
		\item \href{https://www.fao.org/faostat/en/#data/TM}{UNTrade} is a food and agriculture trading weighted network among 181 nations over 30 years \cite{macdonald2015rethinking}. 
		Nodes are countries, and edges are tradings between two countries. 
		The weights of edges are the total sum of normalized
		agriculture import or export values between two given countries \cite{poursafaeitowards}.  
		
		\item  \href{https://github.com/shenyangHuang/LAD}{USLegis} is a senate co-sponsorship network that examines the social relations of legislators in their co-sponsorship relationships on bills~\cite{huang2020laplacian}. 
		Nodes are congress members, and edge weights are the number of times that two members of congress co-sponsor a bill in a given congress \cite{poursafaeitowards}.

		\item \href{https://dataverse.harvard.edu/dataset.xhtml?persistentId=doi:10.7910/DVN/LEJUQZ}{UNVote} is a weighted network of roll-call votes in the UN General Assembly 1946-2021~\cite{DVN/LEJUQZ_2009}. Nodes are nations, and edge weights are the number of times both nations have voted "yes" to an item.
		
		\item \href{https://tianchi.aliyun.com/dataset/649}{Taobao} is a subset of the Taobao user behavior dataset intercepted based on the period 8:00 to 18:00 on 26 November 2017~\cite{zhu2018learning}. This public dataset is a user-item 
		bipartite network. 
		Nodes are users and items, and edges are behaviors between users and items, such as favor, click, purchase, and add an item to shopping cart.
		Each edge has 4 features, corresponding to 4 different types of behaviors~\cite{jin2022neural}.
		
	\end{itemize}
	
	
	\section{Experiment Details}
	\label{appendix:experimentdetails}
	
	DataLoader of the link prediction pipeline introduced in Section \apc{3.2.1} splits and generates 
	training set, validation set, transductive set, and inductive test sets depending on the New-Old and New-New settings, for link prediction task.
	The detailed statistics of these data sets are shown in Table \ref{tab:lp-nodesedges}. 
	DataLoader of the node classification pipeline introduced in Section \apc{3.2.2} follows the traditional transductive setting. 
	The detailed statistics of the training set, validation set, and test set on three available datasets (Reddit, Wikipedia, and MOOC) are given in Table \ref{tab:nc-nodesedges}.

	\begin{table}[!h]
		\scriptsize
		\setlength{\tabcolsep}{5.5pt}
		\centering
		\caption{Statistics of datasets for link prediction task. 
			"New-Old Validation" indicates the validation set under Inductive New-Old setting, and so on.} 
		\label{tab:lp-nodesedges}
		\begin{tabular}{l|ll|ll|ll|ll|ll}
			\toprule
			&\multicolumn{2}{c|}{\textbf{Training}}&\multicolumn{2}{c|}{\textbf{Validation}}&\multicolumn{2}{c|}{\textbf{Transductive Test}}&\multicolumn{2}{c|}{\textbf{Inductive  Validation}}&\multicolumn{2}{c}{\textbf{Inductive  Test}}\\
			\cmidrule(lr){2-3}\cmidrule(lr){4-5}\cmidrule(lr){6-7}\cmidrule(lr){8-9}\cmidrule(lr){10-11}
			
			&\textit{\# nodes}&\textit{\# edges}&\textit{\# nodes}&\textit{\# edges}&\textit{\# nodes}&\textit{\# edges}&\textit{\# nodes}&\textit{\# edges}&\textit{\# nodes}&\textit{\# edges}\\
			\midrule
			Reddit & 9,574  & 389,989  & 9,839  & 100,867  & 9,615  & 100,867  & 3,491  & 19,446  & 3,515  & 21,470   \\ 
			Wikipedia & 6,141  & 81,029  & 3,256  & 23,621  & 3,564  & 23,621  & 2,120  & 12,016  & 2,437  & 11,715   \\ 
			MOOC & 6,015  & 227,485  & 2,599  & 61,762  & 2,412  & 61,763  & 2,333  & 25,592  & 2,181  & 29,179   \\ 
			LastFM & 1,612  & 722,758  & 1,714  & 193,965  & 1,753  & 193,966  & 1,643  & 57,651  & 1,674  & 98,442   \\ 
			Enron & 157  & 79,064  & 155  & 18,786  & 141  & 18,785  & 112  & 5,637  & 110  & 4,859   \\ 
			SocialEvo & 67  & 1,222,980  & 64  & 314,930  & 62  & 314,924  & 62  & 62,811  & 60  & 70,038   \\ 
			UCI & 1,338  & 34,386  & 1,036  & 8,975  & 847  & 8,976  & 816  & 4,761  & 678  & 5,707   \\ 
			CollegeMsg & 1,337  & 34,544  & 1,036  & 8,975  & 847  & 8,975  & 818  & 4,914  & 680  & 5,885   \\ 
			CanParl & 618  & 47,435  & 344  & 11,809  & 342  & 10,113  & 344  & 5,481  & 341  & 5,591   \\ 
			Contact & 617  & 1,372,030  & 632  & 364,005  & 629  & 363,780  & 582  & 68,261  & 590  & 69,617   \\ 
			Flights & 11,230  & 1,107,798  & 10,844  & 279,399  & 10,906  & 287,824  & 6,784  & 54,861  & 6,820  & 58,102   \\ 
			UNTrade & 230  & 291,287  & 230  & 78,721  & 228  & 61,595  & 227  & 17,528  & 226  & 14,001   \\ 
			USLegis & 176  & 38,579  & 113  & 10,005  & 100  & 4,950  & 113  & 5,010  & 100  & 3,297   \\ 
			UNVote & 178  & 600,511  & 194  & 135,298  & 194  & 155,119  & 194  & 28,136  & 194  & 33,083   \\ 
			Taobao & 54,462  & 45,630  & 17,964  & 11,621  & 18,143  & 11,550  & 16,476  & 10,338  & 16,896  & 10,516  \\ 
			\midrule
			&\multicolumn{2}{c|}{\textbf{New-Old  Validation}} &\multicolumn{2}{c|}{\textbf{New-Old  Test}}&\multicolumn{2}{c|}{\textbf{New-New Validation}}&\multicolumn{2}{c|}{\textbf{New-New Test}}&\multicolumn{2}{c}{\multirow{2}{*}{\textbf{Unseen Nodes}}}\\
			\cmidrule(lr){2-3}\cmidrule(lr){4-5}\cmidrule(lr){6-7}\cmidrule(lr){8-9}
			
			&\textit{\# nodes}&\textit{\# edges}&\textit{\# nodes}&\textit{\# edges}&\textit{\# nodes}&\textit{\# edges}&\textit{\# nodes}&\textit{\# edges}&&\\
			\toprule
			Reddit & 3,301  & 16,760  & 3,325  & 18,703  & 488  & 2,686  & 486  & 2,767  & \multicolumn{2}{c}{1,098  } \\ 
			Wikipedia & 1,809  & 8,884  & 1,996  & 8,148  & 468  & 3,132  & 629  & 3,567  & \multicolumn{2}{c}{922 }  \\ 
			MOOC & 2,316  & 23,109  & 2,164  & 25,730  & 553  & 2,483  & 592  & 3,449  & \multicolumn{2}{c}{714}   \\ 
			LastFM & 1,642  & 52,379  & 1,674  & 63,505  & 272  & 5,272  & 331  & 34,937  & \multicolumn{2}{c}{198 }  \\ 
			Enron & 111  & 4,965  & 109  & 4,262  & 19  & 672  & 20  & 597  & \multicolumn{2}{c}{18 }  \\ 
			SocialEvo & 62  & 58,959  & 60  & 65,466  & 7  & 3,852  & 7  & 4,572  & \multicolumn{2}{c}{7  } \\ 
			UCI & 757  & 3,686  & 606  & 4,193  & 247  & 1,075  & 213  & 1,514  & \multicolumn{2}{c}{189  } \\ 
			CollegeMsg & 759  & 3,839  & 608  & 4,328  & 247  & 1,075  & 214  & 1,557  & \multicolumn{2}{c}{189   }\\ 
			CanParl & 344  & 4,543  & 341  & 4,469  & 106  & 938  & 111  & 1,122  & \multicolumn{2}{c}{73  } \\ 
			Contact & 582  & 64,887  & 590  & 65,883  & 62  & 3,374  & 59  & 3,734  & \multicolumn{2}{c}{69  } \\ 
			Flights & 6,711  & 49,796  & 6,739  & 52,504  & 874  & 5,065  & 937  & 5,598  & \multicolumn{2}{c}{1,316  } \\ 
			UNTrade & 227  & 16,420  & 226  & 13,112  & 25  & 1,108  & 25  & 889  & \multicolumn{2}{c}{25  } \\ 
			USLegis & 112  & 4,154  & 100  & 2,436  & 37  & 856  & 42  & 861  & \multicolumn{2}{c}{22  } \\ 
			UNVote & 194  & 26,545  & 194  & 31,166  & 23  & 1,591  & 23  & 1,917  & \multicolumn{2}{c}{20}   \\ 
			Taobao & 9,247  & 5,678  & 8,136  & 4,860  & 7,706  & 4,660  & 9,298  & 5,656  & \multicolumn{2}{c}{8,256}  \\ 
			
			\bottomrule
		\end{tabular}
		
	\end{table}
	
	\begin{table}[!h]
		\footnotesize
		\setlength{\tabcolsep}{4.5pt}
		\centering
		\caption{Statistics of datasets for node classification task.} 
		\label{tab:nc-nodesedges}
		\begin{tabular}{l|ll|ll|ll}
			\toprule
			&\multicolumn{2}{c|}{\textbf{Training}}&\multicolumn{2}{c|}{\textbf{Validation}}&\multicolumn{2}{c}{\textbf{Test}}\\
			\cmidrule(lr){2-3}\cmidrule(lr){4-5}\cmidrule(lr){6-7}
			&\textit{\# nodes}&\textit{\# edges}&\textit{\# nodes}&\textit{\# edges}&\textit{\# nodes}&\textit{\# edges}\\
			\midrule
			Reddit & 10,844  & 470,713  & 9,839  & 100,867  & 9,615  & 100,867   \\ 
			Wikipedia & 7,475  & 110,232  & 3,256  & 23,621  & 3,564  & 23,621   \\ 
			MOOC & 6,625  & 288,224  & 2,599  & 61,762  & 2,412  & 61,763  \\ 
			\bottomrule
		\end{tabular}
		
	\end{table}

	EdgeSampler of link prediction pipeline introduced in Section \apc{3.2.1} uses fixed seeds for different validation sets and test sets to ensure that the test results are reproducible across different runs. 
	
	\begin{table*}[!b]
		\begin{minipage}[t]{0.45\textwidth}
			\centering
			\caption{Experimental parameters of TGAT.} 
			\label{tab:tgat-parameters}
			\begin{tabular}{l|lllll}
				\toprule
				&$d_{n}$&$d_{e}$&$d_{time}$&$d_{pos}$&$n_{head}$\\
				\midrule
				Reddit & 172 & 172 & 172 & 172 & 2  \\ 
				Wikipedia & 172 & 172 & 172 & 172 & 2  \\ 
				MOOC & 172 & 4 & 172 & 172 & 2  \\ 
				LastFM & 172 & 2 & 172 & 172 & 2  \\ 
				Enron & 172 & 32 & 172 & 172 & 2  \\ 
				SocialEvo & 172 & 2 & 172 & 172 & 2  \\ 
				UCI & 172 & 100 & 172 & 172 & 2  \\ 
				CollegeMsg & 172 & 172 & 172 & 172 & 2  \\ 
				CanParl & 172 & 1 & 172 & 172 & 1  \\ 
				Contact & 172 & 1 & 172 & 172 & 1  \\ 
				Flights & 172 & 1 & 172 & 172 & 1  \\ 
				UNTrade & 172 & 1 & 172 & 172 & 1  \\ 
				USLegis & 172 & 1 & 172 & 172 & 1  \\ 
				UNVote & 172 & 1 & 172 & 172 & 1 \\ 
				Taobao & 172 & 4 & 172 & 172 & 2 \\ 
				\bottomrule
			\end{tabular}
		\end{minipage}
		\hfill
		\begin{minipage}[t]{0.45\textwidth}
			\centering
			\caption{Experimental parameters of CAWN.}
			\label{tab:cawn-parameters}
			\begin{tabular}{l|llll}
				\toprule
				&$d_{n}$&$d_{e}$&$d_{time}$&$d_{pos}$\\
				\midrule
				Reddit & 172 & 172 & 172 & 108  \\ 
				Wikipedia & 172 & 172 & 172 & 108  \\ 
				MOOC & 172 & 4 & 172 & 100  \\ 
				LastFM & 172 & 2 & 172 & 102  \\ 
				Enron & 172 & 32 & 172 & 104  \\ 
				SocialEvo & 172 & 2 & 172 & 102  \\ 
				UCI & 172 & 100 & 172 & 100  \\ 
				CollegeMsg & 172 & 172 & 172 & 108  \\ 
				CanParl & 172 & 1 & 172 & 103  \\ 
				Contact & 172 & 1 & 172 & 103  \\ 
				Flights & 172 & 1 & 172 & 103  \\ 
				UNTrade & 172 & 1 & 172 & 103  \\ 
				USLegis & 172 & 1 & 172 & 103  \\ 
				UNVote & 172 & 1 & 172 & 103 \\ 
				Taobao & 172 & 4 & 172 & 100 \\ 
				\bottomrule
			\end{tabular}
		\end{minipage}
	\end{table*}


	\section{Model Implementation Details}
	\label{appendix:ModelImplementation}
	
	We implement JODIE, DyRep, and TGN based on the \href{https://github.com/twitter-research/tgn}{TGN} framework.
	Furthermore, we fix the inconsistencies of implementations between link prediction task and node classification task. 
	
	\href{https://github.com/StatsDLMathsRecomSys/Inductive-representation-learning-on-temporal-graphs}{TGAT} concatenates \textit{node features}, \textit{edge features}, \textit{time features}, and \textit{position features} to perform the multi-head self-attention mechanism. 
	There is a positional encoding in the self-attention mechanism for capturing sequential information.
	Let $d_{n}$, $d_{e}$, $d_{time}$, and $d_{pos}$ denote the dimensions of node features, edge features, time features, and positional encoding, respectively. 
	The number of attention heads is $n_{head}$.
	These parameters must satisfy:
	\begin{equation}
		\label{eq:tgatdimensions}
		(d_{n} + d_{e} + d_{time} + d_{pos}) \% n_{head}=0
	\end{equation}
	
	The experimental parameters of TGAT are summarized in Table \ref{tab:tgat-parameters}.

	Similar to the setup of TGAT, \href{https://github.com/snap-stanford/CAW}{CAWN} adopts a multi-head self-attention mechanism to capture the subtle relevance of \textit{node features}, \textit{edge features}, \textit{time features}, and \textit{positional features}. 
	Those parameters satisfy Formula \eqref{eq:tgatdimensions} as well, and $d_{n}=d_{time}$ . 
	However, CAWN initializes the number of attention heads to 2, so we change the dimension of $d_{pos}$ to conduct experiments. 
	The experimental parameters of CAWN are shown in Table \ref{tab:cawn-parameters}.

	
	
	\href{https://github.com/KimMeen/Neural-Temporal-Walks}{NeurTW} concatenates \textit{node features}, \textit{edge features}, and \textit{positional features} (without \textit{time features}) during the temporal random walk encoding. 
	Regarding the temporal walk sampling strategy, given a node $u$ at time $t$, the sampling probability weight of its neighbor $v$ ($(\{v, u\}, t^{\prime}) \in \mathcal{G}_{u, t}$) is proportion to $exp (\alpha (t^{\prime}-t))$, where $\alpha$ is a temporal bias. 
	This sampling strategy is a temporal-biased sampling method.
	However, the time intervals in some benchmark datasets (Enron, CanParl, UNTrade, USLegis, and UNVote) are relatively large, and the exponential sampling probability weights may encounter overflow. 
	Therefore, we propose a strategy to calculate the sampling probability weights for these datasets: 
	\begin{equation}
		W(v, t^{\prime}) =
		\begin{cases}
			t^{\prime}-t, & t^{\prime}-t > 0, \\
			1,   & t^{\prime}-t = 0, \\
			-1/(t^{\prime}-t), &  t^{\prime}-t < 0,
		\end{cases}
	\end{equation}
	where $W(v, t^{\prime}) > 0$. 
	This strategy can avoid overflow and is also a temporal-biased sampling method.
	Finally, the sampling probability of each neighbor is obtained after normalization:
	\begin{equation}
		Pr_t(v)=\frac{\alpha W(v, t_{v})} {\sum_{v^{\prime} \in \mathcal{G}_{u, t}} \alpha W(v^{\prime}, t_{v^{\prime}})},
	\end{equation}
	where $\alpha$ is a temporal bias. For other hyperparameters that we have not mentioned, we use default values from the original experiments in the corresponding papers. 
	All the experimental codes are publicly available under  \href{https://opensource.org/license/mit/}{MIT license} and can be accessed at \url{https://github.com/qianghuangwhu/benchtemp}.

	\section{Experiment Results}
	\begin{table}[!t]
		\tiny
		\setlength{\tabcolsep}{4.5pt}
		\centering
		\caption{\footnotesize Average Precision (AP) results on link prediction task. 
			"*" denotes that the model encounters runtime error; 
			"---" denotes timeout after 48 hours. 
			The best and second-best results are highlighted as \fv{bold red} and \sv{underlined blue}.
			Some standard deviations are zero because we terminate those models that can only run one epoch within 2 days.
			We do not highlight the second-best if the gap is $>0.05$ compared with the best result.} 
		\label{tab:ap-lp}
		\begin{tabular}{l|lllllll}
			\toprule
			&\multicolumn{7}{|c}{\textbf{Transductive}}\\
			\midrule
			\diagbox{Dataset}{Model} &JODIE&DyRep&TGN&TGAT&CAWN&NeurTW&NAT\\
			\midrule
			Reddit & 0.9718 ± 0.0022 & 0.9808 ± 0.0006 & \sv{0.9874 ± 0.0002} & 0.9822 ± 0.0003 & \fv{0.9904 ± 0.0001} & 0.9855 ± 0.0013 & 0.9868 ± 0.0017  \\ 
			Wikipedia & 0.9471 ± 0.0056 & 0.9464 ± 0.0010 & 0.9852 ± 0.0003 & 0.9536 ± 0.0022 & \sv{0.9906 ± 0.0001} & \fv{0.9918 ± 0.0001} & 0.9819 ± 0.0026  \\ 
			MOOC & 0.7364 ± 0.0370 & 0.7933 ± 0.0348 & \sv{0.883 ± 0.0242} & 0.7185 ± 0.0051 & \fv{0.9369 ± 0.0009} & 0.7943 ± 0.0248 & 0.7537 ± 0.0191  \\ 
			LastFM & 0.6762 ± 0.0678 & 0.6736 ± 0.0768 & 0.7694 ± 0.0276 & 0.5375 ± 0.0044 & \fv{0.8946 ± 0.0006} & 0.8405 ± 0.0 & \sv{0.8729 ± 0.0022}  \\ 
			Enron & 0.7841 ± 0.0254 & 0.7648 ± 0.0418 & 0.8472 ± 0.0173 & 0.6063 ± 0.0194 & \fv{0.9142 ± 0.0052} & 0.8847 ± 0.0079 & \sv{0.9044 ± 0.0036}  \\ 
			SocialEvo & 0.7982 ± 0.0476 & 0.8816 ± 0.0042 & \fv{0.9325 ± 0.0006} & 0.7724 ± 0.0052 & \sv{0.9188 ± 0.0011} & --- & 0.8989 ± 0.0096  \\ 
			UCI & 0.8436 ± 0.0110 & 0.4913 ± 0.0367 & 0.8914 ± 0.0138 & 0.779 ± 0.0052 & \sv{0.9425 ± 0.001} & \fv{0.9702 ± 0.0021} & 0.9253 ± 0.0083  \\ 
			CollegeMsg & 0.5276 ± 0.0493 & 0.5070 ± 0.0049 & 0.8418 ± 0.0847 & 0.7902 ± 0.0033 & \sv{0.9401 ± 0.0025} & \fv{0.9727 ± 0.0001} & 0.9241 ± 0.0086  \\ 
			CanParl & 0.7030 ± 0.0077 & 0.6860 ± 0.0256 & 0.6765 ± 0.0615 & 0.6811 ± 0.0157 & 0.6916 ± 0.0546 & \fv{0.8528 ± 0.0213} & 0.6593 ± 0.0764  \\ 
			Contact & 0.9087 ± 0.0114 & 0.9016 ± 0.0319 & 0.9699 ± 0.0045 & 0.5888 ± 0.0065 & \sv{0.9677 ± 0.0024} & \fv{0.9756 ± 0.0} & 0.945 ± 0.0168  \\ 
			Flights & 0.9389 ± 0.0075 & 0.8836 ± 0.0078 & \sv{0.9764 ± 0.0025} & 0.899 ± 0.0025 & \fv{0.9860 ± 0.0002} & 0.9321 ± 0.0 & 0.9749 ± 0.0048  \\ 
			UNTrade & 0.6329 ± 0.0102 & 0.6099 ± 0.0057 & 0.6059 ± 0.0086 & * & \sv{0.7488 ± 0.0005} & 0.5648 ± 0.0167 & \fv{0.7514 ± 0.0615}  \\ 
			USLegis & 0.7585 ± 0.0032 & 0.6808 ± 0.0368 & 0.7398 ± 0.0027 & 0.7206 ± 0.0071 & \sv{0.9682 ± 0.0048} & \fv{0.9713 ± 0.0013} & 0.7425 ± 0.016  \\ 
			UNVote & 0.6090 ± 0.0076 & 0.5855 ± 0.0225 & \fv{0.6694 ± 0.0095} & 0.5388 ± 0.002 & 0.6175 ± 0.0013 & 0.6008 ± 0.0 & \sv{0.6449 ± 0.033} \\
			Taobao & 0.808 ± 0.0015 & 0.8074 ± 0.0014 & 0.8618 ± 0.0004 & 0.5508 ± 0.0093 & 0.7464 ± 0.0027 & \sv{0.8808 ± 0.0012} & \fv{0.8933 ± 0.0007} \\ 
			\midrule
			&\multicolumn{7}{|c}{\textbf{Inductive}}  \\
			\midrule
			\diagbox{Dataset}{Model} &JODIE&DyRep&TGN&TGAT&CAWN&NeurTW&NAT\\
			\midrule
			Reddit & 0.9427 ± 0.0118 & 0.9582 ± 0.0003 & 0.9767 ± 0.0003 & 0.9667 ± 0.0003 & \sv{0.9889 ± 0.0001} & 0.9821 ± 0.0006 & \fv{0.9912 ± 0.0027}  \\ 
			Wikipedia & 0.9316 ± 0.0049 & 0.9181 ± 0.0037 & 0.9791 ± 0.0004 & 0.9389 ± 0.0035 & 0.9903 ± 0.0002 & \sv{0.9912 ± 0.0004} & \fv{0.9962 ± 0.0021}  \\ 
			MOOC & 0.7282 ± 0.0686 & 0.7985 ± 0.0153 & 0.8726 ± 0.0267 & 0.7204 ± 0.0055 & \fv{0.9394 ± 0.0005} & 0.7903 ± 0.0307 & 0.7474 ± 0.0214  \\ 
			LastFM & 0.8057 ± 0.0424 & 0.7956 ± 0.0631 & 0.8261 ± 0.0145 & 0.5454 ± 0.0094 & \sv{0.9225 ± 0.0009} & 0.8842 ± 0.0 & \fv{0.9235 ± 0.0028}  \\ 
			Enron & 0.7640 ± 0.0310 & 0.6883 ± 0.0635 & 0.7982 ± 0.0237 & 0.5661 ± 0.0134 & \sv{0.916 ± 0.001} & 0.8940 ± 0.0025 & \fv{0.9308 ± 0.0085}  \\ 
			SocialEvo & 0.8527 ± 0.0303 & \sv{0.8954 ± 0.0034} & 0.8944 ± 0.0102 & 0.6497 ± 0.004 & \fv{0.9118 ± 0.0003} & --- & 0.8682 ± 0.0324  \\ 
			UCI & 0.7298 ± 0.0152 & 0.4606 ± 0.0209 & 0.8306 ± 0.0177 & 0.704 ± 0.0046 & 0.9421 ± 0.0012 & \fv{0.9720 ± 0.0024} & \sv{0.9658 ± 0.0125}  \\ 
			CollegeMsg & 0.4960 ± 0.0193 & 0.4858 ± 0.0051 & 0.7983 ± 0.049 & 0.7184 ± 0.0014 & 0.941 ± 0.0026 & \fv{0.9762 ± 0.0} & \sv{0.9642 ± 0.0124}  \\ 
			CanParl & 0.5148 ± 0.0119 & 0.5365 ± 0.0064 & 0.5596 ± 0.0141 & 0.5814 ± 0.0041 & 0.6915 ± 0.0578 & \fv{0.8469 ± 0.0161} & 0.6058 ± 0.0812  \\ 
			Contact & 0.9162 ± 0.0051 & 0.8334 ± 0.0620 & 0.9411 ± 0.0071 & 0.5922 ± 0.0056 & \sv{0.9688 ± 0.0023} & \fv{0.9762 ± 0.0} & 0.9489 ± 0.0091  \\ 
			Flights & 0.9190 ± 0.0081 & 0.8707 ± 0.0121 & 0.9439 ± 0.0043 & 0.8361 ± 0.0039 & \fv{0.9834 ± 0.0002} & 0.9201 ± 0.0 & \sv{0.9817 ± 0.0026}  \\ 
			UNTrade & 0.6392 ± 0.0132 & 0.6232 ± 0.0188 & 0.5603 ± 0.0106 & * & \fv{0.7361 ± 0.0009} & 0.5640 ± 0.0137 & 0.6586 ± 0.0543  \\ 
			USLegis & 0.5557 ± 0.0107 & 0.5687 ± 0.0008 & 0.6048 ± 0.0047 & 0.5637 ± 0.0048 & \sv{0.9694 ± 0.0028} & \fv{0.971 ± 0.0009} & 0.6946 ± 0.0198  \\ 
			UNVote & 0.5242 ± 0.0050 & 0.5118 ± 0.0037 & 0.5702 ± 0.0099 & 0.5204 ± 0.004 & 0.6014 ± 0.0013 & 0.6025 ± 0.0 & \fv{0.7637 ± 0.0023} \\ 
			Taobao & 0.6696 ± 0.0025 & 0.6717 ± 0.0006 & 0.6761 ± 0.0015 & 0.5293 ± 0.0096 & 0.7389 ± 0.0026 & 0.8815 ± 0.0045 & \fv{0.9992 ± 0.0001} \\ 
			\midrule
			&\multicolumn{7}{|c}{\textbf{Inductive New-Old}}  \\
			\midrule
			\diagbox{Dataset}{Model} &JODIE&DyRep&TGN&TGAT&CAWN&NeurTW&NAT\\
			\midrule
			Reddit & 0.9399 ± 0.0112 & 0.9552 ± 0.0027 & 0.9749 ± 0.0006 & 0.9659 ± 0.0004 & \sv{0.9871 ± 0.0003} & 0.9810 ± 0.0015 & \fv{0.9947 ± 0.0014}  \\ 
			Wikipedia & 0.9127 ± 0.0078 & 0.8947 ± 0.0040 & 0.9724 ± 0.0008 & 0.9223 ± 0.0021 & \sv{0.9901 ± 0.0002} & 0.9884 ± 0.0007 & \fv{0.9959 ± 0.0018}  \\ 
			MOOC & 0.7366 ± 0.5977 & 0.8011 ± 0.0092 & 0.8669 ± 0.0328 & 0.7263 ± 0.0059 & \fv{0.9408 ± 0.0022} & 0.7907 ± 0.0336 & 0.7677 ± 0.0175  \\ 
			LastFM & 0.7448 ± 0.0034 & 0.7024 ± 0.0532 & 0.7661 ± 0.0232 & 0.5447 ± 0.0023 & \sv{0.8906 ± 0.0021} & 0.835 ± 0.0 & \fv{0.9235 ± 0.001}  \\ 
			Enron & 0.7526 ± 0.0158 & 0.6742 ± 0.0632 & 0.7918 ± 0.0209 & 0.5729 ± 0.0189 & \sv{0.9168 ± 0.0061} & 0.8925 ± 0.0090 & \fv{0.9319 ± 0.0092}  \\ 
			SocialEvo & 0.8521 ± 0.0403 & \fv{0.8999 ± 0.0046} & \sv{0.8972 ± 0.0107} & 0.6578 ± 0.0041 & 0.8830 ± 0.0008 & --- & 0.8437 ± 0.0569  \\ 
			UCI & 0.6891 ± 0.0166 & 0.4574 ± 0.0207 & 0.8243 ± 0.0205 & 0.6826 ± 0.0084 & 0.9414 ± 0.002 & \sv{0.9732 ± 0.0040} & \fv{0.9768 ± 0.0127}  \\ 
			CollegeMsg & 0.5000 ± 0.0227 & 0.4834 ± 0.0177 & 0.7954 ± 0.0349 & 0.701 ± 0.0058 & 0.9407 ± 0.0017 & \sv{0.9719 ± 0.0014} & \fv{0.9763 ± 0.0133}  \\ 
			CanParl & 0.5143 ± 0.0043 & 0.5168 ± 0.0170 & 0.552 ± 0.0135 & 0.574 ± 0.0054 & 0.6952 ± 0.0518 & \fv{0.8417 ± 0.0132} & 0.6027 ± 0.0787  \\ 
			Contact & 0.9150 ± 0.0058 & 0.8253 ± 0.0637 & 0.9421 ± 0.0055 & 0.5915 ± 0.0049 & \sv{0.9689 ± 0.0029} & \fv{0.9757 ± 0.0} & 0.9384 ± 0.0175  \\ 
			Flights & 0.9128 ± 0.0095 & 0.8657 ± 0.0117 & 0.9412 ± 0.0039 & 0.833 ± 0.0031 & \sv{0.9827 ± 0.0002} & 0.9161 ± 0.0 & \fv{0.9845 ± 0.0033}  \\ 
			UNTrade & 0.6333 ± 0.0102 & 0.6101 ± 0.0196 & 0.5622 ± 0.014 & * & \fv{0.7375 ± 0.001} & 0.5692 ± 0.0185 & 0.5844 ± 0.053  \\ 
			USLegis & 0.5567 ± 0.0106 & 0.5490 ± 0.0143 & 0.5651 ± 0.0131 & 0.5695 ± 0.0099 & \fv{0.9703 ± 0.0027} & \sv{0.9671 ± 0.0027} & 0.5024 ± 0.0511  \\ 
			UNVote & 0.5348 ± 0.0072 & 0.5126 ± 0.0103 & 0.5724 ± 0.0107 & 0.5196 ± 0.0022 & 0.6050 ± 0.0019 & 0.6036 ± 0.0 & \fv{0.7598 ± 0.0167} \\ 
			Taobao & 0.6838 ± 0.0045 & 0.6884 ± 0.0013 & 0.6944 ± 0.0038 & 0.5309 ± 0.0189 & 0.7374 ± 0.0032 & 0.8687 ± 0.0010 & \fv{0.9997 ± 0.0001} \\ 
			\midrule
			&\multicolumn{7}{|c}{\textbf{Inductive New-New}}  \\
			\midrule
			\diagbox{Dataset}{Model} &JODIE&DyRep&TGN&TGAT&CAWN&NeurTW&NAT\\
			\midrule
			Reddit & 0.9199 ± 0.0167 & 0.9384 ± 0.0064 & 0.9727 ± 0.0004 & 0.9523 ± 0.0056 & \fv{0.9958 ± 0.0017} & 0.9890 ± 0.0003 & \sv{0.9951 ± 0.0005}  \\ 
			Wikipedia & 0.9307 ± 0.0060 & 0.9329 ± 0.0028 & 0.9822 ± 0.0009 & 0.9592 ± 0.0039 & 0.9941 ± 0.0004 & \sv{0.9963 ± 0.0003} & \fv{0.9979 ± 0.0009}  \\ 
			MOOC & 0.6623 ± 0.0189 & 0.7135 ± 0.0148 & 0.8651 ± 0.0059 & 0.7239 ± 0.0052 & \fv{0.935 ± 0.0009} & 0.7871 ± 0.0221 & 0.6654 ± 0.0155  \\ 
			LastFM & 0.8558 ± 0.0110 & 0.8388 ± 0.0209 & 0.8121 ± 0.0046 & 0.536 ± 0.0217 & \sv{0.9716 ± 0.0008} & 0.9585 ± 0.0 & \fv{0.9722 ± 0.0013}  \\ 
			Enron & 0.6525 ± 0.0146 & 0.6312 ± 0.0449 & 0.7391 ± 0.0196 & 0.538 ± 0.0093 & \fv{0.9556 ± 0.0055} & 0.9358 ± 0.0008 & \sv{0.9503 ± 0.0079}  \\ 
			SocialEvo & 0.5958 ± 0.0391 & 0.7312 ± 0.0103 & 0.8268 ± 0.003 & 0.5096 ± 0.0097 & \fv{0.9150 ± 0.0013} & --- & \sv{0.9112 ± 0.0563}  \\ 
			UCI & 0.6249 ± 0.0198 & 0.5062 ± 0.0032 & 0.8393 ± 0.0155 & 0.7758 ± 0.0033 & 0.9488 ± 0.0012 & \fv{0.9736 ± 0.0008} & \sv{0.9518 ± 0.0211}  \\ 
			CollegeMsg & 0.5212 ± 0.0244 & 0.5328 ± 0.0117 & 0.8244 ± 0.0098 & 0.7929 ± 0.0029 & 0.9484 ± 0.0039 & \fv{0.9797 ± 0.0008} & \sv{0.95 ± 0.0257}  \\ 
			CanParl & 0.4697 ± 0.0043 & 0.4794 ± 0.0057 & 0.5553 ± 0.0258 & 0.6004 ± 0.0087 & 0.6671 ± 0.0795 & \fv{0.8511 ± 0.0079} & 0.5989 ± 0.0571  \\ 
			Contact & 0.7381 ± 0.0145 & 0.6601 ± 0.0432 & 0.8916 ± 0.0075 & 0.5779 ± 0.0044 & \sv{0.9670 ± 0.0031} & \fv{0.9704 ± 0.0} & 0.9535 ± 0.0044  \\ 
			Flights & 0.9250 ± 0.0065 & 0.6312 ± 0.0449 & 0.9644 ± 0.0015 & 0.8608 ± 0.0049 & \sv{0.9882 ± 0.0009} & 0.9496 ± 0.0 & \fv{0.9906 ± 0.0009}  \\ 
			UNTrade & 0.5801 ± 0.0112 & 0.5344 ± 0.0130 & 0.5164 ± 0.0056 & * & \fv{0.7404 ± 0.0023} & 0.5685 ± 0.0298 & 0.6785 ± 0.0289  \\ 
			USLegis & 0.5250 ± 0.0045 & 0.5523 ± 0.0127 & 0.5582 ± 0.02 & 0.5434 ± 0.0203 & \sv{0.9767 ± 0.0055} & \fv{0.9803 ± 0.0005} & 0.8627 ± 0.0196  \\ 
			UNVote & 0.4973 ± 0.0145 & 0.4856 ± 0.0078 & 0.5502 ± 0.0096 & 0.5337 ± 0.0046 & 0.5830 ± 0.0076 & 0.5964 ± 0.0 & \fv{0.7549 ± 0.035} \\
			Taobao & 0.6764 ± 0.0013 & 0.676 ± 0.0011 & 0.6739 ± 0.0016 & 0.5222 ± 0.0033 & 0.7390 ± 0.0147 & 0.9025 ± 0.0035 & \fv{0.9997 ± 0.0001} \\
			\bottomrule
		\end{tabular}
		
	\end{table}

	\subsection{AP Results for Link Prediction}
	\label{appendix:ap-lp}
	
	We show the average precision (AP) results on link prediction task and highlight the best and second-best numbers for each job in Table \ref{tab:ap-lp}. The overall performance is similar to that of AUC. For the transductive setting, CAWN  gives impressive results and achieves the best or second-best results 
	on 12 datasets out of 15, followed by NeurTW (7 out of 15), TGN (5 out of 15), and NAT (5 out of 15), verifying the effectiveness of temporal walk, temporal memory, and joint neighborhood  on transductive link prediction task. 
	For the inductive setting, CAWN and NAT both rank top-2 on 9 datasets, followed by NeurTW on 6. 
	Results reveal that models based on temporal walks and joint neighborhood can better capture structure patterns on edges that have never been seen. 
	TGN performs relatively poorly for the inductive link prediction task on almost all datasets.
	We can draw similar conclusions from inductive New-Old and inductive New-Old experimental results. 
	We note that DyRep achieves the best AP result under the inductive New-Old setting and the second-best AP result under the inductive setting on the SocialEvo dataset. 
	SocialEvo has the maximum average degree and edge density as shown in Table \apc{2} in the main paper, demonstrating that DyRep performs better for inductive link prediction tasks on dense temporal graphs.

	\subsection{GPU Utilization Comparison for Link Prediction}
	\label{appendix:gpuutilization-lp}
	
	In Table~\ref{tab:gpuutilization-lp}, we report the GPU utilization results on the link prediction task. 
	NAT obtains the best or second-best results on 13 datasets out of 15, followed by TGN (9 out of 15). 
	The data structure, called \textit{N-cache}, designed in NAT supports parallel access and updates of \textit{dictionary-type neighborhood representation} on GPUs.
	Therefore, NAT achieves the best performance regarding GPU utilization.
	TGN proposes a highly efficient parallel processing strategy to handle temporal graph, so that TGN has the second-best performance on GPU utilization.

	\begin{table}[!t]
		\footnotesize
		\setlength{\tabcolsep}{2pt}
		\centering
		\caption{GPU utilization of models on link prediction task. 
			"*" denotes that TGAT layer cannot find suitable neighbors within given time interval and encounters error. The best and second-best results are highlighted as \fv{bold red} and \sv{underlined blue}. }
		\label{tab:gpuutilization-lp}
		\begin{tabular}{l|rrrrrrr}
			\toprule
			&\multicolumn{7}{|c}{\textbf{GPU Utilization} (\%)}\\
			\midrule
			\diagbox{Dataset}{Model} &JODIE&DyRep&TGN&TGAT&CAWN&NeurTW&NAT\\
			\midrule
			Reddit & 21 & 22 & \fv{41} & 39 & 26 & 31 & \sv{40}  \\ 
			Wikipedia & 34 & \fv{46} & 36 & 35 & 28 & 17 & \sv{38}  \\ 
			MOOC & 14 & 29 & \sv{35} & \sv{35} & 18 & 14 & \fv{45}  \\ 
			LastFM & 22 & 28 & \sv{38} & 22 & 23 & 22 & \fv{48}  \\ 
			Enron & 18 & 24 & \sv{41} & 24 & 25 & 22 & \fv{51}  \\ 
			SocialEvo & 24 & 25 & \sv{42} & 25 & 17 & 22 & \fv{46}  \\ 
			UCI & 30 & 25 & 35 & 33 & 27 & \fv{58} & \sv{44}  \\ 
			CollegeMsg & 21 & 32 & 46 & \sv{47} & 25 & 34 & \fv{48}  \\ 
			CanParl & 26 & 27 & \fv{54} & 47 & 24 & 22 & \sv{51}  \\ 
			Contact & 25 & 22 & \sv{40} & 29 & 19 & 22 & \fv{50}  \\ 
			Flights & 20 & 20 & 26 & \sv{35} & 18 & 24 & \fv{43}  \\ 
			UNTrade & 19 & 23 & \sv{50} & * & 15 & 23 & \fv{53}  \\ 
			USLegis & 22 & 28 & \sv{54} & 38 & 26 & \fv{55} & 53  \\ 
			UNVote & 12 & 22 & 37 & 35 & 16 & \sv{46} & \fv{54}  \\ 
			Taobao & 29 & \fv{56} & 31 & \sv{55} & 22 & 53 & 38 \\ 
			\bottomrule
		\end{tabular}
	\end{table}
	
	\subsection{Efficiency Comparison of Node Classification Task}
	\label{appendix:efficiency-nc}

	\begin{table}[!t]
		\tiny
		\setlength{\tabcolsep}{2pt}
		\centering
		\caption{\footnotesize Model efficiency on the node classification task. 
			We report seconds per epoch as \textbf{Runtime}, 
			the averaged number of epochs for convergence before early stopping as \textbf{Epoch}, the maximum RAM usage as \textbf{RAM}, the maximum GPU memory usage  as \textbf{GPU Memory}, and the maximum GPU utilization usage  as \textbf{GPU Utilization}, respectively. "x" indicates that the model cannot converge within 48 hours. The best and second-best results are highlighted as \fv{bold red} and \sv{underlined blue}.}
		\label{tab:nc-efficiency}
		\begin{tabular}{l|rrrrrrr|rrrrrrr}
			\toprule
			&\multicolumn{7}{|c|}{\textbf{Runtime} (second)}&\multicolumn{7}{|c}{\textbf{Epoch}}\\
			\midrule
			\diagbox{Dataset}{Model} &JODIE&DyRep&TGN&TGAT&CAWN&NeurTW&NAT&JODIE&DyRep&TGN&TGAT&CAWN&NeurTW&NAT\\
			\midrule
			Reddit  & 65.98  & 75.07  & 73.99  & \fv{16.99}  & 1,913.83  & 24,016.13  & \sv{28.24}  & \sv{6} & 7 & 8 & \fv{4} & x & x & \sv{6}  \\ 
			Wikipedia  & 15.75  & 14.55  & 14.58  & \fv{4.95}  & 351.54  & 2,723.37  & \sv{6.39}  & \sv{4} & 14 & \fv{3} & 7 & 6 & \sv{4} & \fv{3}  \\ 
			MOOC & 28.57  & 32.91  & 27.95  & \fv{8.51}  & 1,146.76  & 7,466.04  & \sv{17.55}  & 8 & \fv{3} & \sv{5} & 7 & 10 & x & \fv{3} \\ 
			\midrule
			&\multicolumn{7}{|c|}{\textbf{RAM} (GB)}&\multicolumn{7}{|c}{\textbf{GPU Memory} (GB)}\\
			\midrule
			Reddit  & 3.8 & \sv{3.4} & \fv{3.3} & 4.1 & 45.7 & 16.7 & \fv{3.3} & \sv{2.9} & \fv{2.8} & 3.1 & 3.0 & 3.7 & 3.0 & 3.0  \\ 
			Wikipedia  & \fv{2.6} & \fv{2.6} & \fv{2.6} & \sv{3.1} & 8.2 & 8.2 & \fv{2.6} & 2.6 & \sv{2.4} & 2.8 & 2.5 & 3.1 & 2.5 & \fv{1.7}  \\ 
			MOOC & \sv{2.9} & \sv{2.9} & \sv{2.9} & 3.1 & 45.1 & 32.6 & \fv{2.5} & 1.9 & 1.8 & 1.9 & 2.1 & 2.3 & \sv{1.7} & \fv{1.3} \\
			\midrule
			&\multicolumn{7}{|c|}{\textbf{\textbf{GPU Utilization}} (\%)}&\multicolumn{7}{|c}{}\\
			\cmidrule{1-8}
			Reddit  & 41 & 40 & \sv{42} & 38 & 18 & 36 & \fv{52}  & ~ & ~ & ~ & ~ & ~ & ~ & ~ \\ 
			Wikipedia  & 37 & 30 & 41 & 31 & 25 & \fv{80} & \sv{48}  & ~ & ~ & ~ & ~ & ~ & ~ & ~ \\ 
			MOOC & 35 & 33 & 40 & 30 & 13 & \sv{56} & \fv{62} & ~ & ~ & ~ & ~ & ~ & ~ & ~ \\ 
			\bottomrule
		\end{tabular}
		
	\end{table}
	
	We compare the efficiency results on node classification task and show the results in Table \ref{tab:nc-efficiency}.
	Regarding runtime per epoch, TGAT achieves the fastest performance on all three datasets, followed by NAT. 
	Similar to the runtime results on the link prediction task, the training process of CAWN and NeurTW is much slower due to the inefficient temporal walk. As for the averaged number of epochs for convergence, NAT ranks top-2 on all three datasets, followed by JODIE (2 out of 3), TGN (2 out of 3).
	RAM results reveal that CAWN and NeurTW consume much more memory due to the temporal walk and complex sampling strategy. 
	Most models need 1 - 3 GB of GPU memory, similar to the link prediction task. 
	Due to the parallel access and update of representation on GPU, NAT achieves the highest GPU utilization.

	
	

	\begin{figure*}[!t]
		\centering
		\includegraphics[width=\textwidth]{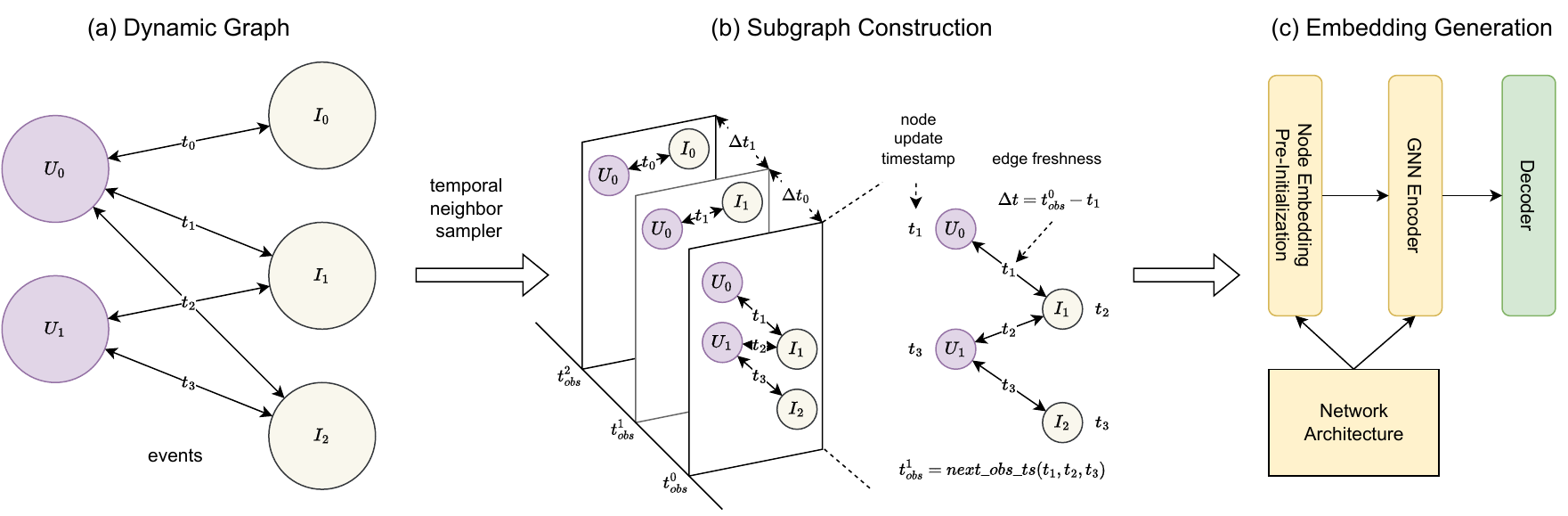}
		\caption{Workflow of TeMP. $U$ denotes user, and $I$ denotes item.}
		\label{fig:temp-workflow}
	\end{figure*}
	

	\section{Results of TeMP}
	\label{appendix:TeMP}
	
	Upon the anatomy of the existing methods, we propose a novel temporal graph neural network, called TeMP.
	As shown in Figure~\ref{fig:temp-workflow}, given a dynamic graph (a), the processing of TeMP is as follows:
	
	\begin{itemize}
		\item{\underline{\textit{Subgraph Construction (b)}}.}
		We construct a subgraph with a temporal neighbor sampler with intervals adaptive to data.
		We try to find a reference timestamp and sample a subgraph before this timestamp. 
		We have conducted experiments at various quantiles, and 
		chosen the mean timestamp since it obtains the overall best performance.
		
		\item{\underline{\textit{Embedding Generation (c)}.}}
		Upon subgraph construction, TeMP generates temporal embeddings for nodes and edges.
		The model architecture consists of three main components: temporal label propagation (LPA), message-passing operators, and a sequence updater. The temporal LPA captures the motif pattern, while the message-passing operators aggregate the original edge features. 
		The sequence updater chooses RNN to update the embeddings with a memory module.
		Furthermore, TeMP uses a pre-initialization strategy to generate initial temporal node embeddings.
	\end{itemize}

	\begin{table}[!t]
		\tiny
		\setlength{\tabcolsep}{2pt}
		\centering
		\caption{AUC and AP results of TeMP on link prediction task. 
			We highlight the numbers as \fv{bold red} and \sv{underlined blue} if TeMP achieves the best and second-best results compared to the TGNN models in the main paper.}
		\label{tab:temp-lp}
		\begin{tabular}{l|lllll}
			\toprule
			&\multicolumn{4}{|c}{\textbf{AUC}} \\
			\midrule
			&Transductive&Inductive&Inductive New-Old&Inductive New-New\\
			\midrule
			Reddit & \fv{0.99 ± 0.0001} & 0.9843 ± 0.0002 & 0.9818 ± 0.0001 & 0.9843 ± 0.0003  \\ 
			Wikipedia & 0.9801 ± 0.0005 & 0.9669 ± 0.0008 & 0.9504 ± 0.0005 & 0.97 ± 0.0013  \\ 
			MOOC & 0.8249 ± 0.0027 & 0.8277 ± 0.0036 & 0.832 ± 0.0031 & 0.7621 ± 0.0086  \\ 
			LastFM & \sv{0.865 ± 0.0012} & 0.8879 ± 0.0015 & 0.8333 ± 0.0014 & 0.9228 ± 0.0007  \\ 
			Enron & 0.8717 ± 0.0122 & 0.8485 ± 0.0153 & 0.8491 ± 0.009 & 0.7798 ± 0.0329  \\ 
			SocialEvo & 0.9303 ± 0.0005 & 0.9039 ± 0.0005 & 0.9017 ± 0.0013 & 0.8485 ± 0.0041  \\ 
			UCI & 0.8925 ± 0.001 & 0.7955 ± 0.0054 & 0.7787 ± 0.0032 & 0.7318 ± 0.0017  \\ 
			CollegeMsg & 0.8873 ± 0.0018 & 0.8027 ± 0.0029 & 0.7848 ± 0.0022 & 0.7342 ± 0.0047  \\ 
			CanParl & 0.7801 ± 0.0139 & 0.5414 ± 0.0176 & 0.5313 ± 0.0253 & 0.5683 ± 0.0293  \\ 
			Contact & 0.958 ± 0.0013 & 0.9474 ± 0.0015 & 0.9474 ± 0.0011 & 0.7835 ± 0.0041  \\ 
			Flights & \fv{0.987 ± 0.0003} &  0.9708 ± 0.0006 & 0.9692 ± 0.0006 & 0.973 ± 0.0009  \\ 
			UNTrade & 0.6011 ± 0.0009 & 0.5732 ± 0.0006 & 0.5703 ± 0.0006 & 0.5539 ± 0.0008  \\ 
			USLegis & 0.7073 ± 0.0114 & 0.5493 ± 0.0183 & 0.5609 ± 0.0123 & 0.5425 ± 0.0234  \\ 
			UNVote & 0.5571 ± 0.0053 & 0.5419 ± 0.006 & 0.5409 ± 0.0066 & 0.5696 ± 0.0038  \\ 
			Taobao & \fv{0.939 ± 0.0005} & 0.8513 ± 0.0018 & 0.8249 ± 0.003 & 0.8502 ± 0.0025 \\
			\midrule
			&\multicolumn{4}{|c}{\textbf{AP}} \\
			\midrule
			&Transductive&Inductive&Inductive New-Old&Inductive New-New\\
			\midrule
			
			Reddit & \fv{0.9904 ± 0.0001} & 0.9849 ± 0.0002 & 0.9828 ± 0.0002 & 0.9757 ± 0.0007  \\ 
			Wikipedia & 0.9817 ± 0.0004 & 0.9696 ± 0.0004 & 0.9562 ± 0.0005 & 0.9666 ± 0.0016  \\ 
			MOOC & 0.7935 ± 0.0033 & 0.7918 ± 0.0039 & 0.7989 ± 0.0035 & 0.7328 ± 0.0089  \\ 
			LastFM & 0.8717 ± 0.0014 & 0.8952 ± 0.0015 & 0.8468 ± 0.0015 & 0.8984 ± 0.0014  \\ 
			Enron & 0.85 ± 0.0141 & 0.8274 ± 0.0167 & 0.8324 ± 0.0104 & 0.757 ± 0.0307  \\ 
			SocialEvo & 0.9098 ± 0.0005 & 0.8767 ± 0.0004 & 0.8752 ± 0.0021 & 0.7907 ± 0.0056  \\ 
			UCI & 0.8968 ± 0.0007 & 0.8104 ± 0.0054 & 0.7969 ± 0.0047 & 0.7594 ± 0.0047  \\ 
			CollegeMsg & 0.8928 ± 0.0026 & 0.8206 ± 0.0041 & 0.8018 ± 0.003 & 0.7505 ± 0.0045  \\ 
			CanParl & 0.6871 ± 0.015 & 0.5388 ± 0.0074 & 0.5341 ± 0.0037 & 0.5554 ± 0.015  \\ 
			Contact & 0.9525 ± 0.0016 & 0.9429 ± 0.0019 & 0.944 ± 0.0012 & 0.7992 ± 0.0031  \\ 
			Flights & \sv{0.9857 ± 0.0003} & 0.9687 ± 0.0005 & 0.9661 ± 0.0006 & 0.9731 ± 0.001  \\ 
			UNTrade & 0.5855 ± 0.0007 & 0.564 ± 0.0007 & 0.5593 ± 0.0011 & 0.5634 ± 0.0008  \\ 
			USLegis & 0.6493 ± 0.0078 & 0.529 ± 0.01 & 0.5519 ± 0.0088 & 0.5375 ± 0.0218  \\ 
			UNVote & 0.5402 ± 0.0038 & 0.536 ± 0.005 & 0.5354 ± 0.0041 & 0.5481 ± 0.0052  \\ 
			Taobao & \fv{0.9385 ± 0.0004} & 0.8493 ± 0.0033 & 0.8243 ± 0.0049 & 0.8448 ± 0.0056 \\ 
			\bottomrule
		\end{tabular}
		
	\end{table}

	\begin{table}[!t]
		\tiny
		\setlength{\tabcolsep}{2pt}
		\centering
		\caption{Efficiency of TeMP on link prediction task. }\label{tab:temp-lp-efficiency}
		\begin{tabular}{l|rrrrr}
			\toprule
			&\multicolumn{5}{|c}{\textbf{Efficiency}} \\
			\midrule
			&Runtime (second)&Epoch&RAM (GB)&GPU Memory (GB)& GPU Utilization (\%)\\
			\midrule
			Reddit & 304.84  & 27 & 5.4 & \sv{2.3} & \fv{86}  \\ 
			Wikipedia & 51.00  & 22 & 3.4 & \sv{1.4}& \fv{71}  \\ 
			MOOC & 100.64  & 23 & 2.9 & \fv{1.3} & \fv{57}  \\ 
			LastFM & 471.25  & 48 & 3.6 & 1.4 & \fv{57}  \\ 
			Enron & 38.68  & 20 & 2.8 & \fv{1.1} & 31  \\ 
			SocialEvo & 670.08  & 30 & 3.77 & \fv{1.1} & 30  \\ 
			UCI & 43.98  & 26 & 2.9 & \fv{1.4} & \fv{50}  \\ 
			CollegeMsg & 11.55  & 30 & 3 & \fv{1.2} & \fv{49}  \\ 
			CanParl & 13.80  & 7 & 2.8 & \fv{1.1} & 37  \\ 
			Contact & 421.15  & 54 & 4.1 & \sv{1.3} & 13  \\ 
			Flights & 859.04  & 38 & 3.9 & \sv{1.4} & \fv{76}  \\ 
			UNTrade & 80.96  & 10 & 3 & 1.3 & 37  \\ 
			USLegis & 10.31  & 14 & 2.6 & \fv{1.1} & \fv{57}  \\ 
			UNVote & 165.32  & 12 & 3.3 & 1.3 & 39  \\ 
			Taobao & 17.78  & 16 & 3.1 & \sv{1.5} & 42 \\ 
			\bottomrule
		\end{tabular}
		
	\end{table}
	
	

	\begin{table}[!t]
		\tiny
		\setlength{\tabcolsep}{2pt}
		\centering
		\caption{AUC and efficiency results of TeMP on node classification task.}\label{tab:temp-nc}
		\begin{tabular}{l|c|rrrrr}
			\toprule
			&\multirow{2}{*}{AUC}&\multicolumn{5}{c}{\textbf{Efficiency}}\\
			\cmidrule{3-7}
			&&Runtime (second) &Epoch &RAM (GB) &GPU Memory (GB) &GPU Utilization (\%) \\
			\midrule
			Reddit  & \sv{0.6357 ± 0.0265} & 170.66 & 7 & 4.4 & 3.0 & \fv{66}  \\ 
			Wikipedia  & \fv{0.8873 ± 0.0078} & 24.18 & 13 & 3.5 & 1.4 & 47  \\ 
			MOOC & 0.6958 ± 0.0017 & 49.90 & 7 & 3.3 & 1.2 & 42 \\ 
			\bottomrule
		\end{tabular}
		
	\end{table}

	\subsection{Experimental Results}
	
	\textbf{Link Prediction.}
	The AUC and AP results of TeMP on link prediction task are presented in Table~\ref{tab:temp-lp}, and the efficiency results are shown in Table~\ref{tab:temp-lp-efficiency}.
	TeMP performs relatively well in the transductive setting, while
	lags behind CAWN, NeurTW, and NAT.
	Due to efficient subgraph sampling and parallel dataloader, TeMP outperforms other baselines regarding GPU memory and GPU utilization.
	
	\textbf{Node Classification.}
	The experimental results of TeMP on node classification task are presented in Table~\ref{tab:temp-nc}.
	TeMP achieves the best AUC on Wikipedia dataset and the second-best AUC on Reddit dataset, demonstrating that TeMP can effectively capture temporal evolution of nodes.
	Similarly, TeMP consumes relatively low GPU memory and can better utilize the computation power of GPU.

	

	\section{Newly Added Datasets}
	We have added \textit{six} datasets (eBay-Small, eBay-Large, Taobal-Large,~DGraphFin,~YouTubeReddit-Small,~YouTubeReddit-Large), including \textit{four} \textbf{large-scale} datasets (eBay-Large,~Taobao-Large,~DGraphFin,~YouTubeReddit-Large). The statistics of the new datasets are shown in Table \ref{tab:datasets}. For easy access, all datasets have been hosted on the open-source platform zenodo (\url{https://zenodo.org/}) with a Digital Object Identifier (DOI) 10.5281/zenodo.8267846 (\url{https://zenodo.org/record/8267846}).
	
	After a discussion with our industrial partner eBay, we are working on sharing the eBay-Small and eBay-Large datasets in a way that ensures availability and justifies the research purpose. eBay provide a Google form for the applicants:
	\url{https://forms.gle/bP1RmyVJ1C6pgyS66} (\textbf{the applicants can remain anonymous}).
	
	\begin{table}[!t]
		\centering
		\caption{\footnotesize Dataset statistics of the new datasets.}
		\label{tab:datasets}
		\begin{tabular}{llrr}
			\toprule
			~ & \textit{Domain} & \textit{\# Nodes} & \textit{\# Edges}  \\ \midrule
			eBay-Small & E-commerce & 38,427  & 384,677   \\ 
			YouTubeReddit-Small~\cite{jin2023predicting} & Social & 264,443  & 297,732   \\
			
			\midrule
			
			eBay-Large & E-commerce & 1,333,594   & 1,119,454 
			\\
			DGraphFin~\cite{huang2022dgraph} & E-commerce  & 3,700,550  & 4,300,999  \\ 
			Youtube-Reddit-Large~\cite{jin2023predicting} & Social  & 5,724,111& 4,228,523    \\
			Taobao-Large~\cite{jin2022neural, zhu2018learning} & E-commerce  & 1,630,453& 5,008,745   \\
			\bottomrule
		\end{tabular}
	\end{table}
	\begin{itemize}
		\item \textbf{eBay-Small} is a subset of the eBay-Large dataset. We sample 38,427 nodes and 384,677 edges from eBay-Large graph according to edge timestamps.
		\item \href{https://drive.google.com/drive/folders/1zgqbBHpQKt_RUJrIRISfZILUXLxPTx1I?usp=sharing}{\textbf{YouTubeReddit-Small }} is a  collection of massive visual contents on YouTube and long-term community activity on
		Reddit. This dataset covers a \textbf{3}-month period from
		January to March 2020. Each row in the dataset represents a YouTube video $v_{i}$
		being shared in a subreddit $s_{j}$
		by some user $u_{k}$
		at time $t$~\cite{jin2023predicting}. Nodes are YouTube videos and subreddits, edges are the users' interactions between videos and subreddits. This dynamic graph has 264,443 nodes and 297,732 edges.
		
		\item \textbf{eBay-Large} is a million-scale dataset consisting of 1.3 million nodes and 1.1 million edges, which comprises the selected transaction records from the eBay e-commerce platform over a two-month period. eBay-Large is modeled as a user-item graph, where items are heterogeneous entities which include information such as phone numbers, addresses, and email addresses associated with a transaction. We selecte one month of transactions as seed nodes and then expand each seed node two hops back in time to enrich the topology while maintaining consistency in the distribution of seed nodes.
		\item \href{https://dgraph.xinye.com/dataset}{\textbf{DGraphFin}} is a collection of large-scale dynamic graph datasets, consisting of interactive objects, events and labels that evolve with time.It is a directed, unweighted dynamic graph consisting of millions of nodes and edges, representing a realistic user-to-user social network in financial industry. Nodes are users, and an edge from one user to another means that the user regards the other user as the emergency contact person~\cite{huang2022dgraph}.
		\item \href{https://drive.google.com/drive/folders/1zgqbBHpQKt_RUJrIRISfZILUXLxPTx1I?usp=sharing}{\textbf{Youtube-Reddit-Large}} dataset
		covers \textbf{54} months of YouTube video propagation history from January 2018
		to June 2022~\cite{jin2023predicting}. This dataset has 5,724,111 nodes and 4,228,523 edges.
		\item \href{https://tianchi.aliyun.com/dataset/649}{\textbf{Taobao-Large}} is a collection  of the Taobao user behavior dataset intercepted based on the period 8:00 to 18:00 on 26 November 2017~\cite{zhu2018learning}. Nodes are users and items, and edges are behaviors between users and items, such as favor, click, purchase, and add an item to shopping cart. This public dataset has 1,630,453 nodes and 5,008,74 user-item interaction edges.
		
	\end{itemize}
	
	\subsection{Experiments}
	\label{sec:Experiments}
	We have conducted extensive experiments on the six newly added temporal graph datasets, including the dynamic link prediction task and dynamic node classification task, diverse workloads (transductive, inductive, inductive New-Old, and inductive New-New scenarios), and efficiency comparison (\textit{inference time}). We have added \textbf{\textit{Average Rank}} metric shown in Table \ref{tab:lp-auc} and Table \ref{tab:nc-auc} for ranking model performances on the newly added large-scale datasets for evaluating TGNN models. 
	
	\subsubsection{Link Prediction Task}
	
	We run the link prediction task on 7 TGNN models and the new datasets under different settings (Transductive, Inductive, Inductive New-Old, and Inductive New-New). The  AUC and AP results for each new datasets are shown in Table \ref{tab:lp-auc} and Table \ref{tab:lp-ap}, respectively. For the four large-scale datasets (eBay-Large, Taobao-Large, DGraphFin, YouTubeReddit-Large), we observe the similar results as in the paper.
	Specifically, NAT and NeurTW achieve the top-2 performance on almost all datasets under transductive  and inductive settings. 

	\begin{table}[!t]
		\tiny
		\setlength{\tabcolsep}{4.5pt}
		\centering
		\caption{\footnotesize ROC AUC results of new datasets on the \textit{dynamic link prediction task}. 
			The best and second-best results are highlighted as \fv{bold red} and \sv{underlined blue}. \textbf{Average Rank} are computed by the experimental results of models on four large-scale datasets (eBay-Large, Taobao-Large, DGraphFin, YouTubeReddit-Large).
			We do not highlight the second-best if the gap is $>0.05$ compared with the best result.}
		\label{tab:lp-auc}
		\begin{tabular}{l|lllllll}
			\toprule
			&\multicolumn{7}{|c}{\textbf{Transductive}}\\
			\midrule
			\diagbox{Dataset}{Model} &JODIE&DyRep&TGN&TGAT&CAWN&NeurTW&NAT\\
			\midrule
			eBay-Small & 0.9946 ± 0.0002 & 0.9941 ± 0.0006 & 0.9984 ± 0.0003 & 0.9838 ± 0.0006 & 0.9985 ± 0.0 & \fv{0.9991 ± 0.0} & \sv{0.9978 ± 0.0003}  \\ 
			YouTubeReddit-Small & \sv{0.8519 ± 0.0007} & 0.8499 ± 0.0012 & 0.8432 ± 0.0032 & 0.8441 ± 0.0014 & 0.7586 ± 0.0031 & \fv{0.9003 ± 0.0031} & 0.8259 ± 0.005  \\ 
			eBay-Large & 0.9614 ± 0.0 & 0.9619 ± 0.0001 & \sv{0.9642 ± 0.0003} & 0.5311 ± 0.0003 & 0.9442 ± 0.0003 & 0.9608 ± 0.0 & \fv{0.9658 ± 0.0002}  \\ 
			DGraphFin & 0.8165 ± 0.0024 & 0.8171 ± 0.0016 & \fv{0.8683 ± 0.0023} & 0.6112 ± 0.0165 & 0.5466 ± 0.0103 & \sv{0.8611 ± 0.0035} & 0.8258 ± 0.0001  \\ 
			Youtube-Reddit-Large & 0.8532 ± 0.0003 & 0.8529 ± 0.0006 & 0.8458 ± 0.0025 & 0.8536 ± 0.0026 & 0.7466 ± 0.0012 & \fv{0.916 ± 0.0025} & \sv{0.8605 ± 0.0009}  \\ 
			Taobao-Large & 0.7726 ± 0.0005 & 0.7724 ± 0.001 & \sv{0.8464 ± 0.0008} & 0.5567 ± 0.0047 & 0.7771 ± 0.0068 & \fv{0.859 ± 0.0091} & 0.8188 ± 0.001 \\ 
			\midrule
			\textbf{Average Rank} &4.5 & 4.5 & 2.75 & 5.75 & 6 & 2.25 & 2.25  \\ 
			\midrule
			&\multicolumn{7}{|c}{\textbf{Inductive}}  \\
			\midrule
			eBay-Small & 0.9696 ± 0.0007 & 0.9674 ± 0.0018 & 0.9913 ± 0.0004 & 0.9698 ± 0.0006 & 0.9964 ± 0.0001 & \sv{0.9982 ± 0.0} & \fv{0.9998 ± 0.0001}  \\ 
			YouTubeReddit-Small & 0.7582 ± 0.0003 & 0.7545 ± 0.0009 & 0.7276 ± 0.0033 & 0.7436 ± 0.0006 & 0.7533 ± 0.0016 & 0.8978 ± 0.0032 & \fv{0.9876 ± 0.0049}  \\ 
			eBay-Large & 0.7536 ± 0.0014 & 0.7515 ± 0.0006 & 0.7657 ± 0.0026 & 0.5224 ± 0.0003 & 0.9459 ± 0.0001 & \sv{0.9608 ± 0.0} & \fv{0.9999 ± 0.0001}  \\ 
			DGraphFin & 0.6884 ± 0.0051 & 0.6876 ± 0.001 & 0.6439 ± 0.0089 & 0.5677 ± 0.0184 & 0.5479 ± 0.009 & \fv{0.8635 ± 0.0021} & \sv{0.7955 ± 0.0201}  \\ 
			Youtube-Reddit-Large & 0.7539 ± 0.0005 & 0.7554 ± 0.0003 & 0.7243 ± 0.0016 & 0.7501 ± 0.0019 & 0.7327 ± 0.0016 & \sv{0.9128 ± 0.0031} & \fv{0.9863 ± 0.006}  \\ 
			Taobao-Large & 0.7075 ± 0.0009 & 0.7042 ± 0.0006 & 0.6812 ± 0.0032 & 0.5222 ± 0.0041 & 0.7787 ± 0.0103 & \sv{0.869 ± 0.010} & \fv{0.9933 ± 0.0008} \\ 
			\midrule
			\textbf{Average Rank} &4 & 4.5 & 5.5 & 6.25 & 4.75 & 1.75 & 1.25\\ 
			\midrule
			&\multicolumn{7}{|c}{\textbf{Inductive New-Old}}  \\
			\midrule
			eBay-Small & 0.9862 ± 0.0003 & 0.9836 ± 0.0016 & 0.9947 ± 0.0009 & 0.9712 ± 0.002 & 0.9985 ± 0.0 & \sv{0.9988 ± 0.0} & \fv{0.9999 ± 0.0}  \\ 
			YouTubeReddit-Small & 0.7695 ± 0.001 & 0.7655 ± 0.0018 & 0.7396 ± 0.0034 & 0.7242 ± 0.0004 & 0.7573 ± 0.0022 & \sv{0.922 ± 0.0002} & \fv{0.9967 ± 0.0014}  \\ 
			eBay-Large & 0.6109 ± 0.0244 & 0.5906 ± 0.0087 & 0.8134 ± 0.0105 & 0.6363 ± 0.0605 & \sv{0.9569 ± 0.0007} & 0.8973 ± 0.0 & \fv{1.0 ± 0.0}  \\ 
			DGraphFin & 0.5768 ± 0.0071 & 0.5735 ± 0.0007 & 0.5564 ± 0.0021 & 0.5742 ± 0.013 & 0.5646 ± 0.0244 & \sv{0.7702 ± 0.0043} & \fv{0.8693 ± 0.0066}  \\ 
			Youtube-Reddit-Large & 0.7844 ± 0.0015 & 0.7894 ± 0.0017 & 0.7623 ± 0.0031 & 0.7457 ± 0.0062 & 0.7511 ± 0.0022 & \sv{0.9356 ± 0.0004} & \fv{0.9958 ± 0.0025}  \\ 
			Taobao-Large & 0.7023 ± 0.0015 & 0.6953 ± 0.0022 & 0.6771 ± 0.0055 & 0.5104 ± 0.0106 & 0.7674 ± 0.005 & \sv{0.8458 ± 0.0043} & \fv{0.9965 ± 0.0005} \\ 
			\midrule
			\textbf{Average Rank} &4.25 & 5 & 5.5 & 5.75 & 4.25 & 2.25 & 1 \\ 
			\midrule
			
			&\multicolumn{7}{|c}{\textbf{Inductive New-New}}  \\
			\midrule
			eBay-Small & 0.9388 ± 0.0009 & 0.9366 ± 0.0037 & 0.9838 ± 0.0007 & 0.9556 ± 0.0007 & 0.9937 ± 0.0 & \sv{0.9975 ± 0.0} & \fv{0.9997 ± 0.0004}  \\ 
			YouTubeReddit-Small & 0.7436 ± 0.0015 & 0.7436 ± 0.0018 & 0.7265 ± 0.0055 & 0.749 ± 0.0011 & 0.7479 ± 0.004 & \sv{0.864 ± 0.0071} & \fv{0.9868 ± 0.0049}  \\ 
			eBay-Large & 0.7526 ± 0.0013 & 0.7500 ± 0.0005 & 0.7639 ± 0.0027 & 0.5196 ± 0.0002 & 0.9542 ± 0.0003 & \sv{0.9615 ± 0.0} & \fv{0.9999 ± 0.0001}  \\ 
			DGraphFin & 0.7307 ± 0.0007 & 0.7323 ± 0.0002 & 0.6843 ± 0.0131 & 0.5649 ± 0.0248 & 0.5417 ± 0.0099 & \fv{0.9051 ± 0.0028} & \sv{0.7584 ± 0.0323}  \\ 
			Youtube-Reddit-Large & 0.6932 ± 0.0026 & 0.7022 ± 0.0007 & 0.6703 ± 0.0024 & 0.7269 ± 0.0 & 0.6942 ± 0.0028 & \sv{0.8716 ± 0.0077} & \fv{0.9796 ± 0.0103}  \\ 
			Taobao-Large & 0.7243 ± 0.0001 & 0.7247 ± 0.0001 & 0.6885 ± 0.0024 & 0.5256 ± 0.0054 & 0.7922 ± 0.0118 & \sv{0.8906 ± 0.0088} & \fv{0.9969 ± 0.0002} \\ 
			\midrule
			\textbf{Average Rank} &5 & 4.25 & 5.5 & 5.75 & 4.5 & 1.75 & 1.25\\ 
			\midrule
			&JODIE&DyRep&TGN&TGAT&CAWN&NeurTW&NAT\\\midrule
			\textbf{Total Rank} &4.44  & 4.56  & 4.81  & 5.88  & 4.88  & \sv{2.00}  & \fv{1.44}\\ 
			\bottomrule
		\end{tabular}
	\end{table}

	\begin{table}[!t]
		\tiny
		\setlength{\tabcolsep}{4.5pt}
		\centering
		\caption{\footnotesize AP results of new datasets on the \textit{dynamic link prediction task}.
			The best and second-best results are highlighted as \fv{bold red} and \sv{underlined blue}.
			We do not highlight the second-best if the gap is $>0.05$ compared with the best result.}
		\label{tab:lp-ap}
		\begin{tabular}{l|lllllll}
			\toprule
			&\multicolumn{7}{|c}{\textbf{Transductive}}\\
			\midrule
			\diagbox{Dataset}{Model} &JODIE&DyRep&TGN&TGAT&CAWN&NeurTW&NAT\\
			\midrule
			eBay-Small & 0.9938 ± 0.0004 & 0.9936 ± 0.0006 &\sv{0.9983 ± 0.0003} & 0.9819 ± 0.0009 & 0.9981 ± 0.0 & \fv{0.9991 ± 0.0} & 0.9975 ± 0.0002  \\ 
			YouTubeReddit-Small & \sv{0.8612 ± 0.0009} & 0.8594 ± 0.0012 & 0.8421 ± 0.0041 & 0.8515 ± 0.0012 & 0.7625 ± 0.0042 & \fv{0.9112 ± 0.0021} & 0.8325 ± 0.0068  \\ 
			eBay-Large & 0.9318 ± 0.0002 & 0.9322 ± 0.0002 &\sv{ 0.9357 ± 0.0006} & 0.5239 ± 0.0002 & 0.9144 ± 0.0004 & 0.9307 ± 0.0 & \fv{0.9398 ± 0.0004}  \\ 
			DGraphFin & 0.7705 ± 0.0009 & 0.7705 ± 0.0024 & \sv{0.8571 ± 0.0009} & 0.6441 ± 0.0123 & 0.5431 ± 0.0095 & \fv{0.8637 ± 0.0014} & 0.7956 ± 0.0012  \\ 
			Youtube-Reddit-Large & 0.8622 ± 0.0007 & \sv{0.8632 ± 0.0004} & 0.8476 ± 0.0022 & 0.8591 ± 0.0026 & 0.7475 ± 0.0017 & \fv{0.9222 ± 0.0013} & 0.8628 ± 0.0015  \\ 
			Taobao-Large & 0.7164 ± 0.0003 & 0.7142 ± 0.0008 & \sv{0.844 ± 0.0011} & 0.5761 ± 0.0023 & 0.7616 ± 0.0069 & \fv{0.8568 ± 0.016} & 0.7904 ± 0.0008  \\ 
			\midrule
			&\multicolumn{7}{|c}{\textbf{Inductive}}  \\
			\midrule
			eBay-Small & 0.9638 ± 0.0007 & 0.9619 ± 0.0017 & 0.9898 ± 0.0005 & 0.9675 ± 0.0007 & 0.9953 ± 0.0002 & \sv{0.9982 ± 0.0} & \fv{0.9998 ± 0.0001}  \\ 
			YouTubeReddit-Small & 0.7866 ± 0.0007 & 0.7833 ± 0.0009 & 0.7387 ± 0.0069 & 0.7551 ± 0.0002 & 0.7568 ± 0.0031 & \sv{0.9086 ± 0.0022} & \fv{0.9872 ± 0.0056}  \\ 
			eBay-Large & 0.6989 ± 0.0018 & 0.6973 ± 0.0007 & 0.7096 ± 0.0030 & 0.518 ± 0.0002 & 0.9174 ± 0.0001 & \sv{0.9308 ± 0.0} & \fv{0.9999 ± 0.0001}  \\ 
			DGraphFin & 0.6563 ± 0.002 & 0.6567 ± 0.0009 & 0.624 ± 0.006 & 0.5866 ± 0.0123 & 0.5428 ± 0.0082 & \fv{0.8626 ± 0.0012} & \sv{0.7053 ± 0.0185}  \\ 
			Youtube-Reddit-Large & 0.7796 ± 0.0009 & 0.7818 ± 0.0009 & 0.73 ± 0.0029 & 0.7587 ± 0.0025 & 0.7353 ± 0.0022 & \sv{0.9192 ± 0.0022} & \fv{0.9849 ± 0.0071}  \\ 
			Taobao-Large & 0.6763 ± 0.0011 & 0.6746 ± 0.0011 & 0.6664 ± 0.0012 & 0.5315 ± 0.0027 & 0.7533 ± 0.011 & \sv{0.8596 ± 0.0205} & \fv{0.9941 ± 0.0007} \\ 
			\midrule
			&\multicolumn{7}{|c}{\textbf{Inductive New-Old}}  \\
			\midrule
			eBay-Small & 0.9849 ± 0.0007 & 0.9836 ± 0.0013 & 0.9931 ± 0.0008 & 0.9682 ± 0.0028 & 0.9985 ± 0.0001 & \sv{0.999 ± 0.0} & \fv{0.9999 ± 0.0}  \\ 
			YouTubeReddit-Small & 0.7963 ± 0.0013 & 0.7937 ± 0.0014 & 0.729 ± 0.0086 & 0.7296 ± 0.0013 & 0.762 ± 0.0041 & \sv{0.9244 ± 0.0015} & \fv{0.9966 ± 0.0016}  \\ 
			eBay-Large & 0.5670 ± 0.0186 & 0.5870 ± 0.0074 & 0.8024 ± 0.0060 & 0.6504 ± 0.0385 & \sv{0.9592 ± 0.0008} & 0.8458 ± 0.0 & \fv{1.0 ± 0.0}  \\ 
			DGraphFin & 0.6005 ± 0.0048 & 0.5872 ± 0.0059 & 0.5753 ± 0.0062 & 0.5927 ± 0.0058 & 0.5669 ± 0.0269 & \sv{0.7572 ± 0.0025} & \fv{0.8184 ± 0.0088}  \\ 
			Youtube-Reddit-Large & 0.808 ± 0.0014 & 0.8142 ± 0.0019 & 0.7472 ± 0.0043 & 0.7526 ± 0.0097 & 0.7553 ± 0.0025 & \sv{0.9368 ± 0.0009} & \fv{0.9953 ± 0.0028}  \\ 
			Taobao-Large & 0.7009 ± 0.0013 & 0.698 ± 0.0014 & 0.6879 ± 0.0008 & 0.5254 ± 0.0074 & 0.7597 ± 0.0053 & \sv{0.8459 ± 0.0103} & \fv{0.9969 ± 0.0004}  \\ 
			\midrule
			&\multicolumn{7}{|c}{\textbf{Inductive New-New}}  \\
			\midrule
			eBay-Small & 0.923 ± 0.001 & 0.9226 ± 0.0024 & 0.98 ± 0.0007 & 0.9505 ± 0.0009 & 0.991 ± 0.0001 & \sv{0.9973 ± 0.0} & \fv{0.9997 ± 0.0004}  \\ 
			YouTubeReddit-Small & 0.7578 ± 0.0015 & 0.7582 ± 0.0021 & 0.7564 ± 0.0043 & 0.7718 ± 0.0023 & 0.7498 ± 0.004 & \sv{0.8868 ± 0.0034} & \fv{0.9861 ± 0.0063}  \\ 
			eBay-Large & 0.6976 ± 0.0016 & 0.6957 ± 0.0007 & 0.7078 ± 0.0031 & 0.5154 ± 0.0001 & 0.93 ± 0.0003 & \sv{0.9318 ± 0.0} & \fv{0.9999 ± 0.0001}  \\ 
			DGraphFin & 0.6802 ± 0.0005 & 0.6811 ± 0.0002 & 0.6526 ± 0.0098 & 0.5831 ± 0.0184 & 0.5379 ± 0.0071 & \fv{0.8977 ± 0.0014} & 0.6529 ± 0.0249  \\ 
			Youtube-Reddit-Large & 0.7038 ± 0.0024 & 0.7115 ± 0.0007 & 0.6979 ± 0.002 & 0.7414 ± 0.0012 & 0.6965 ± 0.004 & \sv{0.8848 ± 0.0023} & \fv{0.9761 ± 0.0134}  \\ 
			Taobao-Large & 0.6738 ± 0.0005 & 0.6742 ± 0.0005 & 0.6611 ± 0.0011 & 0.53 ± 0.0023 & 0.7521 ± 0.0127 & \sv{0.8738 ± 0.0145} &\fv{ 0.9973 ± 0.0001} \\ 
			\bottomrule
		\end{tabular}
	\end{table}

	\subsubsection{ Node Classification Task}
	
	The eBay-Small and eBay-Large datasets have node labels, so
	we conduct dynamic node classification  experiments on both
	the eBay-Small and eBay-Large datasets. The AUC results are shown in Table \ref{tab:nc-auc}.  We can observe the similar results as in the paper. NeurTW achieves the best performance on both eBay-Small and eBay-Large datasets. NAT performs poorly on the node classification task. 
	
	\begin{table}[!t]
		\tiny
		\setlength{\tabcolsep}{4pt}
		\caption{\footnotesize ROC AUC results for the \textit{dynamic node classification task} on the eBay datasets. 
			The top-2 results are highlighted as \fv{bold red} and \sv{underlined blue}.}
		\label{tab:nc-auc}
		\centering
		\begin{tabular}{l|lllllll}
			\toprule
			\diagbox{Dataset}{Model} &JODIE&DyRep&TGN&TGAT&CAWN&NeurTW&NAT\\
			\midrule
			eBay-Small & 0.9274 ± 0.0017 & 0.8677 ± 0.0356 & 0.913 ± 0.0025 & \sv{0.9342 ± 0.0002} & 0.9305 ± 0.0001 & \fv{0.9529 ± 0.0002} & 0.6797 ± 0.0115  \\ 
			eBay-Large & 0.7244 ± 0.0002 & 0.7246 ± 0.0 & 0.6586 ± 0.0129 & 0.672 ± 0.0016 & \sv{0.7710 ± 0.0002} & \fv{0.7859 ± 0.0} & 0.5304 ± 0.0011 \\
			\midrule
			\textbf{Average Rank} &4 & 4.5 & 5.5 & 3.5 & \sv{2.5} & \fv{1} & 7  \\ 
			\bottomrule
		\end{tabular}
	\end{table}
	
	\subsubsection{Efficiency - the inference time}
	
\begin{figure*}[!t]
		\centering
		\includegraphics[width=0.6\textwidth]{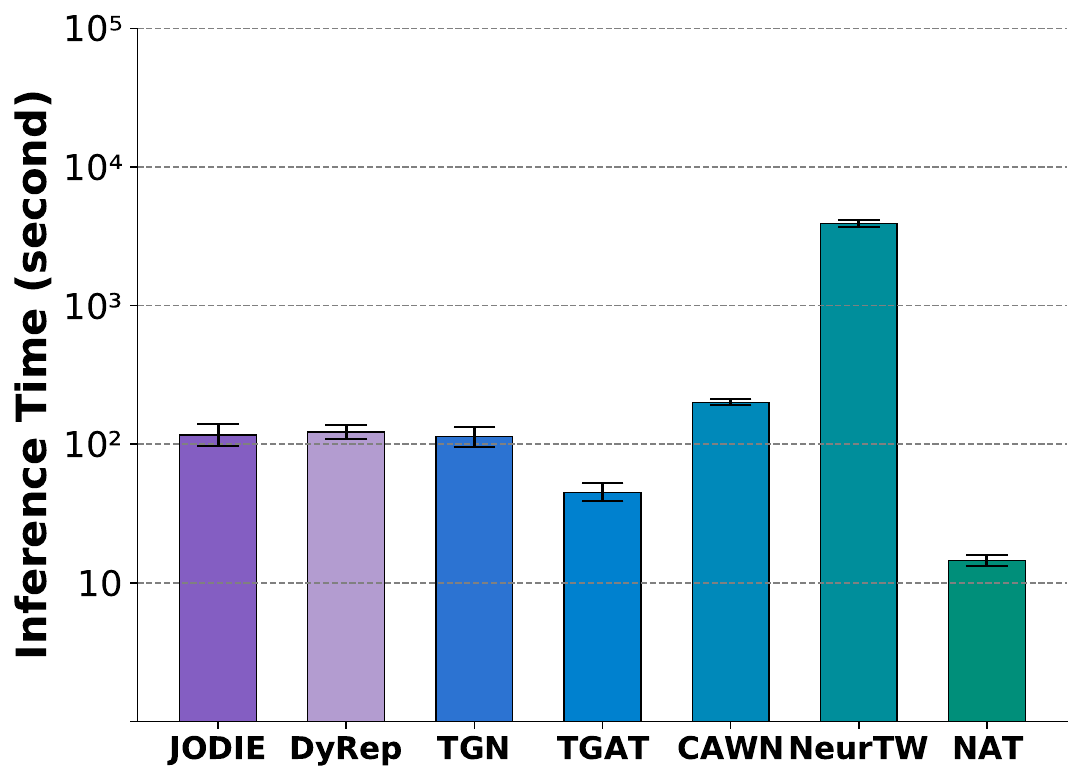}
		\caption{Inference time comparison per 100,000 edges.}
		\label{fig:inferencetime}
	\end{figure*}

	Considering many real world applications,
	we add   \textbf{the inference time} metric to evaluate the efficiency of methods. The inference time comparison per 100,000 edges is shown in Figure \ref{fig:inferencetime}. According to the figure, we can observe the similar model efficiency results  as in the paper. In terms of the inference time, JODIE, DyRep, TGN and TGAT are faster, while CAWN and NeurTW are much slower.
	NAT is relatively faster than temporal walk-based methods through caching and parallelism optimizations, {\em achieving a good trade-off between model quality and efficiency}.

	\subsubsection{Efficiency - Runtime, RAM, GPU}
	We have added model efficiency results for the newly added datasets as follows. We will add all these results to the Appendix (\url{https://openreview.net/attachment?id=rnZm2vQq31&name=supplementary_material}).
	
	Since many real-world graphs are extremely large, we believe efficiency is a vital issue for TGNNs in practice.
	We thereby compare the efficiency of the evaluated models on the newly added datasets (eBay-Small, eBay-Large, Taobal-Large,~DGraphFin,~YouTubeReddit-Small,~YouTubeReddit-Large), and
	present the results for dynamic link prediction task in Table~\ref{tab:newdata-lp-efficiency}, while dynamic node classification task  Table~\ref{tab:newdata-nc-efficiency}. 
	
	The Runtime in Table~\ref{tab:newdata-lp-efficiency} and Table~\ref{tab:newdata-nc-efficiency} shows that NAT is always trained much faster than the others  and need a low RAM and GPU Memory. TGAT obtains the second-best efficiency performance on the newly added datasets. 
	JODIE, DyRep, TGN achieve similar efficiency performance.
	We observe similar results as the main paper, NeurTW performs poorly on model efficiency.\\\\

	\begin{table}[!h]
		\small
		\setlength{\tabcolsep}{4.5pt}
		\centering
		\caption{\footnotesize Model efficiency for the newly added datasets on \textit{the link prediction task}. 
			We report seconds per epoch as \textbf{Runtime}, 
			the maximum RAM usage as \textbf{RAM}, and the maximum GPU memory usage  as \textbf{GPU Memory}, respectively. The best and second-best results are highlighted as \fv{bold red} and \sv{underlined blue}.}
		
		\label{tab:newdata-lp-efficiency}
		\begin{tabular}{l|rrrrrrr}
			\toprule
			&\multicolumn{7}{|c}{\textbf{Runtime} (second)}\\
			\midrule
			\diagbox{Dataset}{Model} &JODIE&DyRep&TGN&TGAT&CAWN&NeurTW&NAT\\
			\midrule
			eBay-Small & 749.80  & 801.58  & 905.19  & \sv{61.05}  & 1,385.54  & 1,556.32  & \fv{25.12}   \\ 
			YouTubeReddit-Small & 213.92  & 227.99  & 214.17  & \sv{85.59}  & 378.94  & 7,459.92  & \fv{29.51}   \\ 
			eBay-Large & 28,203.53  & 30,151.18  & 30,286.88  & \sv{791.86}  & 52,116.62  & 58,540.48  & \fv{117.38}   \\ 
			DGraphFin & 4,579.52  & 4,210.48  & 4,397.32  & \sv{1,708.71}  & 30,144.25  & 81,653.89  & \fv{904.38}   \\ 
			Youtube-Reddit-Large & 4,630.49  & 4,935.05  & 4,635.91  & \sv{1,852.67}  & 8,202.50  & 161,476.80  & \fv{638.77}   \\ 
			Taobao-Large & 3,108.45  & 2,931.87  & 2,860.83  & \sv{2,658.34}  & 12,143.02  & 148,922.55  & \fv{6654.56}  \\ 
			\midrule
			&\multicolumn{7}{|c}{\textbf{RAM} (GB)}  \\
			\midrule
			eBay-Small & 7.8  & \sv{6.2}  & 6.8  & \fv{4.3}  & 9.1  & 7.8  & \fv{4.3}   \\ 
			YouTubeReddit-Small & 6.8  & 7.2  & 6.6  & \sv{5.3}  & 13.1  & 8.1  & \fv{4.5}   \\ 
			eBay-Large & 20.2  & 18.3  & 19.1  & \sv{5.2}  & 17.1  & 10.1  & \fv{5.5}   \\ 
			DGraphFin & 17.5  & 15.3  & 17.5  & \sv{8.3}  & 23.2  & 24.3  & \fv{6.9}   \\ 
			Youtube-Reddit-Large & 26.3  & 16.6  & 18.9  & \sv{7.9}  & 18.5  & 21.3  & \fv{6.3}   \\ 
			Taobao-Large & 14.3  & 12.1  & 13.4  & \sv{7.5}  & 18.1  & 20.7  & \fv{6.2}  \\ 
			\midrule
			&\multicolumn{7}{|c}{\textbf{GPU Memory} (GB)}  \\
			\midrule
			eBay-Small & 2.0  & 1.9  & 2.0  & 1.9  & \sv{1.8}  & \fv{1.6}  & 2.2   \\ 
			YouTubeReddit-Small & \sv{1.3}  & 1.4  & 2.1  & \sv{1.3}  & 1.8  & \fv{1.1}  & \fv{1.1}   \\ 
			eBay-Large & 29.7  & 24.6  & 30.9  & 5.8  &\sv{5.7}  & \fv{3.0}  & 5.9   \\ 
			DGraphFin & 19.3  & 18.5  & 16.1  & 6.3  & 6.9  & \sv{6.1}  & \fv{6.0}   \\ 
			Youtube-Reddit-Large & 22.1  & 23.0  & 23.4  & 7.8  & \fv{6.3}  & 7.2  & \sv{7.1}   \\ 
			Taobao-Large & 20.3  & 21.8  & 19.6  & 7.7  & 7.3  & \sv{6.8}  & \fv{5.6}  \\ 
			\bottomrule
		\end{tabular}
	\end{table}

	\begin{table}[!h]
		\small
		\setlength{\tabcolsep}{4.5pt}
		\centering
		\caption{\footnotesize Model efficiency for the newly added datasets on \textit{the node classification task}. 
			We report seconds per epoch as \textbf{Runtime}, 
			the maximum RAM usage as \textbf{RAM}, and the maximum GPU memory usage  as \textbf{GPU Memory}, respectively. The best and second-best results are highlighted as \fv{bold red} and \sv{underlined blue}.}
		
		\label{tab:newdata-nc-efficiency}
		\begin{tabular}{l|rrrrrrr}
			\toprule
			&\multicolumn{7}{|c}{\textbf{Runtime} (second)}\\
			\midrule
			\diagbox{Dataset}{Model} &JODIE&DyRep&TGN&TGAT&CAWN&NeurTW&NAT\\
			\midrule
			eBay-Small & 765.05  & 794.03  & 718.56  & \sv{55.05}  & 226.56  & 583.08  & \fv{13.05}   \\ 
			eBay-Large & 29,153.28  & 29,867.17  & 27,028.53  & \sv{629.52}  & 8,522.04  & 25,693.71  & \fv{97.54}  \\ 
			
			\midrule
			&\multicolumn{7}{|c}{\textbf{RAM} (GB)}  \\
			\midrule
			eBay-Small & 6.5  & 6.8  & 6.7  & \sv{4.2}  & 6.9  & 7.2  & \fv{4.1}   \\ 
			eBay-Large & 41.8 & 39.2 & 20.5 & \fv{5.2} & 15.1 & 7.4 & \sv{5.8} \\ 
			
			\midrule
			&\multicolumn{7}{|c}{\textbf{GPU Memory} (GB)}  \\
			\midrule
			eBay-Small & 1.8  & \fv{1.2}  & \sv{1.5}  & 1.8  & 1.9  & 1.8  & 2.3   \\ 
			eBay-Large & 31.7 & 31 & 31.4 & \sv{5.8} & \sv{5.8} & \fv{2.9} & 5.9 \\ 
			\bottomrule
		\end{tabular}
	\end{table}
	
	\section{Node Classification Task with Multiple Node Labels}
	We have added experiments of the node classification task with multiple label numbers.
	
	In the previous works, only Reddit, Wikipedia, and MOOC datasets have node labels (two labels: 0 and 1) and are used for binary node classification task. 
	Through unremitting efforts, we have finded a large-scale temporal dataset - DGraphFin~\cite{huang2022dgraph} with  multiple node labels.
	DGraphFin consists of 3,700,550 nodes and 4,300,999 edges.
	4,300,999 edges.
	
	\href{https://dgraph.xinye.com/dataset}{\textbf{DGraphFin}} is a collection of large-scale dynamic graph datasets, consisting of interactive objects, events and labels that evolve with time.It is a directed, unweighted dynamic graph consisting of millions of nodes and edges, representing a realistic user-to-user social network in financial industry. Nodes are users, and an edge from one user to another means that the user regards the other user as the emergency contact person~\cite{huang2022dgraph}.
	
	There four classes. Below are the nodes counts of each class.  
	\begin{itemize}
		\item 0: 1210092 
		\item 1: 15509   
		\item 2: 1620851 
		\item 3: 854098     
	\end{itemize} 
	Nodes of Class 1 are fraud users and nodes of 0 are normal users, and they the two classes to be predicted.    
	Nodes of Class 2 and Class 3 are background users.  
	
	We preprocess DGraphFin as the format of temporal graph. We open source the code of preprocessing DGraphFin dataset at \url{https://github.com/qianghuangwhu/benchtemp/blob/master/DGraphFin/DGraphFin.py}.
	
	DGraphFin dataset has been hosted on the open-source platform zenodo (https://zenodo.org/) with a Digital Object Identifier (DOI) 10.5281/zenodo.8267771 (\url{https://zenodo.org/record/8267846}).
	
	The experimental results for dynamic node classification task on DGraphFin dataset are shown in Table \ref{tab:nc-multi-label}. Different evaluation metrics are available, including Accuracy, Precision, Recall and F1.
	
	\begin{equation}
		\begin{gathered}
			\text { Precision }_{\text {weighted }}=\frac{\sum_{i=1}^N \text { Support }_i \times \text { Precision }_i}{\sum_{i=1}^N \text { Support }_i} \\
			\text { Recall }_{\text {weighted }}=\frac{\sum_{i=1}^N \text { Support }_i \times \text { Recall }_i}{\sum_{i=1}^N \text { Support }_i} \\
			\mathrm{~F} 1_{\text {weighted }}=\frac{2 \times \text { Precision }_{\text {weighted }} \times \text { Recall }_{\text {weighted }}}{\text { Precision }_{\text {weighted }}+\text { Recall }_{\text {weighted }}}
		\end{gathered}
	\end{equation}
	
	where Support $_i$ is the number of supports for the $i$-th class.
	
	\begin{table}[h]
		\tiny
		\setlength{\tabcolsep}{4.5pt}
		\centering
		\caption{\footnotesize The experimental results for dynamic node classification task with multiple labels on DGraphFin dataset. }
		\label{tab:nc-multi-label}
		\begin{tabular}{l|lllllll}
			\toprule
			&JODIE&DyRep&TGN&TGAT&CAWN&NeurTW&NAT\\
			\midrule
			Accuracy & 0.4396 ± 0.0070 & 0.4400 ± 0.0117 & \fv{0.5696 ± 0.0056} & \sv{0.5529 ± 0.0036}& 0.4366±0.0068& 0.4933±0.0870& 0.4131±0.0038\\ 
			\midrule
			Precision & 0.1933 ± 0.0062 & 0.1937 ± 0.0103 & \fv{0.4806 ± 0.0053} & \sv{0.4644 ± 0.0118}  & 0.1906±0.0060&0.3879±0.2849&0.3068±0.0044\\ 
			Recall & 0.4396 ± 0.0070 & 0.4400 ± 0.0117 & \fv{0.5696 ± 0.0056} & \sv{0.5529 ± 0.0036} & 0.4366±0.0068 &0.4933±0.0870 &0.4131±0.0038\\ 
			F1 & 0.2685 ± 0.0073 & 0.2690 ± 0.0121 & \fv{0.4905 ± 0.0063} & \sv{0.4744 ± 0.0020} & 0.2654±0.0071&0.3727±0.1588&0.3402±0.0047\\ 
			\bottomrule
		\end{tabular}
	\end{table}
	
	As shown in Table \ref{tab:nc-multi-label}, TGN achieves the best performance on dynamic node classification task with multiple labels, followed by TGAT. JODIE, DyRep, and CAWN perform poorly.

	\section{Ablation Studies on NODEs of NeurTW}
	CAWN~\cite{wanginductive} and NeurTW~\cite{jin2022neural} perform well on the link prediction task and are both based on motifs and index anonymization operation. However, NeurTW~\cite{jin2022neural} additionally constructs  neural ordinary
	differential equations (NODEs). With a component based on neural ordinary
	differential equations, the extracted motifs allow for irregularly-sampled temporal
	nodes to be embedded explicitly over \textit{multiple different interaction time intervals},
	enabling the effective capture of the underlying spatiotemporal dynamics.
	
	In NeurTW~\cite{jin2022neural}, The \textit{Continuous Evolution} operation is illustrated in Equation 8 in the original paper (\url{https://openreview.net/pdf?id=NqbktPUkZf7}).
	\begin{equation}
		h_i^{\prime}=h_{i-1}+\int_{t_{i-1}}^{t_i} f\left(h_t, \theta\right) dt,
	\end{equation}
	where $f\left(h_t, \theta\right)$ is the ODE function, implemented by an autoregressive gated recurrent unit with a parameter $\sigma$. The corresponding ODE function $\tilde{f}\left(\tilde{h}_s, s, \theta\right)$ follows:
	\begin{equation}
		\begin{aligned}
			\tilde{f}\left(\tilde{h}_s, s, \theta\right) & :=\frac{d \tilde{h}_s}{d s}=\left.\frac{d h_t}{d t}\right|_{t=s\left(t_{\text {end }}^c-t_{\text {start }}^c\right)+t_{\text {start }}^c} \frac{d t}{d s} \\
			& =\left.f\left(h_t, t, \theta\right)\right|_{t=s\left(t_{\text {end }}^c-t_{\text {start }}^c\right)+t_{\text {start }}^c}\left(t_{\text {end }}^c-t_{\text {start }}^c\right) \\
			& =f\left(\tilde{h}_s, s\left(t_{\text {end }}^c-t_{\text {start }}^c\right)+t_{\text {start }}^c, \theta\right)\left(t_{\text {end }}^c-t_{\text {start }}^c\right)
		\end{aligned}
	\end{equation}
	
	Thus, Due to the neural ordinary
	differential equations (NODEs)
	, NeurTW~\cite{jin2022neural} performs better on datasets with a large time granularity.
	
	\href{https://github.com/shenyangHuang/LAD}{CanParl} is a Canadian parliament bill voting
	network extracted from open \href{https://openparliament.ca/}{website} 
	\cite{huang2020laplacian}. Nodes are members of parliament (MPs), and edges are the interactions between  MPs from 2006 to 2019.
	\begin{figure*}[!h]
		\centering
		\includegraphics[width=0.6\textwidth]{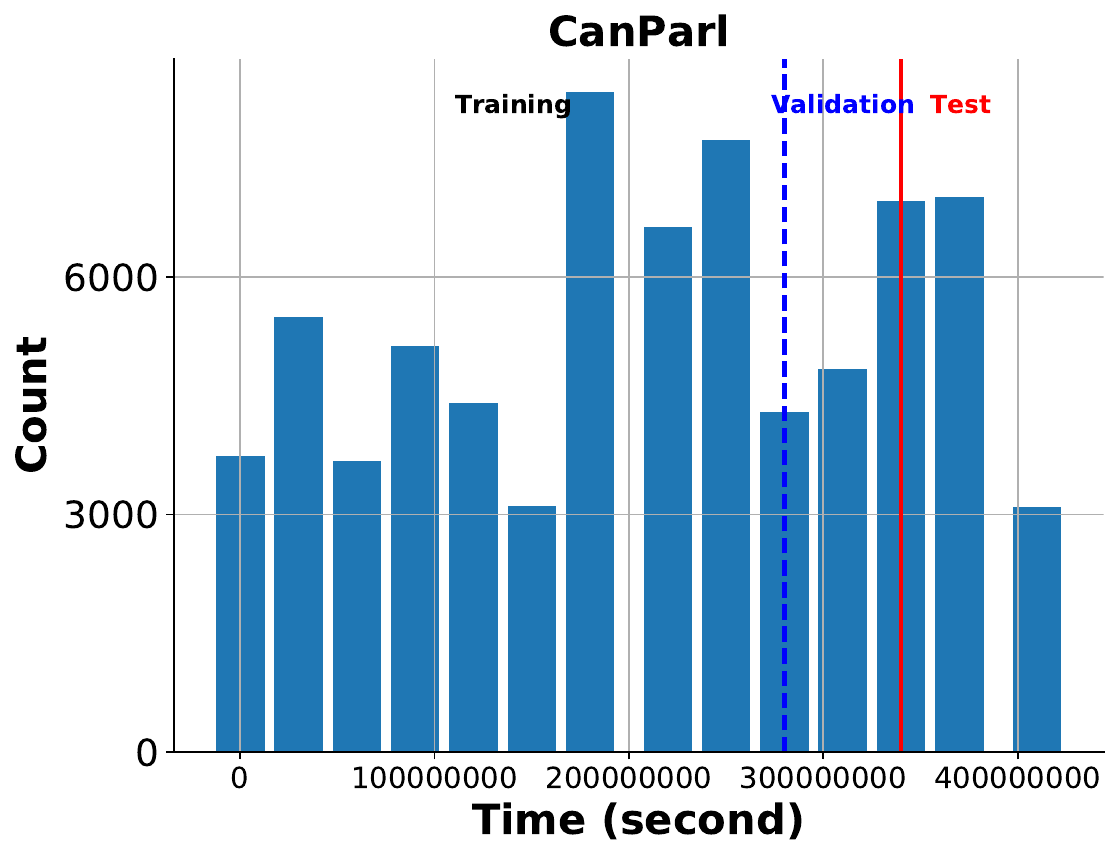}
		\caption{The distribution of temporal edge count for the CanParl dataset, and the illustration on the
			train-validation-test splitting.}
		\label{fig:CanParldistribution}
	\end{figure*}
	We illustrate the distribution of temporal edge count for the CanParl dataset in Figure \ref{fig:CanParldistribution}. As shown in Figure \ref{fig:CanParldistribution},  CanParl dataset has a large time granularity and NeurTW~\cite{jin2022neural} achieves the best performance on CanParl dataset. Inspired by the above analysis, we could infer
	that NeurTW is potentially suitable for datasets with a large time granularity and time intervals, such as CanParl.
	
	We further conduct ablation studies to verify the effectiveness of neural ordinary
	differential equations (NODEs) of NeurTW on datasets with a large time granularity and time intervals. The experimental results are  detailed in Table \ref{tab:Ablation-NODEs}.
	
	\begin{table}[h]
		\tiny
		\setlength{\tabcolsep}{2.5pt}
		\centering
		\caption{Ablation studies on neural ordinary
			differential equations (NODEs) of NeurTW. "\textcolor{red}{\textbf{--}} NODEs" means remove NODEs module.}
		\label{tab:Ablation-NODEs}
		\begin{tabular}{l|l|llll|llll}
			\toprule
			\multirow{2}{*}{Ablation}  &\multirow{2}{*}{Datasets}&\multicolumn{4}{c|}{AUC}&\multicolumn{4}{c}{AP}\\
			\cmidrule(lr){3-6}\cmidrule(lr){7-10}
			
			& &Transductive&Inductive&  New-Old&  New-New&Transductive&Inductive&  New-Old&  New-New\\
			\midrule
			\multirow{2}{*}{original} &CanParl&0.8920 ± 0.0173 &0.8871 ± 0.0139 &0.8847 ± 0.0102 &0.8882 ± 0.0045 &0.8528 ± 0.0213 &0.8469 ± 0.0161 &0.8417 ± 0.0132 &0.8511 ± 0.0079  \\
			&USLegis&0.9715 ± 0.0009&0.9708 ± 0.0009&0.9682 ± 0.0018&0.9787 ± 0.0004&0.9713 ± 0.0013&0.971 ± 0.0009&0.9671 ± 0.0027&0.9803 ± 0.0005\\
			\midrule
			\multirow{2}{*}{\textcolor{red}{\textbf{--}} NODEs} &CanParl&0.5 ± 0.0&0.5001 ± 0.0&0.5001 ± 0.0&0.5 ± 0.0&0.5 ± 0.0&0.5001 ± 0.0&0.5001 ± 0.0&0.5 ± 0.0 \\
			&USLegis&0.898 ± 0.004&0.9186 ± 0.0018&0.9026 ± 0.0025&0.9474 ± 0.0&0.8721 ± 0.0047&0.9037 ± 0.0034&0.8651 ± 0.0029&0.9458 ± 0.0004\\
			\bottomrule
		\end{tabular}
	\end{table}
	
	NeurTW without differential equations (NODEs) module performs much poorly on  datasets with a large time granularity and time intervals (such as, CanParl). However, on  a tiny time granularity and time intervals (such as, USLegis, the timestamp of USLegis is only from 0 to 11.), thus, removing the differential equations (NODEs) module has relatively little negative impact on the performance of the NeurTW.
	
	Ablation studies on neural ordinary
	differential equations (NODEs) of NeurTW verify that "NeurTW introduces a continuous-time operation that can depict evolution trajectory, which is potentially suitable for CanParl with a large time granularity and time intervals". 
	
	\section{Graph Density for CAWN with Random Subgraph Sampling}
	\begin{figure*}[!h]
		\centering
		\includegraphics[width=0.6\textwidth]{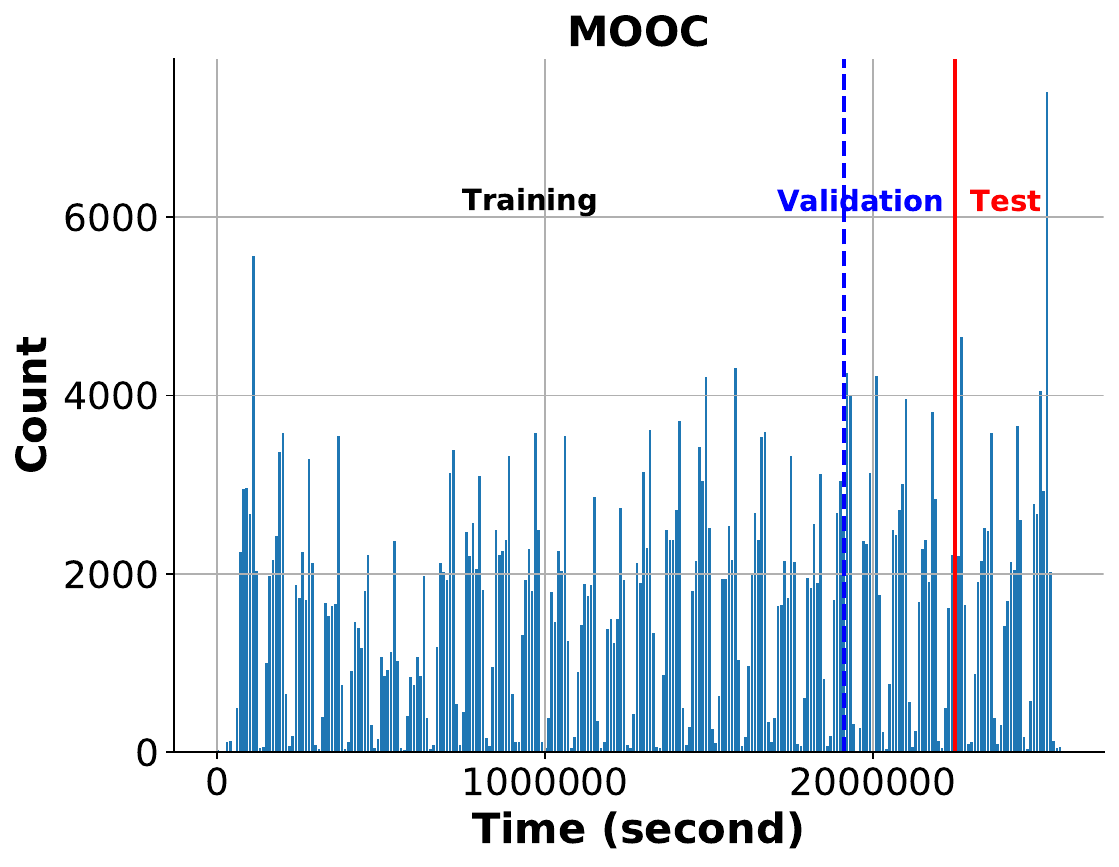}
		\caption{The distribution of temporal edge count for the MOOC dataset, and the illustration on the
			train-validation-test splitting.} 
		\label{fig:MOOCdistribution}
	\end{figure*}
	
	We have added experiments to demonstrate that the effectiveness of the temporal walk mechanism changes in response to changes in graph density.

	As shown in Figure \ref{fig:MOOCdistribution}, MOOC is relatively denser and CAWN based on the temporal walk mechanism (motifs) achieves the best performance on this dataset.

	To demonstrate that the effectiveness of the temporal walk mechanism changes in response to changes in graph density, we adopt edges sampling strategy. We randomly sample a  constant number $N_{e}$ of temporal edges $\{(u_{1}, i_{1}), \dots, (u_{N}, i_{N})\}$ each time as a subgraph $G_{S}$ of the original temporal graph. The number $N_{u}$ of the source nodes is the number of the elements in set $\{u_{1}, \dots, u_{N}\}$. The number $N_{i}$ of the source nodes is the number of the elements in set $\{i_{1}, \dots, i_{N}\}$.
	
	The graph density $\sigma_{D_{S}}$ in our paper is the temoral density of temoral edges $(u_{t}, i_{t})$, the calculation formula is as follows:
	\begin{equation}
		\sigma_{D_{S}} = \frac{N_{e}}{N_{u} \times N_{i}}.
	\end{equation}
	
	The number of sampled edges is a constant $N_{e}$. Thus, we sampled two subgraphs: $G_{S_{1}}$ and $G_{S_{2}}$. 
	The statistics of sampled subgraphs are shown in Table \ref{tab:lp-density}. 
	The temoral graph density of $G_{S_{1}}$ is 0.6320, while $G_{S_{2}}$ 0.3127.
	
	\begin{table}[h]
		\setlength{\tabcolsep}{4.5pt}
		\centering
		\caption{The parameters of sampled graphs $G_{S_{1}}$, $G_{S_{2}}$. }
		\label{tab:lp-density}
		\begin{tabular}{l|l|ll|l}
			\toprule
			&$N_{e}$&$N_{u}$&$N_{i}$&$\sigma_{D_{S}}$\\
			\midrule
			$G_{S_{1}}$ &100000 & 4395 & 36 & 0.6320   \\ 
			$G_{S_{2}}$  &100000 & 4264 & 75 & 0.3127  \\ 
			\bottomrule
		\end{tabular}
	\end{table}

	The experimental results of CAWN based on the temporal walk mechanism (motifs) on $G_{S_{1}}$ and $G_{S_{2}}$ are shown in Table \ref{tab:lp-density-results}.
	
	\begin{table}[h]
		\tiny
		\setlength{\tabcolsep}{2.5pt}
		\centering
		\caption{The experimental results of CAWN  on $G_{S_{1}}$ and $G_{S_{2}}$. }
		\label{tab:lp-density-results}
		\begin{tabular}{l|llll|llll}
			\toprule
			&\multicolumn{4}{c|}{AUC}&\multicolumn{4}{c}{AP}\\
			\midrule
			&Transductive&Inductive&Inductive New-Old&Inductive New-Old&Transductive&Inductive&Inductive New-Old&Inductive New-Old\\
			\midrule
			$G_{S_{1}}$ & 0.886 ± 0.0164 & 0.883 ± 0.0158 & 0.8847 ± 0.0166 & 0.8701 ± 0.0134 & 0.8651 ± 0.021 & 0.8601 ± 0.0219 & 0.8616 ± 0.0225 & 0.8511 ± 0.0174
			
			\\ 
			$G_{S_{2}}$ & 0.8357 ± 0.0073 & 0.8353 ± 0.0105 & 0.8535 ± 0.0087 & 0.7752 ± 0.0157 & 0.8172 ± 0.0088 & 0.8154 ± 0.0128 & 0.8337 ± 0.0103 & 0.7535 ± 0.0197  \\ 
			\bottomrule
		\end{tabular}
	\end{table}
	
	As shown in Table \ref{tab:lp-density-results}, CAWN performs much better on $G_{S_{1}}$ with a larger graph density, $\sigma_{D_{S_{1}}}=0.6320$ > $\sigma_{D_{S_{2}}}=0.3127$. The experimental results demonstrate that the effectiveness of the CAWN based on temporal walk mechanism changes in response to changes in graph density.
	

	\section{BenchTeMP with Historical Negative Sampling and Inductive Negative Sampling}
	As for NAT and NeurTW can achieved 90\%+ or even 95\%+ AUC and AP performance on many of these datasets, we have found that BenchTeMP with \textit{Historical Negative Sampling} and \textit{Inductive Negative Sampling}  can addressed this issue. \cite{poursafaeitowards}.

	\begin{figure*}[!h]
		\centering
		\includegraphics[width=0.7\textwidth]{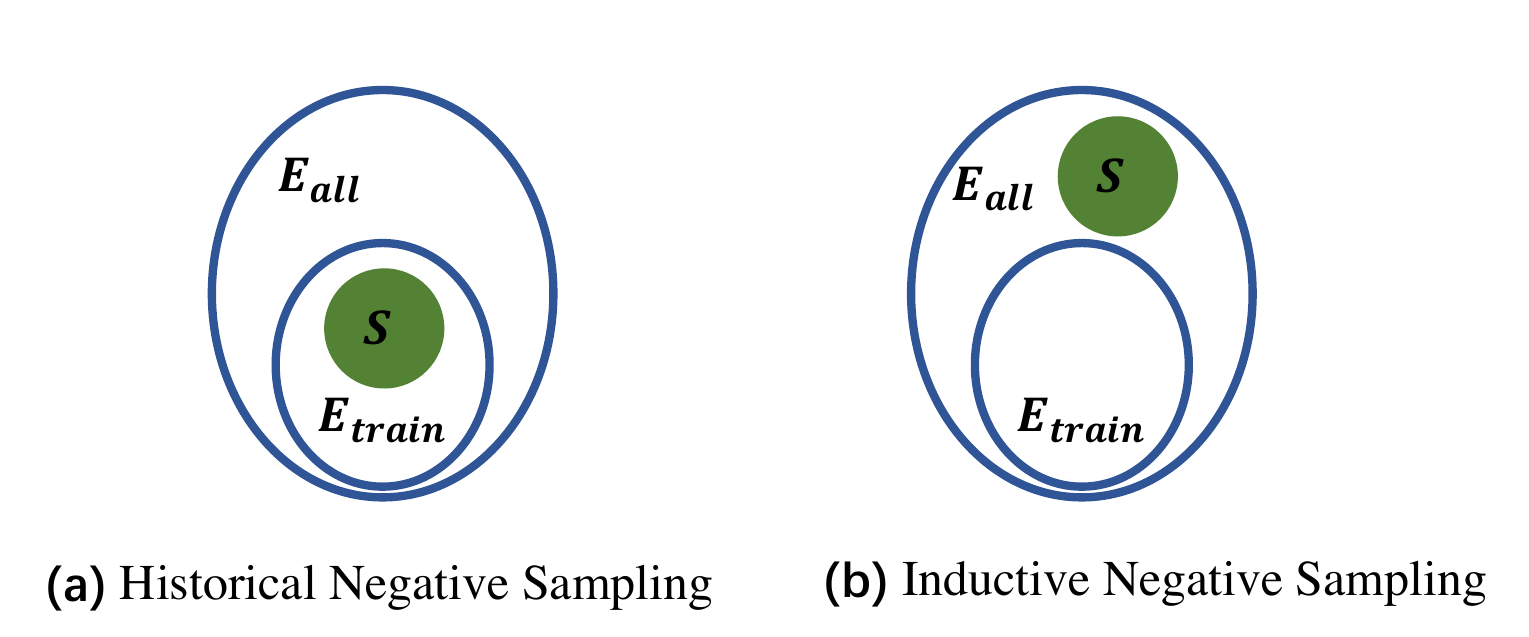}
		\caption{\textit{Historical Negative Sampling} and \textit{Inductive Negative Sampling}.}
		\label{fig:Negative-Sampling}
	\end{figure*}
	
	Let $E_{all}$, $E_{train}$, $S$ be the set of edges in  dataset, the set of edges in train dataset, and the set of negative sampling edges, respectively. \textit{Historical Negative Sampling} and \textit{Inductive Negative Sampling} are illustrated in Figure \ref{fig:Negative-Sampling}.
	
	\begin{itemize}
		\item \textbf{Historical Negative Sampling}. Sampling negative edges from the set
		of edges that have been observed during previous timestamps but are absent in the current step, i.e., Sampling negative edges in the  $E_{train}$.
		\item \textbf{Inductive Negative Sampling}. Sampling negative edges in the  $E_{all}$ but not observed during training.
	\end{itemize}

	We have conducted experiments of NAT on those over-performance datasets (Reddit, Wikipedia, Flights) with \textit{Historical Negative Sampling} and \textit{Inductive Negative Sampling}. Experimental results are shown in Table \ref{tab:negative-sampling-auc}  and Table \ref{tab:negative-sampling-ap}.

	\begin{table}[!h]
		\tiny
		\setlength{\tabcolsep}{4pt}
		\caption{\footnotesize ROC AUC results of NAT with \textit{Historical Negative Sampling} and \textit{Inductive Negative Sampling}   for the 
			dynamic  link prediction  task. 
		}
		\label{tab:negative-sampling-auc}
		\centering
		\begin{tabular}{l|l|llll}
			\toprule
			Sampling  &Datasets &Transductive&Inductive&Inductive New-Old&Inductive New-New\\
			\midrule
			\multirow{3}{*}{\textbf{Historical}} & Reddit & 0.7759 ± 0.0065 & 0.8272 ± 0.0036 & 0.8532 ± 0.0019 & 0.9097 ± 0.0014  \\ 
			&  Wikipedia & 0.6992 ± 0.0027 & 0.7924 ± 0.0022 & 0.8118 ± 0.0034 & 0.8448 ± 0.0015  \\ 
			&  Flights & 0.6145 ± 0.0034 & 0.6443 ± 0.0276 & 0.6418 ± 0.0566 & 0.8019 ± 0.0079 \\ 
			
			\midrule
			\multirow{3}{*}{\textbf{Inductive}}&Reddit & 0.8058 ± 0.0049 & 0.858 ± 0.0045 & 0.8746 ± 0.0035 & 0.9515 ± 0.001  \\ 
			& Wikipedia & 0.731 ± 0.0022 & 0.7609 ± 0.0009 & 0.7593 ± 0.001 & 0.8323 ± 0.0045  \\ 
			& Flights & 0.6145 ± 0.0034 & 0.6443 ± 0.0276 & 0.6418 ± 0.0566 & 0.8019 ± 0.0079 \\ 
			\bottomrule
		\end{tabular}
	\end{table}
	
	\begin{table}[!h]
		\tiny
		\setlength{\tabcolsep}{4pt}
		\caption{\footnotesize AP results of NAT with \textit{Historical Negative Sampling} and \textit{Inductive Negative Sampling}   for the 
			dynamic  link prediction  task. 
		}
		\label{tab:negative-sampling-ap}
		\centering
		\begin{tabular}{l|l|llll}
			\toprule
			Sampling  &Datasets &Transductive&Inductive&Inductive New-Old&Inductive New-New\\
			\midrule

			\multirow{3}{*}{\textbf{Historical}} & Reddit & 0.7958 ± 0.0097 & 0.8406 ± 0.0008 & 0.8558 ± 0.001 & 0.9063 ± 0.0033  \\  
			&  Wikipedia & 0.7128 ± 0.0019 & 0.7859 ± 0.0022 & 0.8017 ± 0.0031 & 0.8267 ± 0.0025  \\ 
			&  Flights & 0.6287 ± 0.0094 & 0.6596 ± 0.0248 & 0.6507 ± 0.0562 & 0.8428 ± 0.0042 \\  
			
			\midrule
			\multirow{3}{*}{\textbf{Inductive}}&Reddit & 0.8523 ± 0.0013 & 0.8902 ± 0.8979 & 0.8979 ± 0.0035 & 0.9648 ± 0.0009  \\ 
			& Wikipedia &  0.733 ± 0.0025 & 0.7525 ± 0.0043 & 0.747 ± 0.0 & 0.8238 ± 0.0088  \\  
			& Flights & 0.6287 ± 0.0094 & 0.6596 ± 0.0248 & 0.6507 ± 0.0562 & 0.8428 ± 0.0042 \\ 
			\bottomrule
		\end{tabular}
	\end{table}

	The experimental results demonstrated the effectiveness of \textit{Historical Negative Sampling} and \textit{Inductive Negative Sampling}. 
	
	We leave the exploration of BenchTeMP with \textit{Historical Negative Sampling} and \textit{Inductive Negative Sampling}  in detail  for future works.

\end{document}